\documentclass[10pt,journal,compsoc]{IEEEtran}

\ifCLASSOPTIONcompsoc
  \usepackage[nocompress]{cite}
\else
  \usepackage{cite}
\fi
\ifCLASSINFOpdf
\else
\fi
\usepackage{hyperref}
\usepackage{microtype}
\usepackage{graphicx}
\usepackage{subfigure}
\usepackage{booktabs} 

\usepackage{xcolor}
\usepackage{color, colortbl}
\definecolor{Gray}{gray}{0.9}
\usepackage{multirow}
\usepackage{amsmath}
\DeclareMathOperator*{\argmin}{arg\,min}
 
\DeclareMathOperator*{\argmaxC}{\arg\max}
\DeclareMathOperator*{\argminC}{\arg\min}
\usepackage{eucal}
\usepackage{bm} 

\usepackage{amsmath}
\usepackage{amssymb}
\usepackage{mathtools}
\usepackage{amsthm}

\usepackage[capitalize,noabbrev]{cleveref}

\newcommand{\R}{\mathbb{R}}
\newcommand{\iid}{\overset{\mathrm{iid}}{\sim}}

\usepackage[format=plain]{caption}
\usepackage{float}

\usepackage{algorithm}
\usepackage{algpseudocode}
\algrenewcommand\algorithmicrequire{\textbf{Input:}}
\algrenewcommand\algorithmicensure{\textbf{Output:}}
\algdef{SE}[VARIABLES]{Variables}{EndVariables}
   {\algorithmicvariables}
   {\algorithmicend\ \algorithmicvariables}
\algnewcommand{\algorithmicvariables}{\textbf{global variables}}

\newtheorem{Prop}{Property}

\begin{document}

\title{Simplex Clustering via $\mathtt{sBeta}$ with Applications to Online Adjustment of Black-Box Predictions}

\author{Florent~Chiaroni,
        Malik~Boudiaf,
        Amar~Mitiche
        and~Ismail~Ben~Ayed
\IEEEcompsocitemizethanks{\IEEEcompsocthanksitem
F. Chiaroni was with ETS Montreal and INRS, Montreal, QC, Canada during this work and is now with Thales Research and Technology (TRT) Canada, Thales Digital Solutions (TDS), CortAIx, Montreal, QC, Canada.\protect\\
E-mail: florent.chiaroni.ai@gmail.com
\IEEEcompsocthanksitem M. Boudiaf is with ETS Montreal, Montreal, QC, Canada.\protect\\
E-mail: malik.boudiaf.1@etsmtl.net
\IEEEcompsocthanksitem A. Mitiche is with INRS, Montreal, QC, Canada.\protect\\
E-mail: Amar.Mitiche@inrs.ca
\IEEEcompsocthanksitem I. Ben Ayed is with ETS Montreal, Montreal, Canada.\protect\\
E-mail: Ismail.BenAyed@etsmtl.ca}
}

\IEEEtitleabstractindextext{%
\begin{abstract}
We explore clustering the softmax predictions of deep neural networks and introduce a novel probabilistic clustering method, referred to as \textsc{k-sBetas}. In the general context of clustering discrete distributions, the existing methods focused on exploring distortion measures tailored to simplex data, such as the KL divergence, as alternatives to the standard Euclidean distance. We provide a general maximum a posteriori (MAP) perspective of clustering distributions, emphasizing that the statistical models underlying the existing distortion-based methods may not be descriptive enough. Instead, we optimize a mixed-variable objective measuring data conformity within each cluster to the introduced $\mathtt{sBeta}$ density function, whose parameters are constrained and estimated jointly with binary assignment variables. Our versatile formulation approximates various parametric densities for modeling simplex data and enables the control of the cluster-balance bias. This yields highly competitive performances for the unsupervised adjustment of black-box model predictions in various scenarios. Our code and comparisons with the existing simplex-clustering approaches and our introduced softmax-prediction benchmarks are publicly available: \url{https://github.com/fchiaroni/Clustering_Softmax_Predictions}.
\end{abstract}

}

\maketitle

\IEEEdisplaynontitleabstractindextext

\IEEEpeerreviewmaketitle

\ifCLASSOPTIONcompsoc
\IEEEraisesectionheading{\section{Introduction}\label{sec:introduction}}

\IEEEPARstart{O}{ver} the last decade, deep neural networks have continuously gained wide interest for the semantic analysis of real-world data. However, under real-world conditions, potential shifts in the feature or class distributions may affect the prediction performances of the pre-trained source models.
To address this issue, several recent adaptation studies investigated fine-tuning all or a part of the model's parameters using standard gradient-descent and back-propagation procedures \cite{wang2021tent,wang2018deep,liang2021source}. However, in a breadth of practical applications, e.g., real-time predictions, it might be cumbersome to perform multiple forward and backward passes. This is particularly true when dealing with large (and continuously growing) pre-trained networks, such as the recently emerging foundational vision-language models \cite{radford2021clip}. In such scenarios, even fine-tuning only parts of the pre-trained model, such as the parameters of the normalization layers \cite{wang2021tent}, might be computationally intensive. In addition, the existing adaptation techniques assume knowledge of the source model and its training procedures. Such assumptions may not always hold in practice due to data-privacy constraints \cite{Xia_2021_ICCV} and
the fact that many large-scale pre-trained models are only available through APIs \cite{ColomboEMNLP2023}, i.e., their pre-trained weights are not shared\footnote{This is the case, for instance, of large-scale foundation models in NLP such as the GPT family, Anthropic’s Claude or Google’s PaLM.}. Therefore, model-agnostic solutions, which enable computationally efficient adaptation of black-box pre-trained models, would be of significant interest in practice.

In this work, we propose to cluster the softmax predictions of deep neural networks. Solely based on the model's probabilistic outputs, this strategy yields computationally efficient adaptation while preserving the privacy of both the model and data.
Fig. \ref{fig_GTA_to_Cityscapes_road_seg} depicts an application example for road segmentation under domain shifts.
            \begin{figure}%
                \centering
                \begin{minipage}{0.23\textwidth}%
                \includegraphics[width=\columnwidth]{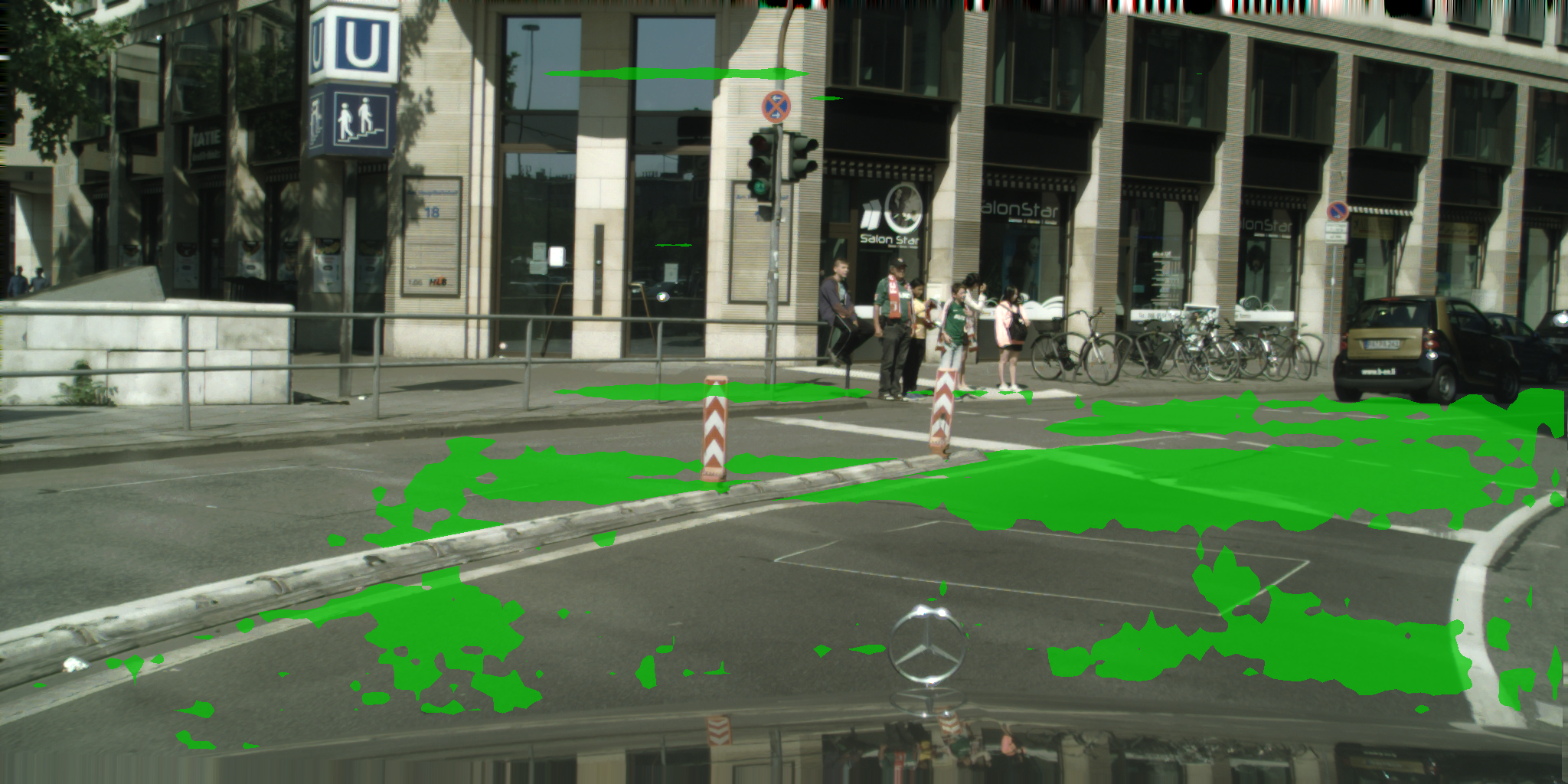}%
                \centering
                \\(a) argmax
                \end{minipage}%
                \hspace{0.005\textwidth}
                \begin{minipage}{0.23\textwidth}%
                \includegraphics[width=\columnwidth]{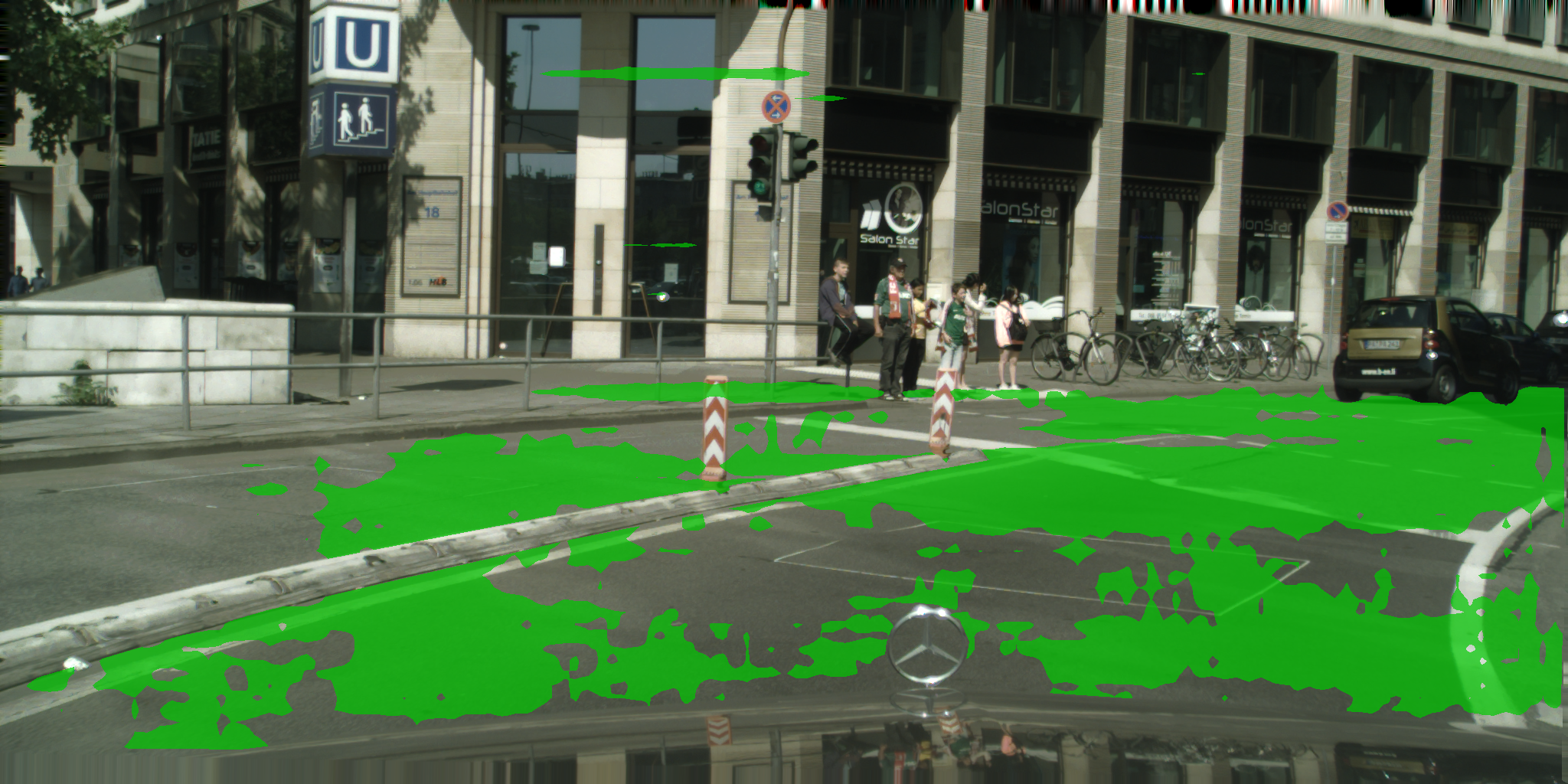}%
                \centering
                \\(b) k-means
                \end{minipage}%
                \vspace{0.005\textwidth}
                \begin{minipage}{0.23\textwidth}%
                \includegraphics[width=\columnwidth]{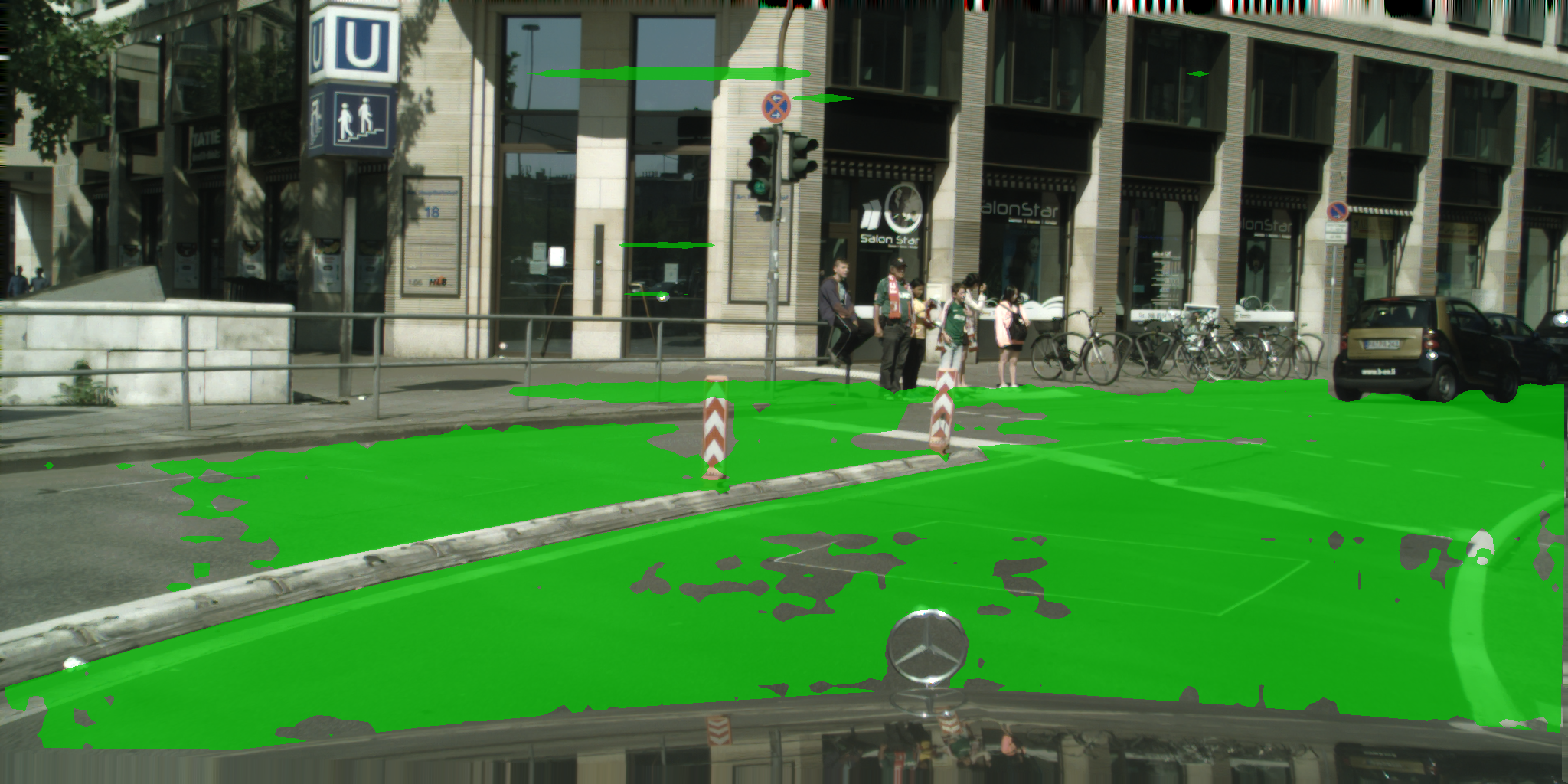}%
                \centering
                \\(c) \textbf{k-sBetas} (ours)
                \end{minipage}%
                \hspace{0.005\textwidth}
                \begin{minipage}{0.23\textwidth}%
                \includegraphics[width=\columnwidth]{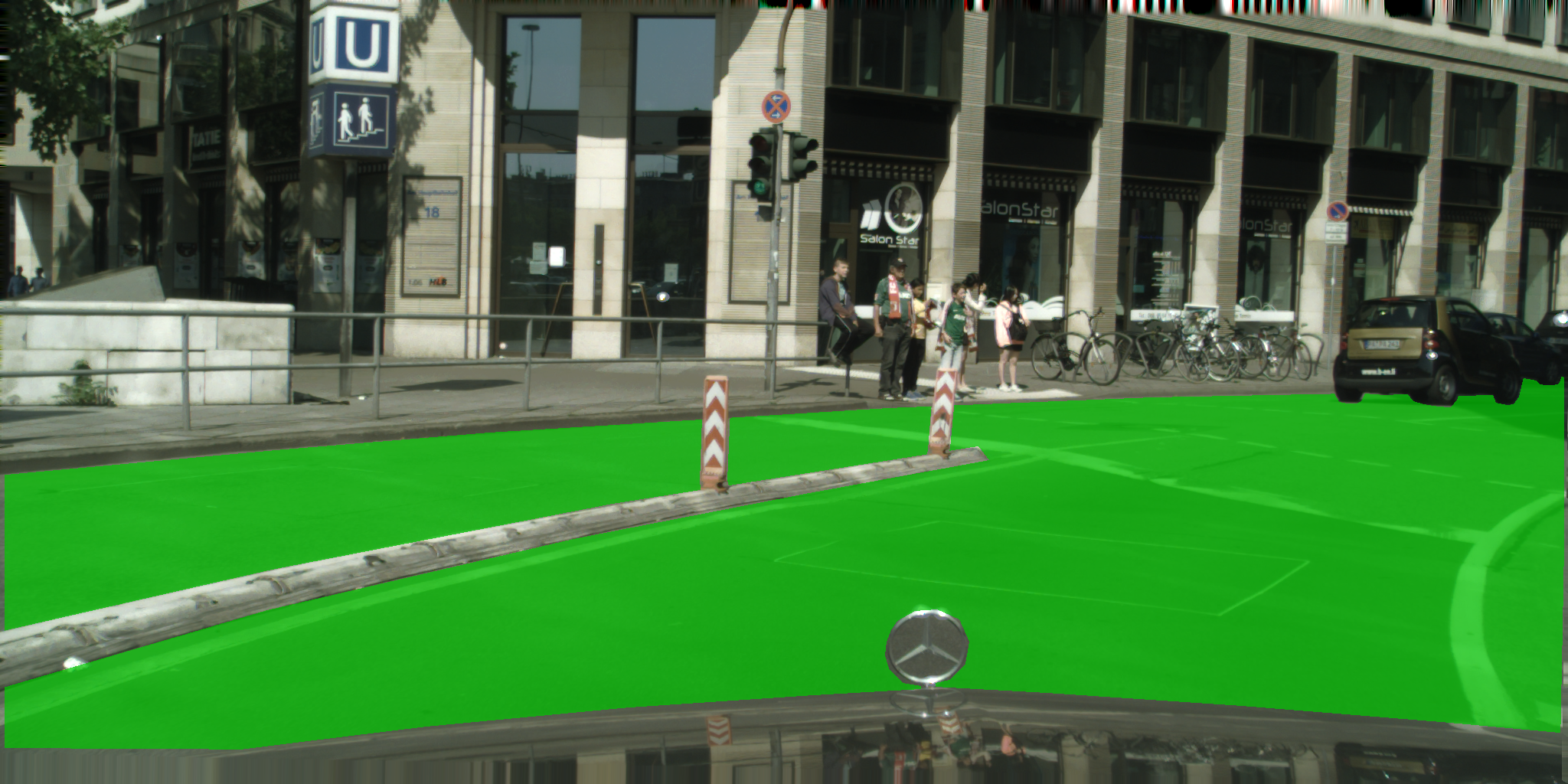}%
                \centering
                \\(d) Ground truth
                \end{minipage}%
                \vskip -0.07in
                \caption{\textbf{Real-time} (45 fps) black-box adaptation for road segmentation on images of size $2048 \times 1024$ by clustering the softmax predictions from a source model, pre-trained on GTA5 and applied to Cityscapes. See details in Sec. \ref{subsec_real_time_UDA_road_seg}.}%
                \vspace{-0.015\textwidth}
                \label{fig_GTA_to_Cityscapes_road_seg}%
                \vskip -0.1in
            \end{figure}

Clustering the softmax output predictions of deep networks can be formulated as a clustering problem on the probability simplex domain, i.e., clustering discrete distributions. Probability simplex clustering is not a new problem and has been the subject of several early studies in other application domains such as text analysis \cite{pereira1994distributional, banerjee2005clustering, wu2008sail, chaudhuri2008finding}. However, in the context of deep learning, clustering methods have been typically applied on deep feature maps, as in self-training frameworks \cite{caron2018deep}. This requires access to the internal feature maps of the source models, thereby violating the black-box assumption. In contrast, simplex-clustering-based adjustment of the probabilistic outputs of a deep pre-trained model complies with the black-box assumption but has remained largely under-explored, to the best of our knowledge.

\begin{figure*}%
\centering
\begin{minipage}{0.25\textwidth}%
\includegraphics[width=\columnwidth]{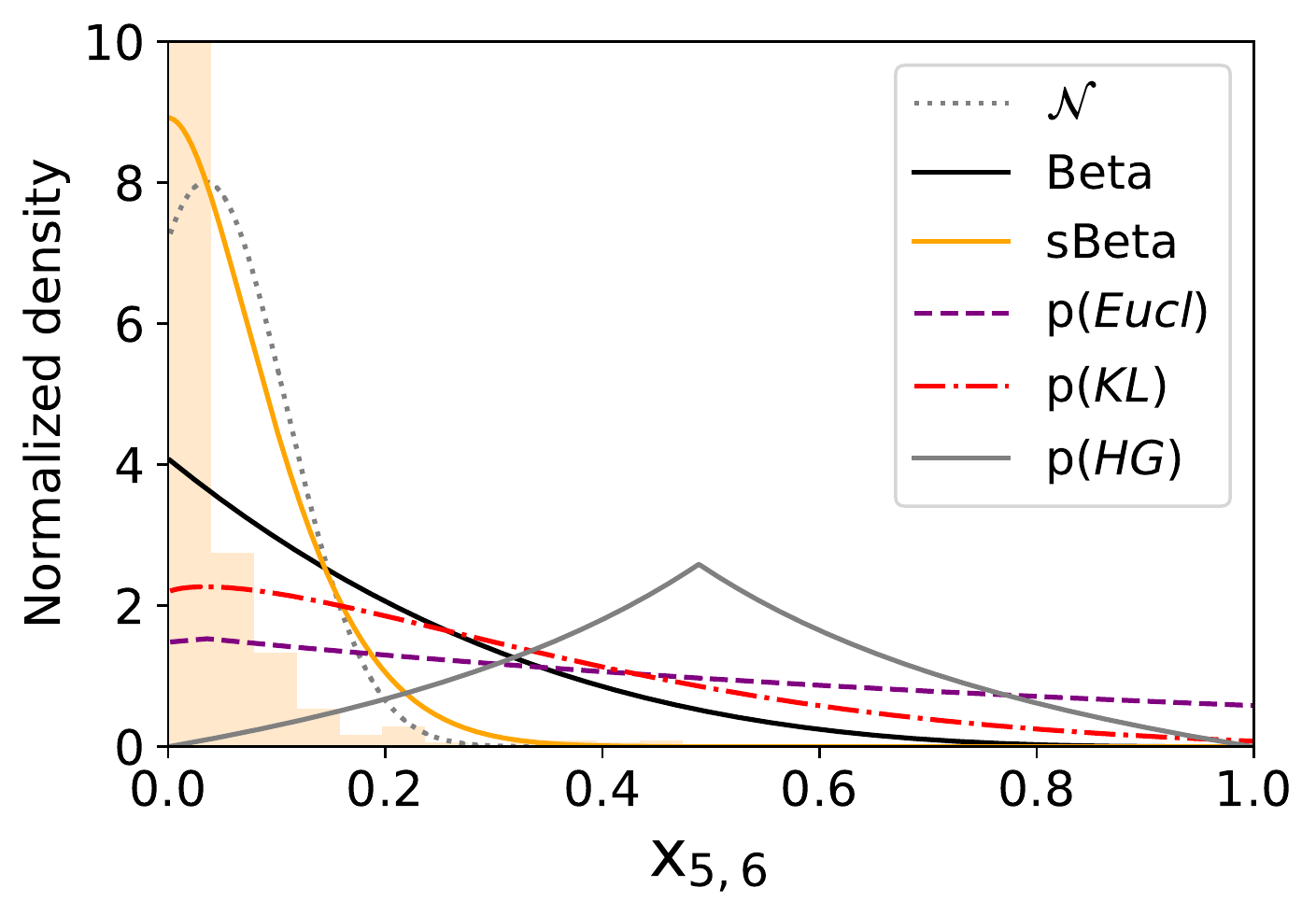}%
\centering
\\(a)
\end{minipage}%
\begin{minipage}{0.25\textwidth}%
\includegraphics[width=\columnwidth]{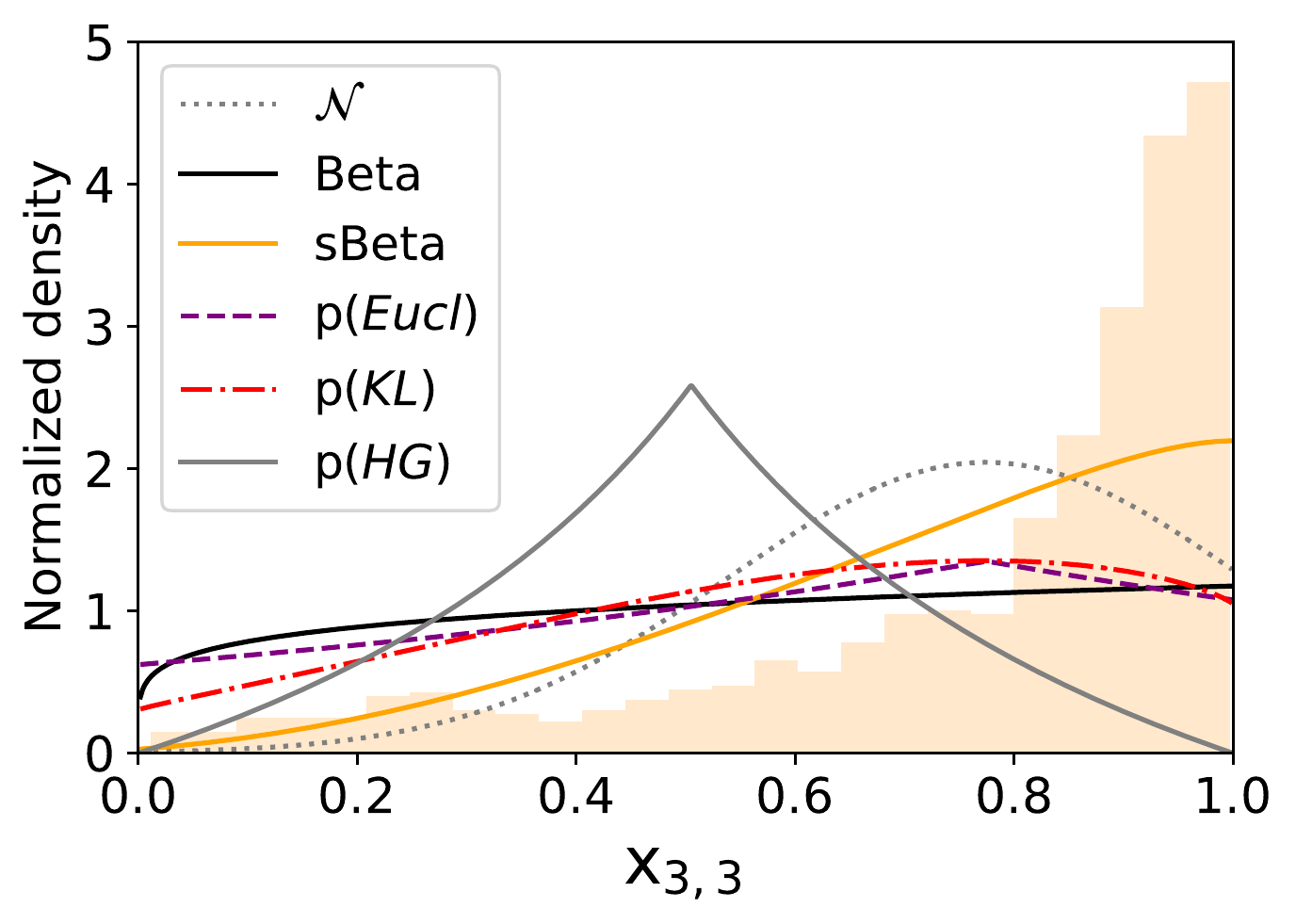}%
\centering
\\(b) 
\end{minipage}%
\begin{minipage}{0.25\textwidth}%
\includegraphics[width=\columnwidth]{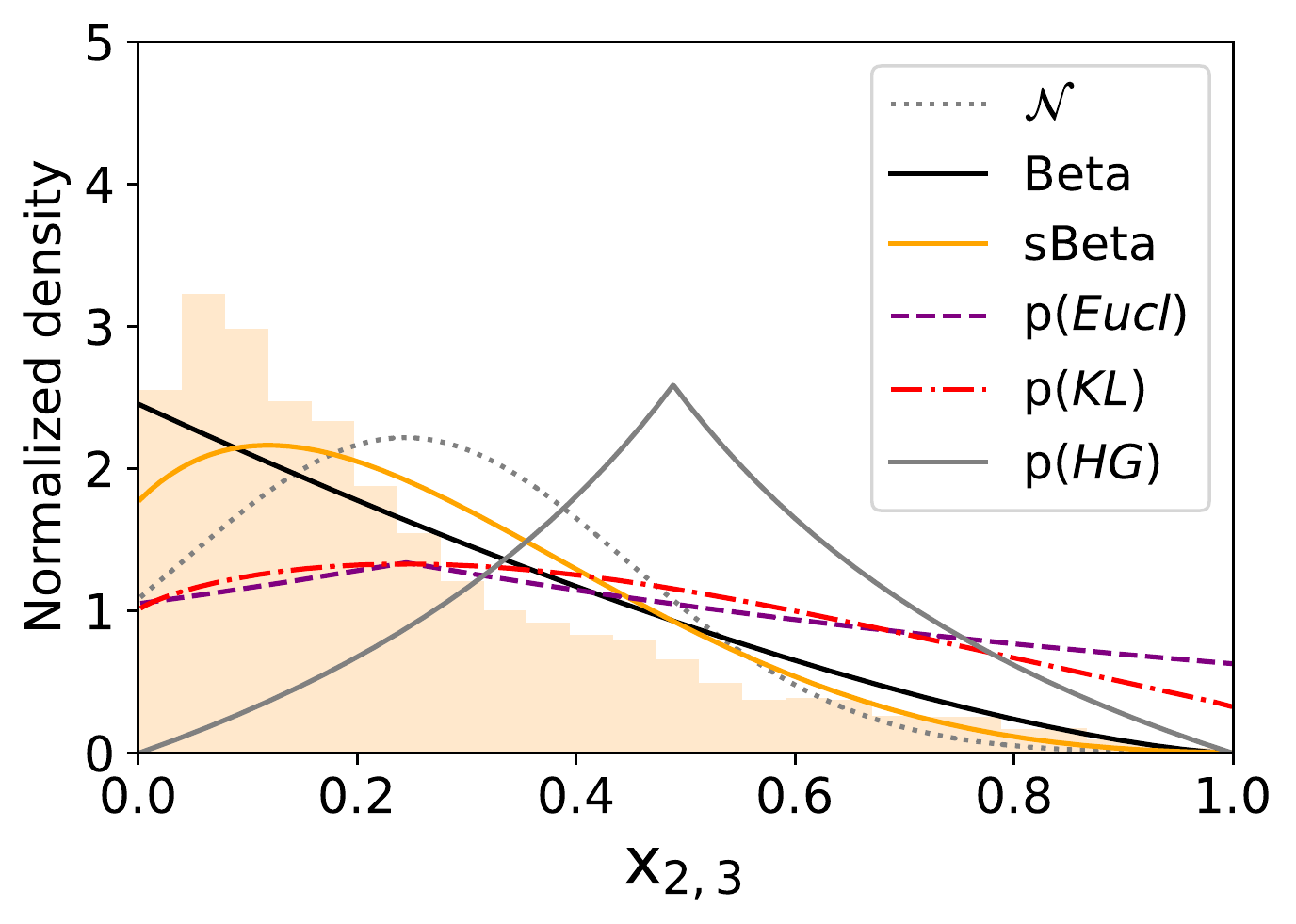}%
\centering
\\(c) 
\end{minipage}%
\begin{minipage}{0.25\textwidth}%
\includegraphics[width=\columnwidth]{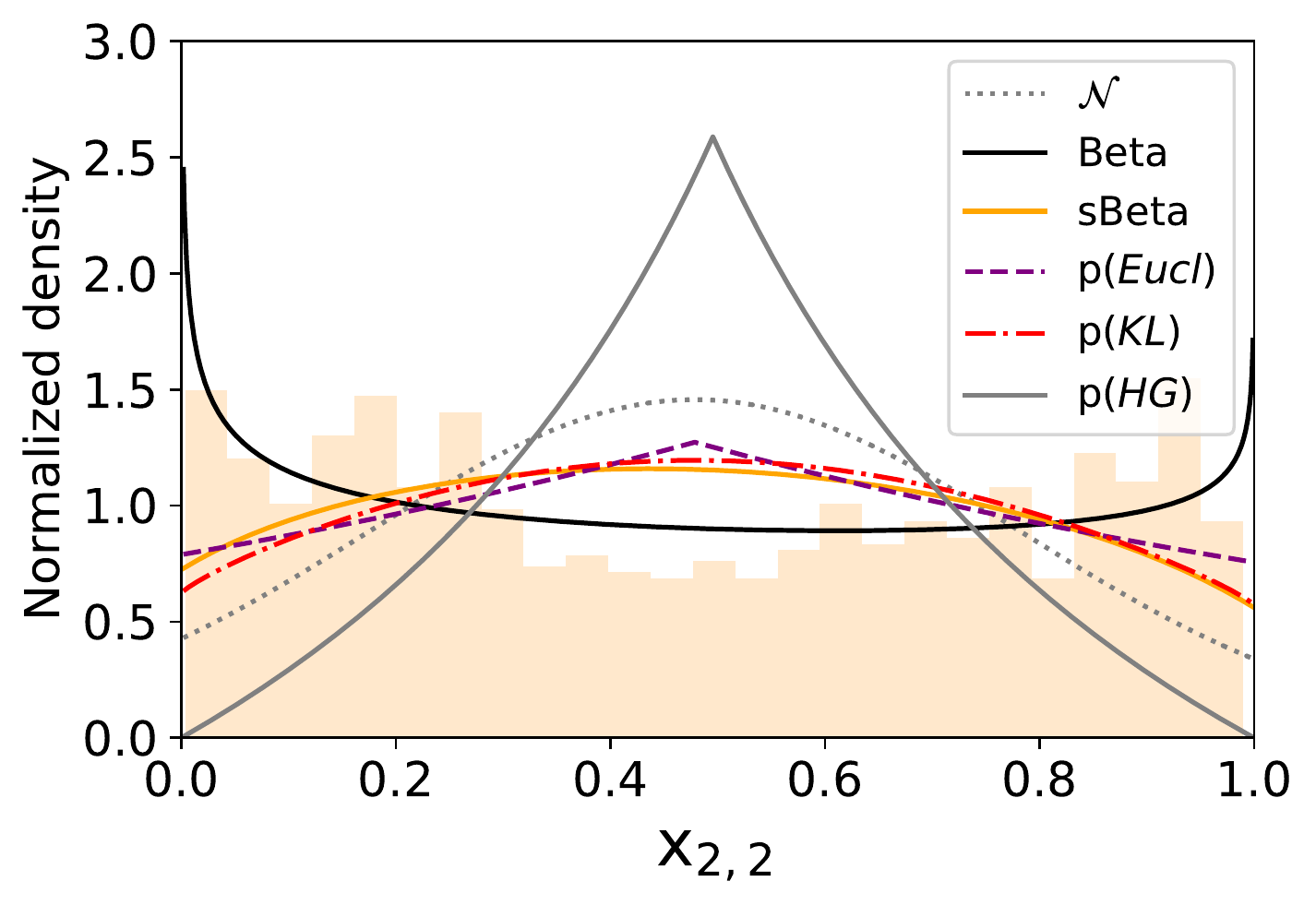}%
\centering
\\(d) 
\end{minipage}%
\caption{\textbf{Modelling softmax marginal distributions.} Figures (a), (b), (c) and (d) compare the density fittings of real-world marginal distributions of softmax predictions, represented by their respective histograms (orange bars). Specifically, the histograms in Fig. (a), (b), and (d) were extracted from the SVHN$\rightarrow$MNIST benchmark and the one in (c) from the VISDA-C one (details in Sec. \ref{subsec_softmax_preds_comp}). Figs. \ref{fig_S_to_M_hists} and \ref{fig_VISDA_C_hists} in the Appendix depict the whole set of histograms per class. Normalized probability density functions $\mathcal{N}$, $p(Eucl)$, $p(KL)$, $p(HG)$, $\mathtt{Beta}$ and $\mathtt{sBeta}$ are obtained, respectively, from the normal density, Euclidean distance, Kullback-Leibler divergence, Hilbert distance, $\mathtt{Beta}$, and $\mathtt{sBeta}$, all listed in Table. \ref{prob_to_metric_table}.}%
\label{fig_real_models_fitting}%
\end{figure*}
     
The simplex clustering literature is often based on optimizing {\em distortion-based} objectives. 
The goal is to minimize, within each cluster, some distortion measure evaluating the discrepancy between a cluster representative and each sample within the cluster. 
Besides standard objectives like \textsc{k-means}, which correspond to an $\|.\|_2$ distortion, several simplex-clustering methods motivated and used distortion measures that are specific to simplex data. This includes, for instance, the Kullback-Leibler (KL) divergence \cite{chaudhuri2008finding,wu2008sail} and the Hilbert geometry distance \cite{nielsen2019clustering}.
In this work, we explore a general maximum a posteriori (MAP) perspective of clustering discrete distributions. Using the Gibbs model, we emphasize that the density functions underlying current distortion-based objectives (Table \ref{prob_to_metric_table}) may not properly approximate the empirical marginal distributions of real-world softmax predictions.

Fig. \ref{fig_real_models_fitting} illustrates our empirical observations through histograms representing the marginal (per-coordinate) distributions of real softmax predictions. These predictions were obtained using a black-box source model applied to data from an unseen target domain. We juxtapose these empirical histograms to parametric density functions estimated on the same predictions via various clustering methods. For instance, curve $p(Eucl)$ represents the parametric Gibbs model corresponding to the Euclidean distance used in \textsc{k-means}. A close matching between the shape of the parametric density function and the one of the empirical histogram indicates a good fit of the simplex predictions. However, a significant disparity implies that the assumed parametric density function may not effectively model these predictions. For example, in Fig. \ref{fig_real_models_fitting} (c), the modes of parametric functions $\mathcal{N}$, $p(Eucl)$, $p(KL)$ and $p(HG)$ do not correspond to the one of the empirical distribution depicted with the orange histogram. Furthermore, Fig. \ref{fig_real_models_fitting} (d) depicts a particular scenario with an approximately uniform empirical distribution. In such cases, it is crucial for clustering models to generate unimodal density functions in order to preserve the assumption that the distribution of each class is unimodal. This avoids erroneously fitting multiple unimodal distributions to the data within a single cluster. As depicted in Fig. \ref{fig_real_models_fitting} (d), the standard $\mathtt{Beta}$ density may generate a bimodal distribution, thereby violating the unimodality assumption.

\textbf{Contributions.} Driven by the above observations, we introduce a novel probabilistic clustering objective integrating a generalization of the $\mathtt{Beta}$ density, which we coin $\mathtt{sBeta}$. We derive several properties of $\mathtt{sBeta}$, which enable us to impose different constraints on our clustering model, referred to as \textsc{k-sBetas}, enforcing 
uni-modality of the densities within each cluster while avoiding degenerate solutions. We proceed with a block-coordinate descent approach, alternating optimization w.r.t assignment variables, and inner Newton iterations for solving the non-linear optimality conditions w.r.t the sBeta parameters. 
Furthermore, using the density moments, we derive a closed-form alternative to Newton iterations for parameter estimation.
Our versatile formulation approximates various parametric densities for modeling simplex data, including highly peaked distributions at the vertices of the simplex, as observed empirically in the case of deep-network predictions. It also enables the control of the cluster balance. We report comprehensive experiments, comparisons, and ablation studies, which show highly competitive performances of our simplex clustering for unsupervised adjustment of black-box predictions in a variety of scenarios. Fig. \ref{fig_real_models_fitting} illustrates the ability of our method to consistently fit the empirical marginal distributions of real-world softmax predictions. For instance, Fig. \ref{fig_real_models_fitting} (c) shows that the density peak of $\mathtt{sBeta}$ (orange curve) matches the highest bar of the histogram, enabling a more accurate fitting than the other methods.
In addition, unlike the standard $\mathtt{Beta}$ density, our approach deliberately avoids fitting bimodal distributions (Fig. \ref{fig_real_models_fitting} (d)), thereby preventing degenerate solutions. This refinement not only ensures a more accurate representation of real data, but also maintains the assumption that the distribution of each class is unimodal.
To reproduce our comparative experiments, we have made the codes publicly available, including the proposed \textsc{k-sBetas} model, the explored state-of-the-art approaches, as well as the introduced softmax prediction benchmarks \footnote{\url{https://github.com/fchiaroni/Clustering_Softmax_Predictions}}. 

\section{Problem formulation} \label{sec_problem_formulation}

We start by introducing the basic notations, which will be used throughout the 
article:
\begin{itemize}
    \item $\Delta^{D-1}= \{\bm{x} \in [0,1]^D~|~ \bm{x}^T \bm{1} = 1\}$ stands for the ($D-1$)-probability simplex.
    \item $ \mathcal{X} = \{\bm{x}_i\}_{i=1}^N \iid X$, with $\mathcal{X}$ a given data set and $X$ denoting the random 
    simplex vector in $\Delta^{D-1}$.
    \item $\mathcal{X}_k = \{ \bm{x}_i \in \mathcal{X} ~| ~ u_{i,k}=1 \}$ denotes cluster $k$, where $\bm{u}_i = (u_{i,k})_{1 \leq k \leq K} \in \Delta^{K-1} \cup \{0, 1\}$ is a latent binary vector assigning point $\bm{x}_i$ to cluster $k$: $u_{i,k} = 1$ if $\bm{x}_i$ belongs to cluster $k$ and $u_{i,k}=0$ otherwise. Let $\mathbf U \in \{0, 1\}^{NK}$ denote the latent assignment matrix whose column vectors are given by $\bm{u}_i$. 
    Note that $\mathcal{X} = \mathcal{X}_1 \cup ... \cup \mathcal{X}_{|\mathcal{Y}|}$. 
\end{itemize}
Clustering consists of partitioning a given set $\mathcal{X}$ into $K$ different subsets $\mathcal{X}_k$ referred to as clusters. 
In general, this is done by optimizing some objective functions with respect to the latent assignment variables $\mathbf U$. Often, the 
basic assumption underlying objective functions for clustering is that data 
points belonging to the same cluster $k$ should be relatively close to each other, according to some pre-defined notion of closeness (e.g., via some distance or 
distortion measures). 

In this study, we tackle the clustering of probability simplex data points (or distributions), with a particular focus on the output predictions 
from deep-learning models. Softmax \footnote{Also referred to as the normalized exponential function, the softmax function is defined as ${\sigma(\bm{z})}_j = \frac{\exp(z_j)}{\sum_{l=1}^D \exp(z_l)}$, with $\bm{z}=\{z_1,...,z_{D}\}$ denoting a vector of logits.} is commonly used as the last activation 
function of neural-network classifiers to produce such output probability distributions. Each resulting softmax data point $\bm{x}$ is a $D$-dimensional probability vector of $D$ continuous random variables, which are in $[0, 1]$ and sum to one. Thus, the target clustering challenge is defined on probability simplex domain $\Delta^{D-1}$. Fig. \ref{fig_motiv_PSC_and_Hung} depicts mixtures of three class distributions on $\Delta^{2}$, simulating the impact of a domain shift on the softmax predictions, as well as the benefits of simplex clustering over the standard argmax prediction in such a scenario. Specifically, Fig. \ref{fig_motiv_PSC_and_Hung} (a) shows the softmax predictions from a model trained on and applied to the same source domain. In this case, the standard argmax function, commonly used in deep networks, effectively separates the three classes. In contrast, Fig. \ref{fig_motiv_PSC_and_Hung} (b) reveals that, in the case of a domain shift, the argmax function may lead to erroneous predictions. Fig. \ref{fig_motiv_PSC_and_Hung} (c) illustrates how a simplex clustering successfully maintains class cohesion in this case. This synthetic example points to the potential of simplex clustering in dealing with domain-shift challenges while operating solely on the outputs (i.e., in a black-box setting).

\begin{figure}%
\begin{minipage}{0.05\columnwidth}%
\textcolor{white}{ }
\end{minipage}%
\begin{minipage}{0.05\columnwidth}%
\rotatebox{90}{ NO SHIFT}
\end{minipage}%
\begin{minipage}{0.225\columnwidth}%
\textcolor{white}{ }
\end{minipage}%
\begin{minipage}{0.4\columnwidth}%
\includegraphics[width=\columnwidth]{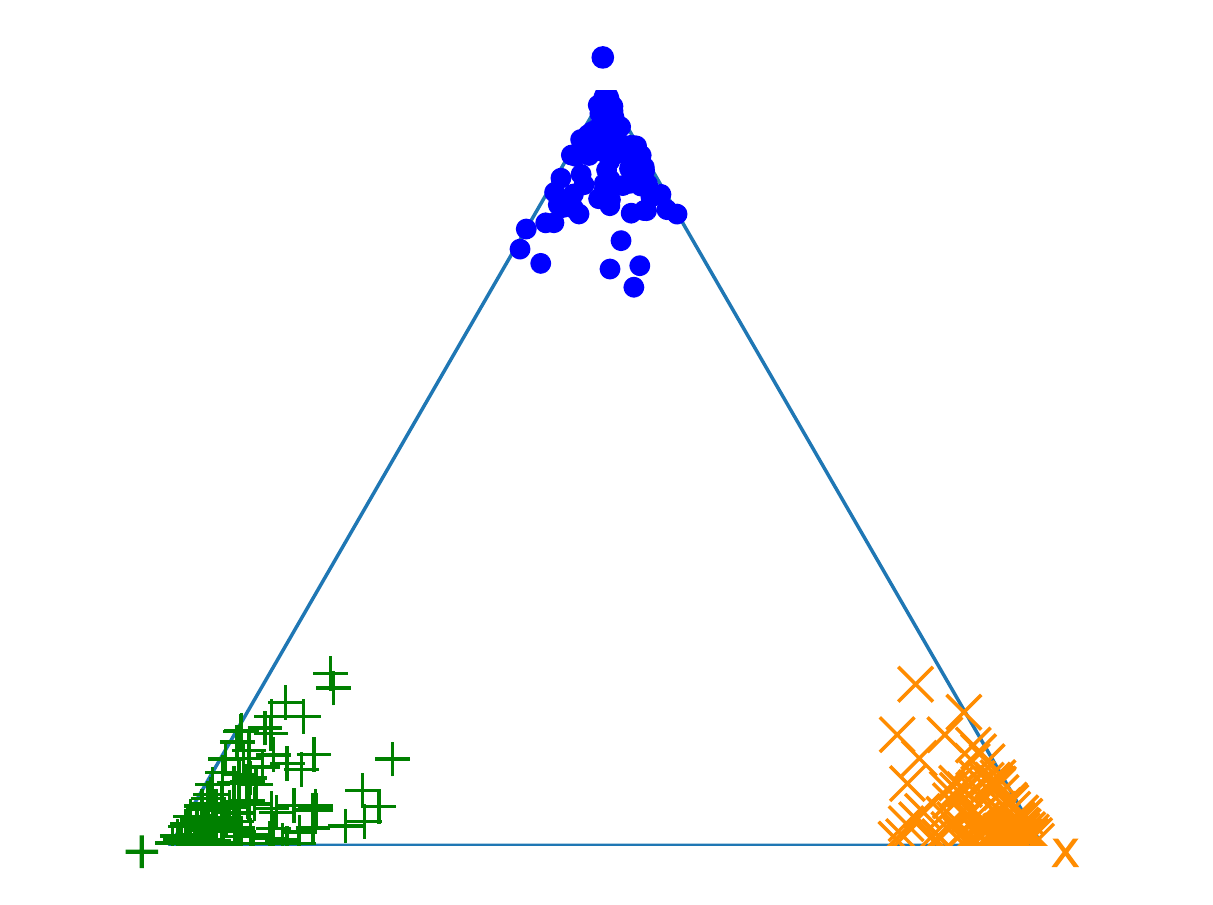}%
\centering
\\(a) Argmax or PSC
\end{minipage}%
\\
\begin{minipage}{0.05\columnwidth}%
\textcolor{white}{ }
\end{minipage}%
\begin{minipage}{0.05\columnwidth}%
\rotatebox{90}{SHIFT}
\end{minipage}%
\begin{minipage}{0.43\columnwidth}%
\includegraphics[width=\columnwidth]{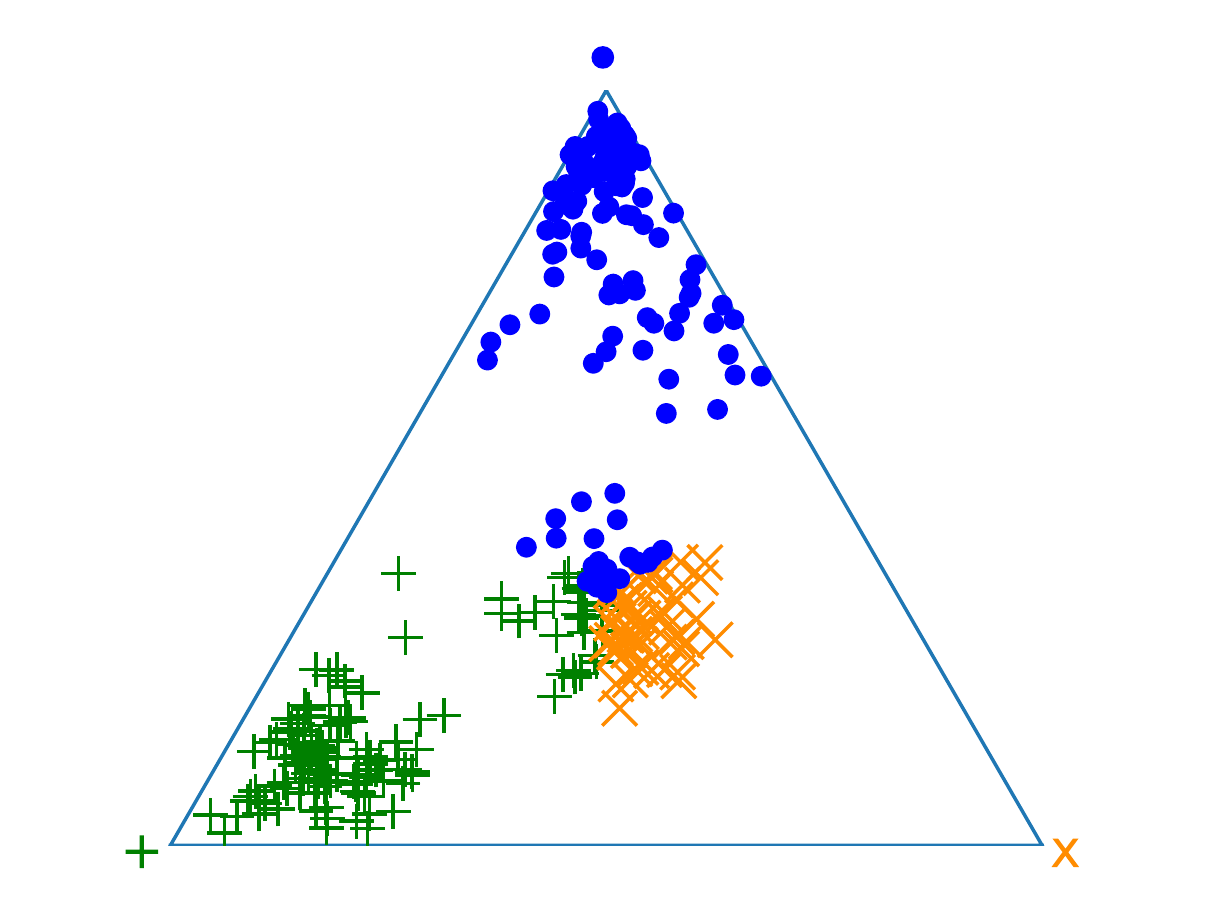}%
\centering
\\(b) Argmax
\end{minipage}%
\begin{minipage}{0.43\columnwidth}%
\includegraphics[width=\columnwidth]{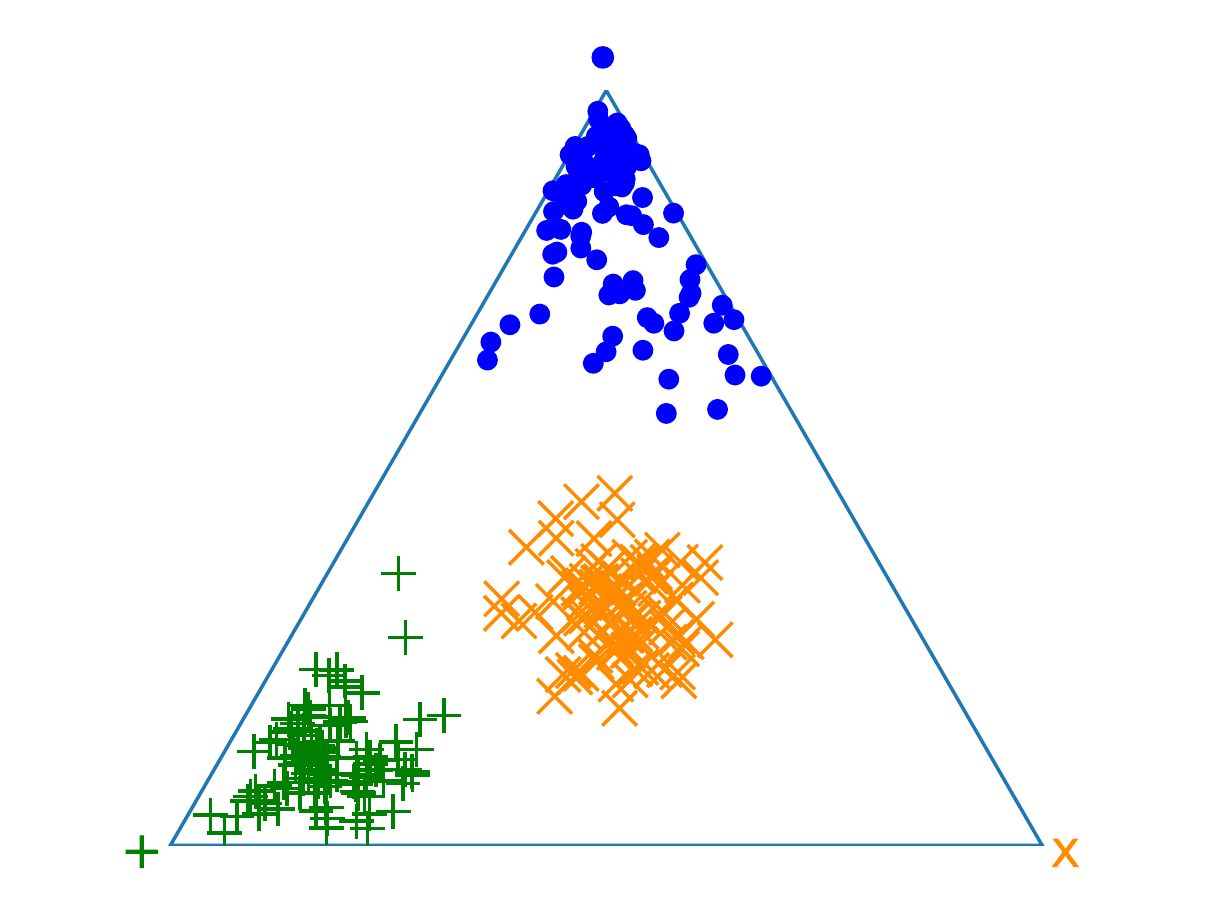}%
\centering
\\(c) PSC
\end{minipage}%
\caption{\textbf{Partitioning points within probability simplex $\Delta^{2}$}. Specifically, these points are projected into a 2D equilateral triangle for a clear visualization, where each vertex represents one dimension of the probability space. The figures depict the probabilistic predictions from a hypothetical source model, with and without distribution shift. Fig. (b) depicts the standard argmax assignment, whereas Fig. (c) represents a Probability Simplex Clustering (PSC) using a cluster-to-class alignment. Symbols \textbf{$+$} (green), x (orange) and $\bullet$ (blue) indicate separate class assignments. 
}%
\label{fig_motiv_PSC_and_Hung}%
\end{figure}

\section{Related work}
\label{sec_Related_Work}
This section presents related work in the general context of clustering, focusing on the probability simplex domain.
\subsection{Distortion-based clustering objectives}
The most widely used form of clustering objectives is based on some {\em distortion} measures, for instance, a distance in general-purpose clustering or a divergence measure in the case of probability simplex data. 
This amounts to minimizing w.r.t both assignment variables and cluster representatives $\bm{\Theta} = (\bm{\theta}_{k})_{1\leq k \leq K}$ a mixed-variable function of 
the following general form:
\begin{equation}
\label{general_distortion_objective}
    L_{dist}({\mathbf U};\bm{\Theta}) = \sum_{k=1}^K \sum_{i=1}^N  u_{i,k} \|\bm{x}_i- \bm{\theta}_{k}\|_d = \sum_{k=1}^K \sum_{\bm{x}_i \in \mathcal{X}_k} \|\bm{x}_i- \bm{\theta}_{k}\|_d. 
\end{equation}
The goal is to minimize, within each cluster $\mathcal{X}_k$, some distortion measure $\|.\|_d$ evaluating the discrepancy between cluster representative $\bm{\theta}_{k}$ and each point 
belonging to the cluster. When $\|.\|_d$ is the Euclidean distance, general form \eqref{general_distortion_objective} becomes the standard and widely used \textsc{k-means} objective \cite{DudaHart2nd}.
In general, optimizing distortion-based objective \eqref{general_distortion_objective} is NP-hard\footnote{For example, a proof of the NP-hardness of the standard \textsc{k-means} objective could be found in \cite{Aloise-2009}.}. One standard iterative solution to tackle the problem in \eqref{general_distortion_objective} is to follow a block-coordinate descent approach, which alternates two steps, one optimizes the objective w.r.t to the cluster prototypes and the other w.r.t assignment variables:
\begin{itemize}
    \item {\bf U}-update:  
       $     u_{i,k} = \left\{
                \begin{array}{ll}
                    1 & \mbox{if } \argminC_{k} \|\bm{x}_i- \bm{\theta}_{k}\|_d = k \\
                    0 & \mbox{otherwise.}
                \end{array}
            \right.$
    \item $\bm{\Theta}$-update: Find $\argmin_{\bm{\Theta}}  L_{dist}({\mathbf U};\bm{\Theta})$
\end{itemize}
In the specific case of \textsc{k-means} (i.e., using the $\|.\|_2$ distance as distortion measure), optimization w.r.t the parameters in the $\bm{\Theta}$-update 
step yields closed-form solutions, which correspond to the {\em means} of features within the clusters: 
\[\bm{\theta}_{k} = \frac{\sum_{i=1}^N  u_{i,k} \bm{x}_i} {  \sum_{i=1}^N  u_{i,k}}, \forall k. \]

One way to generalize \textsc{k-means} is to replace the Euclidean distance with other distortion measures. In this case, the optimal value of the parameters
$\bm{\theta}_{k}$ may no longer correspond to cluster means. For instance, when using the $\|.\|_1$ distance as distortion measure 
in \eqref{general_distortion_objective}, the optimal $\bm{\theta}_{k}$ corresponds to the cluster median, and the ensuing algorithm is the 
well-known \textsc{k-median} clustering \cite{bradley1997clustering}. For exponential distortions, the optimal parameters may correspond to cluster modes \cite{carreira2013k, Ziko2018}. 
Such exponential distortions enable the model to be more robust to outliers, but this comes at the price of additional computations (inner iterations), as cluster modes could not be obtained in closed-form. 

\subsection{Distortion measures for simplex data} 

In our case, data points are probability vectors within simplex domain $\Delta^{D-1}$, e.g. the softmax predictions of deep networks. These points are $D$-dimensional vectors of $D$ continuous random variables bounded between $0$ and $1$, and summing to one. To our knowledge, the simplex clustering literature is often based on distortion objectives of the general form in \eqref{general_distortion_objective}. Besides standard objectives like \textsc{k-means}, several simplex-clustering works motivated and used distortion measures that are specific to simplex data. This includes the Kullback-Leibler (KL) divergence \cite{chaudhuri2008finding,wu2008sail} and Hilbert geometry distance \cite{nielsen2019clustering}.   

\textbf{Information-theoretic k-means.} The works in \cite{chaudhuri2008finding, wu2008sail} discussed \textbf{\textsc{KL k-means}}, a distortion-based clustering tailored to simplex data and whose objective fits the general form in \eqref{general_distortion_objective}. \textsc{KL k-means} uses the Kullback-Leibler (KL) divergence as a distortion measure 
instead of the Euclidean distance in the standard \textsc{k-means}:
\begin{align}
\|\bm{x}_i- \bm{\theta}_{k}\|_d = \mbox{KL}(\bm{x}_i||\bm{\theta}_{k})= \sum_{n=1}^D x_{i,n} \log \left ( \frac{x_{i,n}}{\theta_{k,n}} \right ),
\end{align}
where $x_{i,n}$ and $\theta_{k,n}$ stand, respectively, for the $n$-th component of simplex vectors $\bm{x}_i$ and $\bm{\theta}_{k}$.   
$\mbox{KL}(\bm{x}_i||\bm{\theta}_{k})$ is a Bregman divergence \cite{banerjee2005clustering} measuring the dissimilarity between two distributions $\bm{x}_i$ and $\bm{\theta}_{k}$. Despite being asymmetric \cite{chaudhuri2008finding}, the machine learning community widely uses the KL 
divergence in a breadth of problems where one has to deal with probability simplex vectors \cite{krause2010discriminative, rezende2014stochastic, hu2017learning, claici2020model}.

\textbf{Hilbert Simplex Clustering (HSC).} More recently, the study in \cite{nielsen2019clustering} investigated the Hilbert Geometry (HG) distortion with \textit{minimum enclosing ball (MEB)} centroids \cite{gonzalez1985clustering,panigrahy1998ano} for clustering probability simplex vectors. For a given cluster set $k$, the MEB center represents the midpoint of the two farthest points within the set.
Given two points $\bm{x}_i$ and $\bm{\theta}_k$ within simplex domain $\Delta^{D-1}$, let $\bm{x}_i + t (\bm{\theta}_k-\bm{x}_i)$, $t \in {\mathbb R}$, denote the line passing through points $\bm{x}_i$ and $\bm{\theta}_k$, with $t_0 \leq 0$ and $t_1 \geq 1$ the two intersection points of this line with the simplex domain boundary $\partial \Delta^{D-1}$. Then, HG simplex distortion is given by: 
\begin{equation}
    \begin{aligned}
        \text{HG}_{\Delta^{D-1}} (\bm{x}_i;\bm{\theta}_k) &= 
        |\log(\frac{(1-t_0)t_1}{(-t_0)(t_1-1)})| \\
        &= \log(1-\frac{1}{t_0}) - \log(1-\frac{1}{t_1}).
    \end{aligned}
\label{eq_HG_simplex}
\end{equation}
Note that if one deals with centroids close or equal to the vertices of the simplex, 
then HG distortion would inconsistently output large or infinite distance values near such centroids.

\subsection{From distortions to probabilistic clustering}
\label{subsec_prob_clust_form}

Beyond distortions, a more versatile approach is to minimize the following well-known maximum a posteriori (MAP) generalization 
of \textsc{K-means} \cite{boykov_volumetric_2015,tang2019kernel,kearns1998information}, coined by \cite{kearns1998information} as 
{\em probabilistic} \textsc{K-means}: 
\begin{equation}\label{eq_prob_formulation}
    L_{prob}({\mathbf U};\bm{\pi};\bm{\Theta})= - \sum_{k=1}^K \sum_{i=1}^N u_{i,k} \log( \pi_k f(\bm{x}_i;\bm{\theta}_{k})),
\end{equation}
where $f(\cdot;\bm{\theta}_{k})$ is a parametric density function modeling the likelihood probabilities of the samples within cluster $k$, and $\pi_k$ the prior probability of cluster $k$, with $\bm{\pi}=(\pi_k)_{1\leq k \leq K} \in \Delta^{K-1}$. Hence, instead of minimizing a distortion $\|\bm{x}_i- \bm{\theta}_{k}\|_d$ in \eqref{general_distortion_objective}, we maximize a more general quantity measuring the posterior probability of cluster $k$ given sample $\bm{x}_i$ and parameters $\bm{\theta}_{k}$. It is easy to see that, for any choice of distortion measure, the objective
in \eqref{general_distortion_objective} corresponds, up to a constant, to a particular case of probabilistic \textsc{K-means} in \eqref{eq_prob_formulation}, with $\pi_k = 1/K, \forall k$, and parametric density $f$ chosen to be the {\em Gibbs} distribution:
\begin{equation}
\label{Gibbs-distribution}
f(\bm{x}_i;\bm{\theta}_{k}) \propto \exp(-\|\bm{x}_i- \bm{\theta}_{k}\|_d).
\end{equation}
For instance, the \textsc{K-means} objective corresponds, up to additive and multiplicative constants, to choosing likelihood probability $f(\bm{x}_i;\bm{\theta}_{k})$ to be a Gaussian 
density with mean equal to $\bm{\theta}_{k}$, identity covariance matrix and prior probabilities verifying $\pi_k = 1/K, \forall k$. This uncovers hidden assumptions in \textsc{K-means}: 
The samples are assumed to follow Gaussian distribution within each cluster, and the clusters are assumed to 
be balanced\footnote{In fact, it is well-known that \textsc{K-means} has a strong bias towards balanced partitions \cite{boykov_volumetric_2015}}. 
Therefore, the general probabilistic \textsc{K-means} formulation in \eqref{eq_prob_formulation} is more versatile as it enables to make more appropriate assumptions about the parametric statistical models of the samples within the clusters (e.g., via prior knowledge or validation data), and to manage the cluster-balance bias \cite{boykov_volumetric_2015}.

\subsection{Do current simplex clustering methods model softmax predictions properly?}
\label{sec_Related_Work_limits}

\begin{table}[t]
    \caption{Comparison of metrics and density functions, depending on the probability simplex vector $\bm{x}_i=( x_{i,n} )_{1 \leq n \leq D} \in \Delta^{D-1}$ 
    and the parameters $\bm{\theta}_k=(\theta_{k,n})_{1 \leq n \leq D}$ for Euclidean, KL, and $HG_{\Delta^{D-1}}$ distortions. $\bm{\theta}_k$ respectively represents the density parameters $\bm{\theta}_k=(\bm{\mu}_k, \Sigma_k)$ with $\bm{\mu}_k=(\mu_{k,n})_{1 \leq n \leq D}$ for $\mathcal{N}$, $\bm{\theta}_k=(\alpha_{k,n})_{1 \leq n \leq D}$ for $f_{\mathtt{Dir}}$, and $\bm{\theta}_k=(\alpha_{k,n}, \beta_{k,n})_{1 \leq n \leq D}$ for $g_{\mathtt{Beta}}$ and $g_{\mathtt{sBeta}}$.}
    \label{prob_to_metric_table}
    \vskip 0.15in
    \begin{center}
    \begin{small}
    \begin{sc}
    \resizebox{\columnwidth}{!}{%
        \begin{tabular}{lcc}
            Metric-Based & $\displaystyle \exp\left(-\textrm{metric}(\bm{x}_i;\bm{\theta}_k)\right)$ \\
            \toprule
            $\| \bm{x}_i - \bm{\theta}_k \|_2$  & $\exp\left(-\sqrt{\sum_{n=1}^D(x_{i,n}-\theta_{k,n})^2}\right)$ \\ 
            \midrule
            $\mbox{KL}(\bm{x}_i||\bm{\theta}_k)$ & $\displaystyle \prod_{n=1}^D \exp\left(- x_{i,n} \cdot \log \frac{x_{i,n}}{\theta_{k,n}}\right)$ \\
            \midrule
            $\mbox{HG}_{\Delta^{D-1}} (\bm{x}_i;\bm{\theta}_k)$ (Eq. \eqref{eq_HG_simplex}) & $(1-\displaystyle \frac{1}{t_1}) / (1-\displaystyle \frac{1}{t_0})$ \\
            \bottomrule
        \end{tabular}
    }
    \vskip 0.3in
    \resizebox{\columnwidth}{!}{%
        \begin{tabular}{lcc}
            Probabilistic & $f(\bm{x}_i; \bm{\theta}_k)$ \\
            \midrule
            $\mathcal{N}(\bm{x}_i; \bm{\theta}_k)$ & $\frac{1}{\sqrt{(2\pi)^n|\Sigma_k|}}exp\left(-\frac{1}{2}(\bm{x}_i-\bm{\mu}_k)^T \Sigma_k^{-1}(\bm{x}_i-\bm{\mu}_k)\right)$ \\
            \midrule
            $f_{\mathtt{Dir}}(\bm{x}_i; \bm{\theta}_k)$ & $\displaystyle \frac{\prod_{n=1}^D {x_{i,n}}^{(\alpha_{k,n}-1)}}{B(\bm{\theta}_k)}$ \\
            \midrule
            $g_{\mathtt{Beta}} (\bm{x}_i; \bm{\theta}_k)$ & $\displaystyle \prod_{n=1}^D \frac{(x_{i,n})^{\alpha_{k,n}-1}(1-x_{i,n})^{\beta_{k,n}-1}}{B(\alpha_{k,n},\beta_{k,n})}$ \\
            \midrule
            $g_{\mathtt{sBeta}} (\bm{x}_i; \bm{\theta}_k)$
             & $\displaystyle \prod_{n=1}^D \frac{(x_{i,n}+\delta)^{\alpha_{k,n}-1}(1+\delta-x_{i,n})^{\beta_{k,n}-1}}{B(\alpha_{k,n},\beta_{k,n})(1+2\delta)^{\alpha_{k,n}+\beta_{k,n}-2}}$ \\
            \bottomrule
        \end{tabular}
    }
    \end{sc}
    \end{small}
    \end{center}
    \vskip -0.1in
\end{table}

Eq. \eqref{Gibbs-distribution} emphasized that there are implicit density functions underlying distortion-based clustering objectives via Gibbs models; see Table \ref{prob_to_metric_table}. Fig. \ref{fig_real_models_fitting} 
illustrates how the Gibbs models corresponding to the existing simplex clustering algorithms may not properly approximate the empirical marginal density functions of real-world softmax predictions:

\textbf{Standard \textsc{k-means}: Fast, but not descriptive enough and biased.}
The well-known \textsc{k-means} clustering falls under the probabilistic clustering framework, as defined in Sec. \ref{subsec_prob_clust_form}. In particular, it instantiates the data likelihoods 
as Gaussian densities with unit covariance matrices. An important advantage of \textsc{k-means} is that the associated ${\mathbf \Theta}$-update step can be analytically solved by simply updating each $\theta_k$ as the
mean of samples within cluster $k$, according to the current latent assignments. Such closed-form solutions enable \textsc{k-means} to be among the most efficient clustering algorithms. However, the Gaussian 
assumption yielding such closed-form solutions might be limiting when dealing with asymmetric data densities, such as exponential densities, with the particularity that the mode and the mean are distinct. 
As depicted by Fig. \ref{fig_real_models_fitting} (a) and (b), the softmax outputs from actual deep-learning models commonly follow such asymmetric distributions.
Note that \textsc{k-means} and other distortion-based objectives correspond to setting $\pi_k = 1/K, \forall k$ in the general probabilistic formulation in Eq. \eqref{eq_prob_formulation}. Therefore, they encode 
the implicit strong bias towards balanced partitions \cite{boykov_volumetric_2015}, and might be sub-optimal when dealing with imbalanced clusters, as is often the case of realistic deep-network predictions.

\textbf{Elliptic \textsc{k-means}.} Elliptic \textsc{k-means} is a generalization of the standard K-means: It uses the Mahalanobis distance instead of the Euclidean distance, which assumes the 
data within each cluster follows a Gaussian distribution (i.e. Gibbs model for the Mahalanobis distance) whose covariance matrix is also a variable of the model (as opposed to unit covariance in  \textsc{k-means}); see Table 
\ref{prob_to_metric_table}. Fig. \ref{fig_real_models_fitting} (b) and (c) confirm that Gaussian density function $\mathcal{N}$ is not appropriate for skewed distributions, as is the case of network predictions.

\textbf{HSC yields poor approximations of highly peaked distributions located at the vertices of the simplex.} Fig. \ref{fig_real_models_fitting}
shows that HSC, which combines MEB centroids and HG metric, is not relevant to asymmetric distributions whose modes are close to the vertices of the simplex. Specifically, in Fig. \ref{fig_real_models_fitting} (a) and (b), the corresponding Gibbs probability is small near the vertices of the simplex and high in the middle of the simplex. This yields a poor approximation of the target empirical histograms (depicted in orange). In addition, the overall HSC algorithm is computationally demanding (see Table. \ref{table_comp_time}).

\textbf{KL \textsc{k-means}.} 
While KL could yield asymmetric density, it may poorly model highly peaked distributions at the vertices of the simplex (e.g. Fig. \ref{fig_real_models_fitting} a), as is the case of softmax predictions.

\section{Proposed approach}
\label{sec_method}

\subsection{Background}
\label{sec_Background}

\textbf{Multivariate Dirichlet density function:} Commonly used 
for modeling categorical events, the Dirichlet density operates on $D$-dimensional discrete distributions within the simplex, $\Delta^{D-1}$. Its expression, parameterized by vector $\bm{\alpha}=(\alpha_{n})_{1 \leq n \leq D} \in {\mathbb R}^D$, is given by:
\begin{align}
    f_{\mathtt{Dir}}(\bm{x};\bm{\alpha}) = \frac{1}{B(\bm{\alpha})} \prod_{n=1}^D x_n^{\alpha_{n} - 1},
\end{align}
with $\bm{x} \in \Delta^{D-1}$, and $B(\bm{\alpha})$ the multivariate Beta function expressed using the Gamma function\footnote{The Gamma function $\Gamma(\alpha) = \int_{0}^{\infty} t^{\alpha-1} \exp(-t) dt$ for $\alpha > 0$. $\Gamma(\alpha) = (\alpha-1)!$ when $\alpha$ is a strictly positive integer.}. While the Dirichlet density is designed for simplex vectors, its complex nature leads to computationally intensive optimization, often involving iterative and potentially sub-optimal parameter estimation \cite{minka2000estimating}. To address this computational challenge, one could adopt the mean-field approximation principle \cite{wainwright2008graphical}. This uses a relaxed product density, which is 
based on Dirichlet's  marginal density, i.e. $\mathtt{Beta}$ with parameters $\bm{\alpha}$ and $\bm{\beta}$:
\begin{equation}
\label{beta-density-mean-field}
    \begin{aligned}
        g_{\mathtt{Beta}} (\bm{x};\bm{\alpha},\bm{\beta})
        &= \prod_{n=1}^D {f_{\mathtt{Beta}}}(x_n;\alpha_n,\beta_n) \\
        &= \prod_{n=1}^D \frac{(x_n)^{\alpha_n-1}(1-x_n)^{\beta_n-1}}{B(\alpha_n,\beta_n)} \\
        &\approx {f_{\mathtt{Dir}}}(\bm{x};\bm{\alpha}).
    \end{aligned}
\end{equation}
This approximation simplifies parameter estimation by treating each simplex coordinate independently. However, apart from the computational load, $\mathtt{Dir}$ and its marginal $\mathtt{Beta}$ may yield poor approximations of the
empirical marginal distributions of real-world softmax predictions. For example, Fig. \ref{fig_real_models_fitting} (a), (b), and (c) show how $\mathtt{Beta}$ has difficulty capturing empirical marginal distributions. In addition, 
$\mathtt{Dir}$ allows multimodal distributions, as shown with its $\mathtt{Beta}$ marginal in Fig. \ref{fig_real_models_fitting} (d), opposing the common assumption that each semantic class distribution is unimodal. Indeed, discriminative deep learning models are optimized to output probability simplex vertices (i.e. one-hot vectors). For each class, the model is optimized to output softmax predictions following an unimodal distribution. In this context, $\mathtt{Dir}$ may erroneously represent two unimodal distributions as a single bimodal distribution.

In the next section, we propose $\mathtt{sBeta}$, an alternative capable of effectively addressing these limitations.

\subsection{Proposed density function: A generalization of Beta constrained to be unimodal}

\begin{figure}%
\centering
\begin{minipage}{0.5\columnwidth}%
\includegraphics[width=\columnwidth]{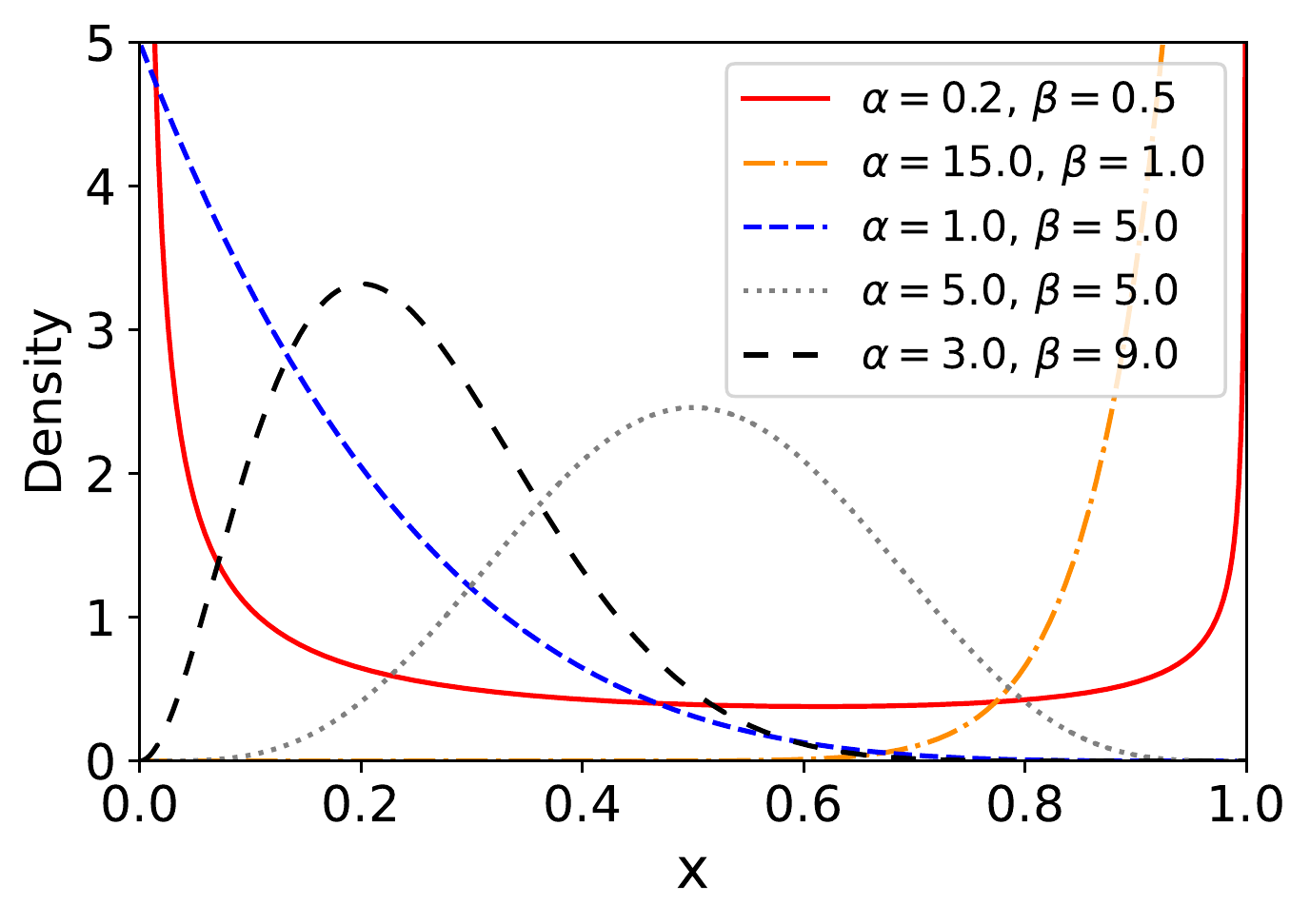}%
\centering
\\(a) $\mathtt{Beta}$
\end{minipage}%
\begin{minipage}{0.5\columnwidth}%
\includegraphics[width=\columnwidth]{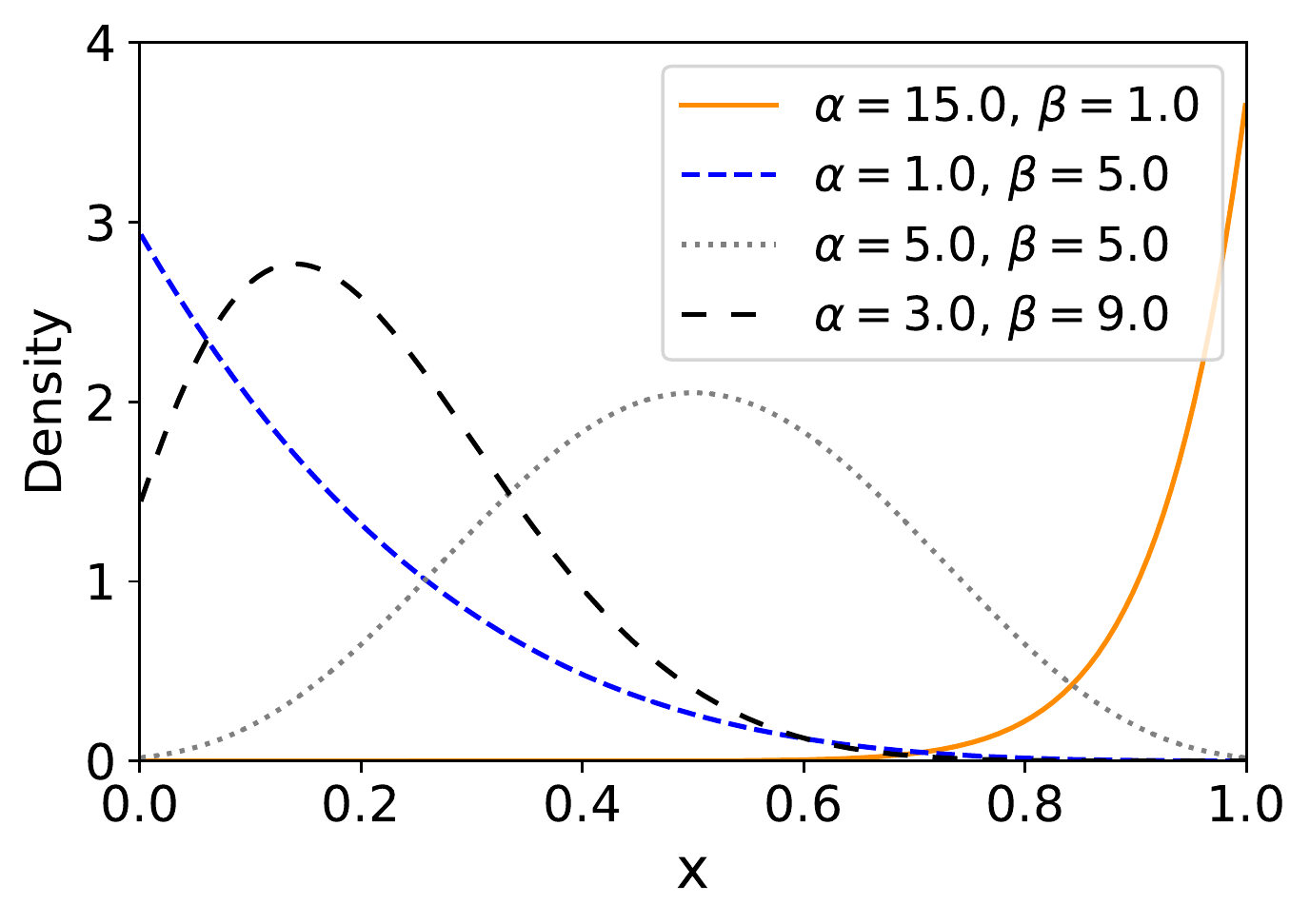}%
\centering
\\(b) $\mathtt{sBeta}$
\end{minipage}%
\vspace{0.015\textwidth}
\begin{minipage}{0.5\columnwidth}%
\includegraphics[width=\columnwidth]{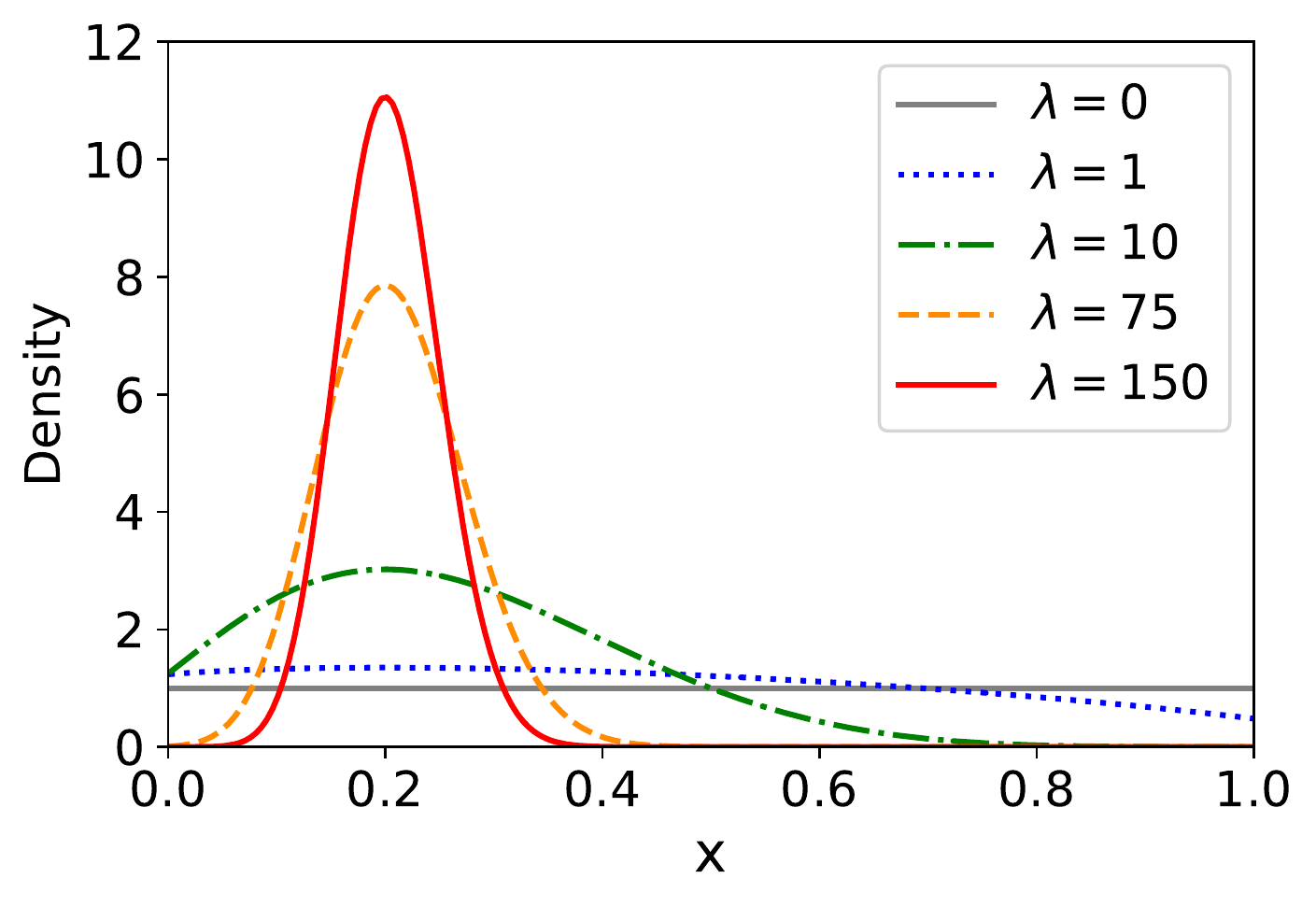}%
\centering
\\(c) $\mathtt{sBeta}$ concentration
\end{minipage}%
\caption{\textbf{Beta and sBeta density functions.} Figures (a) and (b) respectively illustrate $\mathtt{Beta}$ density function, and the presented variant referred to as $\mathtt{sBeta}$. 
Fig. (c) shows $\mathtt{sBeta}$ depending on the concentration parameter $\lambda$ while maintaining the same mode.}%
\label{fig_Beta_sBeta_pdfs}%
\end{figure}

Unlike $\mathtt{Beta}$, we seek a density function that can approximate highly peaked densities near the vertices of the simplex 
while satisfying a uni-modality constraint, thereby avoiding degenerate solutions and accounting for the statistical properties of real-world softmax predictions. \\

\subsubsection{Generalization of Beta} 

We propose the following generalization of $\mathtt{Beta}$, which we refer to as $\mathtt{sBeta}$ (scaled $\mathtt{Beta}$) in the sequel: 
\begin{equation}
    f_{\mathtt{sBeta}}{(x;\alpha,\beta)} = \frac{(x+\delta)^{\alpha-1}(1+\delta-x)^{\beta-1}}{B(\alpha,\beta)(1+2\delta)^{\alpha+\beta-2}},
\label{eq_sBeta_density}
\end{equation}
with $\delta \in \R^{+}$. Clearly, when parameter $\delta$ is set equal to $0$, the generalization in \eqref{eq_sBeta_density} reduces to $\mathtt{Beta}$ density $f_{\mathtt{Beta}}$ in \eqref{beta-density-mean-field}.  
Figures \ref{fig_Beta_sBeta_pdfs} (a) and (b) depict $\mathtt{Beta}$ and $\mathtt{sBeta}$ density functions for different values of parameters $\alpha$ and $\beta$, showing the difference between the two densities.
For instance, for $\alpha=3$ and $\beta=9$, one may observe that $\mathtt{sBeta}$ is characterized by a higher density than $\mathtt{Beta}$ in the proximity of the simplex vertex in this particular example. 
As illustrated in Figs. \ref{fig_real_models_fitting} (a), (b) and (c), $\mathtt{sBeta}$ approximations of the empirical distributions (orange histograms) of the softmax predictions are better than those obtained with the $\mathtt{Beta}$ density. Overall, $\mathtt{sBeta}$ can be viewed as a scaled variant of $\mathtt{Beta}$. It is relatively more permissive, enabling it to fit a wider range of unimodal simplex distributions.

In the following, we provide expressions of the moments (first and second central) and mode of $\mathtt{sBeta}$ as 
functions of density parameters $\alpha$ and $\beta$. As we will see shortly, these properties will enable us to integrate different constraints with our $\mathtt{sBeta}$-based probabilistic clustering objective, enforcing 
uni-modality of the distribution within each cluster while avoiding degenerate solutions. Furthermore, they will enable us to derive a computationally efficient, moment-based estimation of the density parameters. 

\subsubsection{Mean and Variance} 

\begin{Prop}
\label{prop_mean_sBeta}
The mean $E_{sB}[X]$ of $\mathtt{sBeta}$ could be expressed as a function of the density parameters as follows:
\begin{equation}
    E_{sB}[X] = \frac{\alpha}{\alpha+\beta}(1+2\delta) - \delta.
\label{eq_mean_sBeta}
\end{equation}
\end{Prop}

\begin{proof}
The mean $E_{sB}[X]$ of $\mathtt{sBeta}$ could be found by integrating $x f_{\mathtt{sBeta}}(x)$. A detailed derivation is provided in Appendix \ref{subsec_all_properties_demos}. 
\end{proof}

\begin{Prop}
\label{prop_var_sBeta}
{\em The variance $V_{sB}[X]$ of $\mathtt{sBeta}$ could be expressed as a function of the density parameters as follows}:
\begin{equation}
    V_{sB}[X] = \frac{\alpha\beta}{(\alpha+\beta)^2(\alpha+\beta+1)}(1+2\delta)^2.
    \label{eq_var_sBeta}
\end{equation}
\end{Prop}

\begin{proof}
We use $V_{sB}[X] = E_{sB}[X^2] - E_{sB}[X]^2$, with $E_{sB}[X^2]$ denoting the second moment of $\mathtt{sBeta}$, which could be found by integrating $x^2 f_{\mathtt{sBeta}}(x)$. The details of the computations of $E_{sB}[X^2]$ and $V_{sB}[X]$ are provided in the Appendix. \ref{subsec_all_properties_demos}.
\end{proof}

\subsubsection{Mode and Concentration} \label{subsubsec_mode_concentration}

The mode of a density function $f(x)$ corresponds to the value of $x$ at which $f(x)$ achieves its maximum.

\begin{Prop}
The mode $m_{sB}$ of $\mathtt{sBeta}$ could be expressed as a function of the density parameters as follows:
\begin{equation}
    m_{sB} = \frac{\alpha-1 + \delta(\alpha-\beta)}{\alpha + \beta - 2}.
\label{eq_mode_sBeta}
\end{equation}
\end{Prop}

\begin{proof}
The mode $m_{sB}$ of $\mathtt{sBeta}$ can be found by estimating at which value of $x$ the derivative of $f_{\mathtt{sBeta}}$ is equal to $0$. A detailed derivation can be found in Appendix \ref{subsec_all_properties_demos}. 
\end{proof}

Note that, when $\alpha=\beta$, we have $m_{sB} = E_{sB}[X] = \frac{1}{2}$. Otherwise, $m_{sB} \neq E_{sB}[X]$. 
Thus, $\mathtt{sBeta}$ is asymmetric when $\alpha \not = \beta$, and the variance is consequently inadequate to measure the dispersion around $\mathtt{sBeta}$ mode. 

\textbf{Concentration parameter.} Motivated by the previous observation, we present the concentration parameter $\lambda$ to measure how much a sample set is condensed around the mode value. The concept of concentration parameter has been previously discussed in \cite{kruschke2014doing}. 
With respect to the mode equation \ref{eq_mode_sBeta}, we express $\lambda=\alpha+\beta-2$ such that we have
\begin{equation}
    m_{sB} = \frac{\alpha-1 + \delta(\alpha-\beta)}{\lambda}.
    \label{eq_sBeta_mode_using_lambda}
\end{equation}
Using Eq. \eqref{eq_sBeta_mode_using_lambda}, we can express $\alpha$ and $\beta$ as functions of the mode and concentration parameter:
\begin{equation}
    \left\{
    \begin{array}{ll}
        \alpha =& 1 + \lambda \displaystyle  \frac{m_{sB} + \delta}{1 + 2\delta} \\
        \beta =& 1 + \lambda \displaystyle  \frac{1+\delta - m_{sB}}{1 + 2\delta}
    \end{array}
    \right.
    \label{params_depending_on_lambda_and_mode}
\end{equation}
Fig. \ref{fig_Beta_sBeta_pdfs} (c) shows different concentrations around a given mode when changing $\lambda$.  \\

\subsection{Proposed clustering model: \textsc{k-sBetas}}
\label{subsec_clustering_algo}

\textbf{Probabilistic clustering.} We cast clustering distributions as minimizing the following probabilistic objective (\textsc{k-sBetas}), which measures the conformity of data within each cluster to a 
multi-variate $\mathtt{sBeta}$ density function, subject to constraints that enforce uni-modality and discourage degenerate solutions:
\begin{equation}
    L_{\mathtt{sBeta}}({\mathbf U};\bm{\pi}; \bm{\Theta}) = - \sum_{i=1}^N \sum_{k=1}^K u_{i,k} \log \left (\pi_k g_{\mathtt{sBeta}}(\bm{x}_i; \bm{\alpha}_k, \bm{\beta}_k) \right )
    \label{eq_unbiased_ksbetas}
\end{equation}
\begin{equation}
    \mbox{s.t.} \quad 
    \left\{
    \begin{array}{ll}
        \lambda_{k,n} \geq \tau^- \forall {k,n}:\textrm{ Unimodal constraint, Sec. \ref{app_subsec_sBeta_constraints}} \\
        \lambda_{k,n} \leq \tau^+ \forall {k,n}:\textrm{ Avoids Dirac solutions, Sec. \ref{app_subsec_sBeta_constraints}}
    \end{array}
    \right.
\end{equation}
where $g_{\mathtt{sBeta}}$ is a multivariate extension of $\mathtt{sBeta}$:
\[{g_{\mathtt{sBeta}}}(\bm{x_i};\bm{\alpha}_k, \bm{\beta}_k) = \prod_{n=1}^D {f_{\mathtt{sBeta}}}(x_{i,n};\alpha_{k,n}, \beta_{k,n}),\]
and $\lambda_{k,n}$ denotes the concentration parameter of univariate $f_{\mathtt{sBeta}}(x_{i,n};\alpha_{k,n}, \beta_{k,n})$. In our model in \eqref{eq_unbiased_ksbetas}, $\bm{\Theta}=(\bm{\alpha}_k, \bm{\beta}_k)_{1 \leq k \leq K}$,  
with $\bm{\alpha}_k = (\alpha_{k,n})_{1 \leq n \leq D}$ and $\bm{\beta}_k = (\beta_{k,n})_{1 \leq n \leq D}$, $\bm{\pi} = (\pi_k)_{1 \leq k \leq K} \in \Delta^{K-1}$ and, following the general notation we introduced in Sec. \ref{sec_problem_formulation}, ${\mathbf U}$ denotes binary point-to-cluster assignments.    

\subsubsection{Block-coordinate descent optimization} 
\label{subsec_params_estim}

Our objective in \eqref{eq_unbiased_ksbetas} depends on three different sets of variables: ${\mathbf U}$; $\bm{\pi}$; and sBeta parameters $\bm{\Theta}$. 
Therefore, we proceed with a block-coordinate descent approach, alternating three steps. Each step optimizes the objective w.r.t a set of variables while keeping the rest of the variables fixed. 

\begin{itemize}
\item \textbf{$\bm{\Theta}$ updates, with ${\mathbf U}$ and $\bm{\pi}$ fixed:}
This section presents two different strategies for updating the $\mathtt{sBeta}$ parameters: \\

{\em (a) Solving the necessary conditions for the minimum of $L_{\mathtt{sBeta}}$ w.r.t $\alpha_{k,n}$ and $\beta_{k,n}$:} 

    Our objective in \eqref{eq_unbiased_ksbetas} is convex in each $\alpha_{k,n}$ and $\beta_{k,n}$. The global optima could be obtained by setting the gradient to zero. This could be viewed as a maximum likelihood estimation approach (MLE). Unfortunately, this yields a non-linear system of equations that, cannot be solved in closed-form.
    Therefore, we proceed to an inner iteration. In the appendix, we derive the following iterative and alternating updates to solve the non-linear system of equations:
    \begin{align} \label{eq:mle_updates}
        \bm{\alpha}_k^{(t+1)} =& \psi^{-1}(\psi(\bm{\alpha}_k^{(t)}+\bm{\beta}_k^{(t)}) \nonumber \\ 
        &+ \frac{1}{\sum_{i=1}^N u_{i,k}} \sum_{i=1}^N u_{i,k} \log(\frac{\bm{x}_i +\delta}{1+2\delta})) \\
        \bm{\beta}_k^{(t+1)} =& \psi^{-1}(\psi(\bm{\alpha}_k^{(t)}+\bm{\beta}_k^{(t)}) \nonumber \\
        &+ \frac{1}{\sum_{i=1}^N u_{i,k}} \sum_{i=1}^N u_{i,k} \log(\frac{1+\delta - \bm{x}_i}{1+2\delta})) 
    \end{align}  
    where $t$ refers to iteration number and $\psi$ corresponds to the digamma function. Note that neither $\psi$ nor $\psi^{-1}$ admit analytic expressions. Instead, we follow \cite{minka2000estimating} to approximate those functions, at the cost of additional Newton iterations. The full derivation of the updates and all required details could be found in the Appendix. \\

{\em (b)  Method of moments (MoM):}

As a computationally efficient alternative to solving a non-linear system within each outer iteration, we 
introduce an approximate estimation of sBeta parameters $\bm{\alpha}_k$ and $\bm{\beta}_k$, which we denote $\bm{\alpha}_k$ and $\bm{\beta}_k$, as the solutions to the first and second central moment equations, following Prop. \ref{prop_mean_sBeta} and Prop. \ref{prop_var_sBeta}: 
\begin{equation}
\label{moment-system}
    \left\{
    \begin{array}{ll}
        \bm{\mu}_k &= \frac{\bm{\alpha}_k}{\bm{\alpha}_k+\bm{\beta}_k}(1+2\delta) - \delta \\
        \bm{v}_{k} &= \frac{\bm{\alpha}_k\bm{\beta}_k}{(\bm{\alpha}_k+\bm{\beta}_k)^2(\bm{\alpha}_k+\bm{\beta}_k+1)}(1+2\delta)^2
    \end{array}
    \right.
\end{equation}
where $\bm{\mu}_{k}$ and $\bm{v}_{k}$ denote, respectively, the empirical means and variances of cluster $k$:
\begin{equation}
\bm{\mu}_{k} = \frac{ \sum_{i=1}^N  u_{i,k} \bm{x}_i }{ \sum_{i=1}^N  u_{i,k} } \; \mbox{and} \; 
\bm{v}_{k}=\frac { \sum_{i=1}^N  u_{i,k}(\bm{x}_i-\bm{\mu}_k)^2 }{ \sum_{i=1}^N  u_{i,k} }.
\end{equation}

The system of two equations and two unknowns in \eqref{moment-system} could be solved efficiently in closed-form, which yields the following estimates of $\mathtt{sBeta}$ parameters: 
\begin{equation}
    \left\{
    \begin{array}{ll}
        \bm{\alpha}_k &= ( \frac{\displaystyle \bm{\mu}_k^{\delta}(1-\bm{\mu}_k^{\delta})(1+2\delta)^2}{\bm{v}_{k}}-1 ) \bm{\mu}_k^{\delta} \\
        \bm{\beta}_k &= ( \frac{\displaystyle \bm{\mu}_k^{\delta}(1-\bm{\mu}_k^{\delta})(1+2\delta)^2}{\bm{v}_{k}}-1 ) (1-\bm{\mu}_k^{\delta})
    \end{array}
    \right.
\end{equation}
with $\bm{\mu}_k^{\delta}=\frac{\bm{\mu}_k+\delta}{1+2\delta}$.

\item \textbf{${\mathbf U}$ updates, with $\bm{\Theta}$ and $\bm{\pi}$ fixed:} With variables $\bm{\Theta}$ and $\bm{\pi}$ fixed, the global optimum of our objective
in \eqref{eq_unbiased_ksbetas} with respect to assignment variables ${\mathbf U}$, subject to constraints $\bm{u}_i = (u_{i,k})_{1 \leq k \leq K} \in \Delta^{K-1} \cup \{0, 1\}$, corresponds to the following closed-form solution for each assignment variable ${u}_{i,k}$:
        \begin{equation}
            u_{i,k} = \left\{
                \begin{array}{ll}
                    1 & \mbox{if } \argmaxC_{k} \pi_k \cdot g_{\mathtt{sBeta}}(\bm{x}_i; \bm{\alpha}_k, \bm{\beta}_k) = k \\
                    0 & \mbox{otherwise.}
                \end{array}
            \right.
        \label{eq_latent_hard_assignment}
        \end{equation}

\item \textbf{$\bm{\pi}$ updates, with ${\mathbf U}$ and $\bm{\Theta}$ fixed:} 
Solving the Karush–Kuhn–Tucker (KKT) conditions for the minimum of \eqref{eq_unbiased_ksbetas} with respect to $\bm{\pi}$, s.t $\bm{\pi} \in \Delta^{K-1}$, yields the following closed-form solution for each $\pi_k$:
\begin{equation}
    \pi_k=\frac{\sum_{i=1}^N u_{i,k}}{N}.
\end{equation}

\end{itemize}

\subsubsection{Handling parameter constraints}
\label{app_subsec_sBeta_constraints}

    \begin{algorithm}[t]
    \caption{Constraints for $\mathtt{sBeta}$ parameters}\label{alg_sBeta_params_constraint}
    \begin{algorithmic}[1]
    \Variables
    \State $\delta$, the $\mathtt{sBeta}$ hyper-parameter
    \State $\tau^-$, the unimodal constraint
    \State $\tau^+$, the constraint to avoid Dirac solutions
    \EndVariables
    \Function{constrain}{$\bm{\alpha}_k$, $\bm{\beta}_k$}
        
        \State $\bm{m}_{sB} \gets \frac{\bm{\alpha}_k-1 + \delta(\bm{\alpha}_k-\bm{\beta}_k)}{\bm{\alpha}_k + \bm{\beta}_k - 2}$ \Comment{Estimate the mode}
        \State $\bm{\lambda}_k \gets \bm{\alpha}_k+\bm{\beta}_k-2$ \Comment{Estimate the concentration}
        \For{$n \gets 1\dots D$}  
            \If{$\lambda_{k,n} < \tau^-$}
            \State $\lambda_{k,n} \gets \tau^-$ \Comment{Ensure strictly unimodal solutions}
            \ElsIf{$\lambda_{k,n} > \tau^+$}
            \State $\lambda_{k,n} \gets \tau^+$ \Comment{Avoid Dirac solutions}
        \EndIf
        \EndFor
        \State $\bm{\alpha}_k \gets 1 + \bm{\lambda}_k \displaystyle  \frac{\bm{m}_{sB} + \delta}{1 + 2\delta}$ \Comment{Update $\bm{\alpha}_k$ using \eqref{params_depending_on_lambda_and_mode}}
        \State $\bm{\beta}_k \gets 1 + \bm{\lambda}_k \displaystyle  \frac{1+\delta - \bm{m}_{sB}}{1 + 2\delta}$ \Comment{Update $\bm{\beta}_k$ using \eqref{params_depending_on_lambda_and_mode}}
        \State \Return $\bm{\alpha}_k$, $\bm{\beta}_k$
    \EndFunction
    \end{algorithmic}
    \end{algorithm}  

Our model in \eqref{eq_unbiased_ksbetas} integrates two constraints on concentration parameters $\lambda_{k,n}$; one discourages multi-modal solutions, and the other avoids degenerate, highly peaked (Dirac) densities.
    
Notice that the $\mathtt{sBeta}$ density is bimodal when $\alpha_{k,n}<1$ and $\beta_{k,n}<1$, which corresponds to $\lambda_{k,n}<0$ (as $\lambda_{k,n} = \alpha_{k,n} + \beta_{k,n} - 2$). 
Moreover, setting $\alpha_{k,n}=\beta_{k,n}=1$, which yields $\lambda_{k,n}=0$, corresponds to the uniform density. Thus, to constrain the $\mathtt{sBeta}$ densities to be strictly unimodal, we must restrict $\lambda_{k,n}$ to positive values. 
In practice, we constrain $\lambda_{k,n}$ to be greater or equal to a threshold $\tau^->0$, which we simply set to $1$.
    
A high-density concentration around the $\mathtt{sBeta}$ mode corresponds to high values $\lambda_{k,n}$. Thus, to avoid degenerate, highly peaked (Dirac) densities, we constrain $\lambda_{k,n}$ to stay smaller than or equal to a fixed, strictly positive threshold $\tau^+$. 
    
Overall, we enforce $ \tau^- \leq \lambda_{k,n} \leq \tau^+$ on $\mathtt{sBetas}$ parameters using Eq. \eqref{params_depending_on_lambda_and_mode}, as detailed in Algorithm \ref{alg_sBeta_params_constraint}. This procedure allows for maintaining the same mode.
    Fig. \ref{fig_avoid_dirac_modelling_using_sBeta} (a) shows $\mathtt{sBeta}$ density estimation, following the constraint $\lambda_{k,n} \geq \tau^-$, on a bimodal distribution sample. Fig. \ref{fig_avoid_dirac_modelling_using_sBeta} (b) shows $\mathtt{sBeta}$ estimation, following the constraint $\lambda_{k,n} \leq \tau^+$, on a Dirac distribution sample.

    \begin{figure}[h]
    \centering
    \begin{minipage}{0.70\columnwidth}%
    \includegraphics[width=\columnwidth]{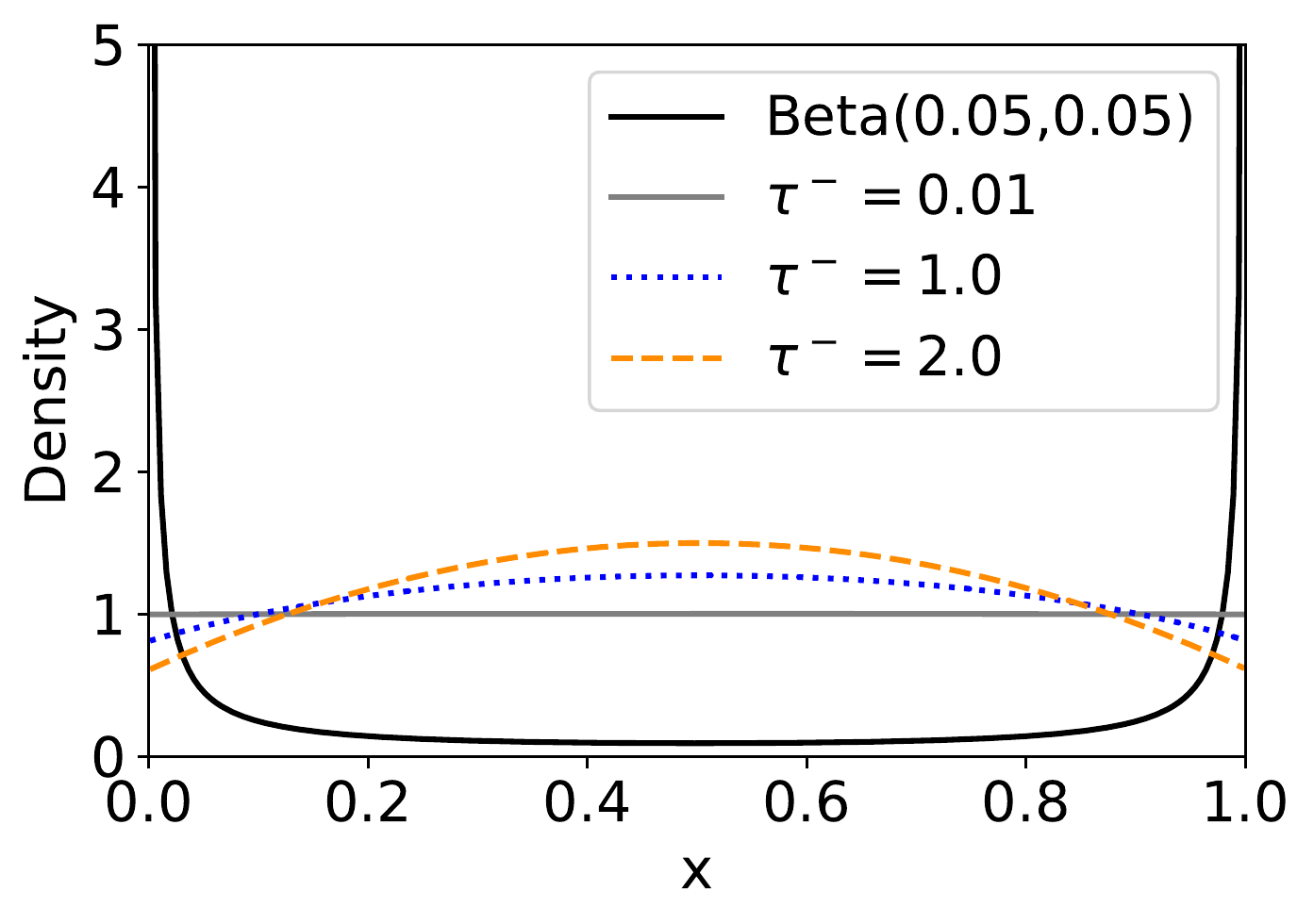}%
    \centering
    \\(a) Avoid bimodal solutions 
    \vspace{0.15in} 
    \end{minipage}%
    
    \begin{minipage}{0.70\columnwidth}%
    \includegraphics[width=\columnwidth]{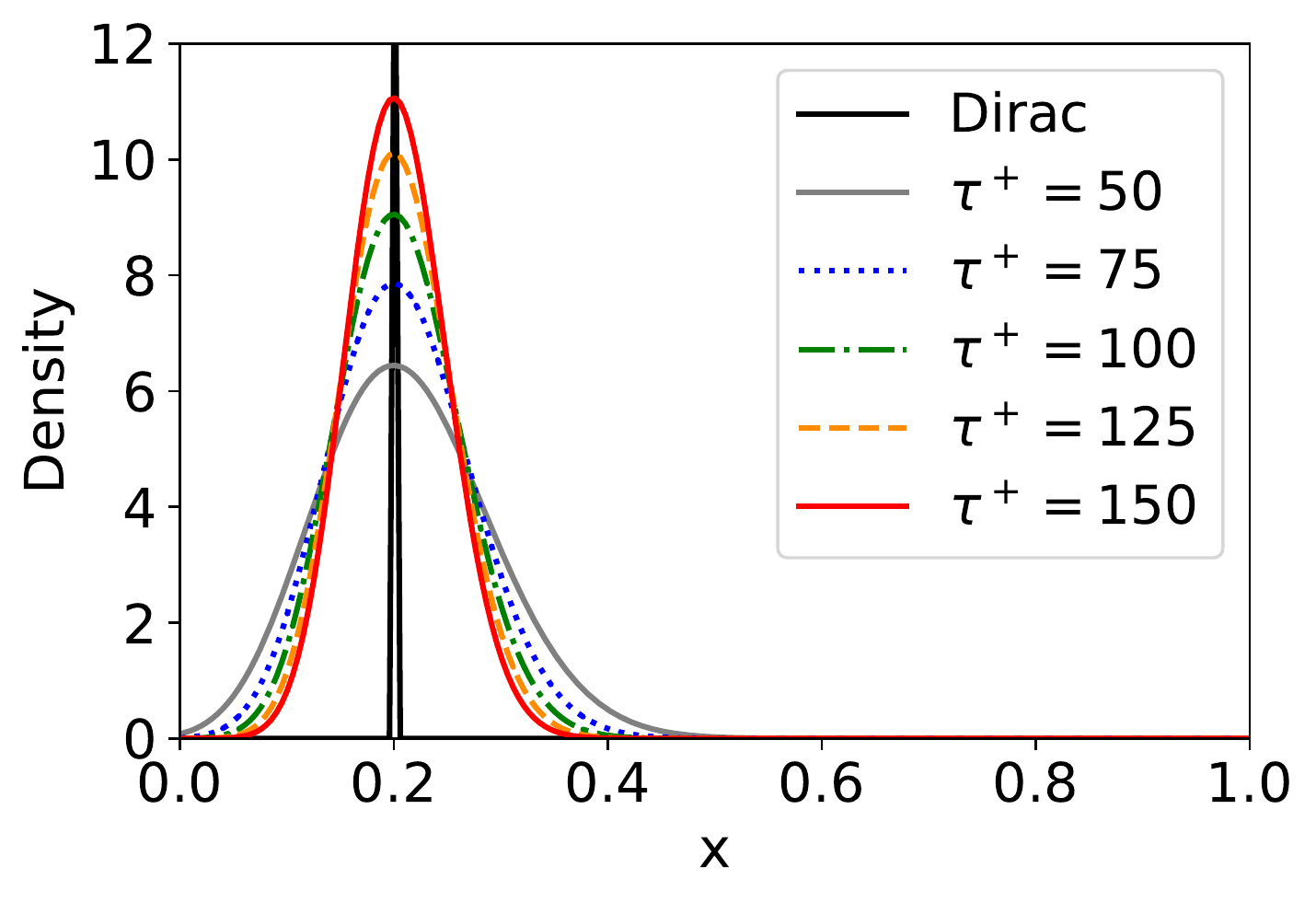}%
    \centering
    \\(b) Avoid Dirac solutions
    \end{minipage}%
    \vskip -0.05in
    \caption{\textbf{Visualization of the introduced constraints.} Fig. (a) shows $\mathtt{sBeta}$ density estimation, constrained with the threshold $\tau^-$, on a sample following a bimodal $\mathtt{Beta}$ distribution. Fig. (b) shows $\mathtt{sBeta}$ density estimation, constrained with the threshold $\tau^+$, on a sample following a Dirac distribution.}%
    \label{fig_avoid_dirac_modelling_using_sBeta}%
    \end{figure}
    
\subsubsection{Centroid initialization} 
\label{sssec_ctrds_init}
Clustering algorithms are known to be sensitive to their parameter initialization. They often converge to a local optimum. To reduce this limitation, the seeding initialization strategy k-means++, proposed in \cite{arthur2006k} and then thoroughly studied in \cite{bachem2016approximate}, is broadly used for centroid initialization. However, in the context of the clustering of softmax predictions, we can assume that softmax predictions generated by deep learning models were optimized upstream to be sets of one-hot vectors\footnote{A one-hot vector both refers to a semantic class and to a vertex on the probability simplex.}. Thus, we propose initializing the centroids as vertices of the target probability simplex domain. More specifically, we initialize all $\alpha_{k,n}$ and $\beta_{k,n}$ parameters such that they model exponential densities on $[0,1]$ at the start. In other words, each initial mode is set as a vertex among all possible vertices on the probability simplex domain. Beyond improving \textsc{k-sBetas}, this simple initialization strategy unanimously improves the scores of every tested clustering method\footnote{Table \ref{table_kmeansplusplus_init_vs_vertices_init} in Appendix empirically supports this statement.}.

\subsubsection{From cluster-label to class-label assignment}
\label{sssec_lbl_alignment}
After clustering, we must align the obtained clusters with the target classes. One can align each cluster centroid with the closest one-hot vector by using the argmax function. However, in this case, a one-hot vector corresponding to a given class could be matched to several cluster centroids, thereby assigning them to the same class. To prevent this problem on closed-set challenges, i.e., with the prior knowledge $K=D$, we use the optimal transport Hungarian algorithm \cite{kuhn1955hungarian}. Specifically, we compute the Euclidean distances between all the centroids and one-hot vectors.
Then, we apply the Hungarian algorithm to this matrix of distances to find the lowest-cost way, assigning a separate 
class to each cluster.

    \begin{algorithm}[t]
    \caption{k-sBetas algorithm.}\label{alg_k_sBetas}
    \begin{algorithmic}[1]
    \Require Dataset of probability simplex points defined as $\mathcal{X}=\{\bm{x}_i\}_{i=1}^N \in \Delta^{D-1}$, with $N$ the total number of points. Number of clusters $K$. The maximum number $t_{end}$ of iterations.
    \Ensure ${\mathbf U}$, the clustering hard label assignments
    \Variables
    \State $\delta$, the $\mathtt{sBeta}$ hyper-parameter
    \State $\tau^-$, the unimodal constraint
    \State $\tau^+$, the constraint to avoid Dirac solutions
    \EndVariables
    \State ${\mathbf U} \gets [0]_{N,K}$ \Comment{Arbitrary label initialization}
    \State $\bm{\pi} \gets \{ \frac{1}{K} \}_{k=1}^K$ \Comment{Initialize uniform class proportions}
    \State Initialize $\bm{\Theta}$ using \textit{vertex init}. \Comment{Sec. \ref{sssec_ctrds_init}.}
    \State $t \gets 0$
    \Repeat
        \If{$t > 0$}
            \For{$k \gets 1\dots K$}  \Comment{Loop over clusters}
                \State Estimate $\bm{\alpha}_k$, $\bm{\beta}_k$ with MoM/MLE
                    \Comment{Sec. \ref{subsec_params_estim}.} 
                \State  $\bm{\alpha}_k$, $\bm{\beta}_k$ $\gets$ \Call{constrain}{$\bm{\alpha}_k$, $\bm{\beta}_k$} 
                \Comment{Algorithm \ref{alg_sBeta_params_constraint}}
            \EndFor
        \EndIf
        \For{$i \gets 1\dots N$}  \Comment{hard label assignment}
            \For{$k \gets 1\dots K$}  
                \If{$\argmaxC_{k} \pi_k \cdot g_{\mathtt{sBeta}}(\bm{x}_i; \bm{\alpha}_k, \bm{\beta}_k) = k$}
                    \State $u_{i,k} \gets 1$
                \Else
                    \State $u_{i,k} \gets 0$
                \EndIf
            \EndFor
        \EndFor
        \State $\bm{\pi} \gets \frac{\sum_{i=1}^N \bm{u}_i}{N}$ \Comment{Estimate class proportions}
        \State $t \gets t+1$
    \Until{convergence or $t=t_{end}$}
    \State \Return ${\mathbf U}$
    \end{algorithmic}
    \end{algorithm}

\section{Experiments}
\label{sec_Experiments}

    Throughout this section, we show different applications of the proposed method and compare it with the state-of-the-art clustering methods discussed in Sec. \ref{sec_Related_Work}. We first validate our implementations on synthetic datasets and on real softmax predictions from unsupervised domain adaptation (UDA) source models (Sec. \ref{UDA-experiements}). Then, in Sec. \ref{subsec_zero_shot_using_clip_embedding}, we demonstrate the usefulness of our simplex-clustering method in the context of transductive zero- and one-shot predictions using contrastive language-image pre-training (CLIP) \cite{radford2021clip} for the source model. Finally, we provide an application example for a dense prediction task: real-time UDA for road segmentation (Sec. \ref{subsec_real_time_UDA_road_seg}). In each of these settings, our method is used as a plug-in on top of the output predictions from a black-box deep learning model. Along with the experiments and comparisons that follow, we also provide ablation studies, which highlight the effect of each component of the proposed framework.
    
    \textbf{Comparisons.} We compare our method to several state-of-the-art clustering methods. Specifically, we include {\textsc{KL k-means}} \cite{chaudhuri2008finding} \cite{wu2008sail} and \textsc{HSC} \cite{nielsen2019clustering}, which are specifically designed to deal with probability simplex vectors. We additionally compare our method to standard clustering methods: \textsc{k-means}, \textsc{GMM} \cite{biernacki2000assessing}, \textsc{k-medians} \cite{bradley1997clustering}, \textsc{k-medoids} \cite{kaufman1990partitioning} and \textsc{k-modes} \cite{carreira2013k}. 
    Both \textsc{k-means} and \textsc{KL k-means} use the mean as the cluster's parameter vector, with the former being based on the Euclidean distortion and the latter on the Kullback-Leibler one. Extensively studied in \cite{bradley1997clustering}, \textsc{k-medians} uses the median as a cluster representative and the Manhattan distance. Clustering with \textsc{k-medoids} uses the Euclidean distance while estimating medoid representations of the clusters with the standard PAM algorithm \cite{kaufman1990partitioning}. Finally, \textsc{k-modes} \cite{carreira2013k} uses a kernel-induced distortion and computes cluster modes via the Meanshift algorithm \cite{cheng1995mean}.
    
    \textbf{\textsc{k-Dirs}.} We also implemented a simplex-clustering strategy, which estimates a multivariate density per cluster using the 
    Dirichlet density function. We used the iterative parameter estimation proposed in \cite{minka2000estimating}. This algorithm, which we refer to as \textsc{k-Dirs}, represents a baseline for the artificial datasets built from the Dirichlet distributions.
    
    \textbf{Proposed \textsc{k-sBetas}.} Our clustering algorithm uses $f_{sBetas}$ as density function and the method of moments (MoM) detailed in Sec. \ref{subsec_params_estim} for parameter estimation. We empirically consolidate this choice in the ablation study in section \ref{subsec_ablation_study}. Note that \textbf{\textsc{k-Betas}} corresponds to a non-scaled variant of \textsc{k-sBetas}, i.e., when we set $\delta=0$.
    
    \textbf{Hyper-parameters.} Across all the experiments, we set the scaling hyper-parameter $\delta$ used in \textsc{k-sBetas} to the same value $0.15$. We set $\tau^-=1$ and $\tau^+=165$ for parameter constraints. The maximum number of clustering iterations is set to $25$ for all the clustering algorithms. For each method, we use the parameter initialization detailed in Sec. \ref{sssec_ctrds_init}. This technique is also applied to the methods based on the feature maps, for consistency and fairness: For each probability simplex vertex (i.e., one-hot vector), we select the feature-map point whose softmax prediction most closely aligns with this one-hot vector. We provide comparisons and justifications for our design choices in the ablation studies in Sec. \ref{subsec_ablation_study}.

    \textbf{Reproducibility.} We used the implementation of the scikit-learn library for GMM \footnote{The \textsc{GMM} code is available at: \url{https://github.com/scikit-learn/scikit-learn/blob/7e1e6d09b/sklearn/mixture/_gaussian_mixture.py}.}, and the implementation provided by the authors for \textsc{HSC}\footnote{The \textsc{HSC} code is available at: \url{https://franknielsen.github.io/HSG/}.}. We have implemented all the other clustering algorithms \footnote{The codes for \textsc{KL k-means}, \textsc{k-means}, \textsc{k-medians}, \textsc{k-medoids}, \textsc{k-modes}, \textsc{k-Dirs}, \textsc{k-Betas} and \textsc{k-sBetas} are available at our repository: \url{https://github.com/fchiaroni/Clustering_Softmax_Predictions}}.
        
    \textbf{Evaluation metrics.} We use the standard \textit{Normalized Mutual Information} (NMI) to evaluate the clustering task for each dataset. 
    To evaluate the class-prediction scores, we use the classification \textit{Accuracy} (Acc) for the balanced datasets or the \textit{Intersection over Union} (IoU) measure for the imbalanced ones. The process of aligning the clusters with the class labels, which is necessary for computing the ACC and mIoU, is detailed in Sec. \ref{sssec_lbl_alignment}.
    
    \subsection{Unsupervised Domain Adaptation}
    \label{UDA-experiements}
        
    \subsubsection{Synthetic experiments}
        The purpose of the following synthetic experiments is to benchmark the different clustering methods on a simple, artificially generated task. \\
        
        \textbf{\textit{Simu} dataset}. We generate balanced mixtures composed of three Dirichlet distributions, defined on the 3-dimensional probability simplex, $\Delta^{2}$. The corresponding Dirichlet parameters are $\bm{\alpha_1} = (1,1,5)$, $\bm{\alpha_2}=(25,5,5)$, $\bm{\alpha_3}=(5,7,5)$. Each component is made biased towards one different vertex of the simplex, thereby simulating the softmax predictions of a deep model for a certain class. Additionally, each component captures a different type of distribution\footnote{We provide the 2D visualizations of these simulated distributions in Appendix, Fig. \ref{fig_three_simul_diri_distr}.}: $Dir(\bm{\alpha_1})$ corresponds to an exponential distribution on $\Delta^2$, $Dir(\bm{\alpha_2})$ to an off-centered distribution with small variance, and $Dir(\bm{\alpha_2})$ to a more centered distribution, with a wider variance. All the 
        components contribute equally to the final mixture. In total, $10^5$ examples are sampled. We perform 5 random runs, each using new examples.
      
        \textbf{Results.} The third column of Table \ref{table_bal_scores} reports the NMI and Acc scores of different approaches on the Simu dataset. As expected, \textsc{k-Dirs} and \textsc{k-Betas} yielded close scores as they both model the $\mathtt{Beta}$ marginal densities. \textsc{k-sBetas} yields lower scores due to the scaling factor, $\delta>0$. However, it behaves better than all the other clustering methods thanks to a proper statistical modeling of simplex data.

    \subsubsection{Clustering the softmax predictions of deep models}
    \label{subsec_softmax_preds_comp}
        We now compare the different clustering approaches on real-world distributions, where data points correspond to the softmax predictions of 
        deep-learning models. \\
        
        \textbf{Setup.} First, we employ the \textbf{SVHN$\rightarrow$MNIST} challenge, where a source model is trained on SVHN \cite{netzer2011reading} and applied to the MNIST \cite{lecun1998gradient} test set, which is composed of $10^4$ images labelled with 10 different semantic classes. Additionally, we experiment with the more difficult \textbf{VISDA-C} challenge \cite{peng2018visda}, which contains 55388 samples split into 12 different semantic classes. For both \textbf{SVHN$\rightarrow$MNIST} and \textbf{VISDA-C}, we use the common network architectures and training procedures detailed in \cite{liang_we_2020}. 
        For these closed-set UDA experiments, we use the optimal-transport Hungarian algorithm
        to obtain cluster-to-class label assignment, as detailed in Sec. \ref{sssec_lbl_alignment}.
        \\ 
        
        \textbf{Results.} Table \ref{table_bal_scores} shows interesting comparative results for clustering real-world softmax predictions. In the SVHN$\rightarrow$MNIST setting, simplex-tailored methods \textsc{KL k-means} and \textsc{k-sBetas} clearly stand out in terms of both the NMI and Accuracy. In particular, the proposed \textsc{k-sBetas} achieves the best scores. On the VISDA-C challenge, however, \textsc{KL k-means} yielded a performance below the baseline, similarly to the other standard clustering methods. \textsc{k-Dirs} failed to converge. In contrast, \textsc{k-sBetas} outperforms the baseline by a margin, achieving the best scores. 
        
        Table \ref{table_comp_time} reports the running times of all the methods. When using MoM for parameter estimation, \textsc{k-sBetas} has a running time close to or even lower than the standard GMM approach. Furthermore, a GPU-based implementation of \textsc{k-sBetas} could be made even faster, which makes it appealing for large-scale datasets such as VISDA-C.

            \begin{table}[t]
                \caption{Comparative \textbf{probability simplex clustering} on \textit{Simu}, SVHN$\rightarrow$MNIST and VISDA-C datasets. The \textsc{feature map} layer is the bottleneck layer on top of the convolutional layers. It outputs a feature vector of size 256 for each input image. The \textsc{logit} points are obtained with the fully connected classifier on top of the bottleneck layer so that the logit point predicted for each input example is a vector of size $K$. The probability \textsc{simplex} points, which represent the predictions of a black-box model, are obtained with the standard softmax function on top of the \textsc{logit} layer.}
                \label{table_bal_scores}
                \begin{center}
                \begin{small}
                \begin{sc}
                \resizebox{\columnwidth}{!}{%
                \begin{tabular}{lcccccc}
                    \toprule
                    \multirow{2}{*}{Method} & \multirow{2}{*}{Layer used} & \multicolumn{1}{c}{Simu}  & \multicolumn{2}{c}{SVHN$\rightarrow$MNIST}  & \multicolumn{2}{c}{VISDA-C}  \\
                     \cmidrule(r){3-3} \cmidrule(r){4-5} \cmidrule(r){6-7}
                    &  & (NMI) & (NMI) & (Acc) & (NMI) & (Acc) \\
                    \midrule
                    k-means & feature map & -  & 67.3 & 74.3 & 30.0 & 22.7 \\ 
                    \midrule
                    k-means & logit & - & 67.1 & 74.5 & 32.7 & 33.1 \\ 
                    \midrule
                    argmax & &  60.1 & 58.6 & 69.8 & 36.5 & 53.1 \\ 
                    k-means & & 76.6 & 58.4 & 68.9 & 37.6 & 47.9 \\ 
                    GMM & & 75.8 & 61.9 & 69.2 & 36.3 & 49.4 \\ 
                    k-medians & & 76.8 & 58.6 & 68.8 & 35.9 & 40.0 \\ 
                    k-medoids & simplex & 60.8 & 58.9 & 71.3 & 36.5 & 46.8 \\ 
                    k-modes & (Black-Box) & 76.2 & 59.5 & 71.3 & 34.2 & 51.8 \\ 
                    KL k-means & & 76.2 & 63.3 & 75.5 & 39.8 & 51.2 \\ 
                    HSC & & 9.3 & 59.2 & 68.9 & 28.7 & 18.1 \\ 
                    k-Dirs & & 81.3 & 57.7 & 68.8 & \multicolumn{2}{c}{fails} \\ 
                    k-Betas & & 81.1 & 53.3 & 51.5 & 36.7 & 19.8 \\
                    \rowcolor{Gray}
                    k-sBetas & & 79.2 & \textcolor{black}{\textbf{65.0}} & \textcolor{black}{\textbf{76.6}} & \textcolor{black}{\textbf{40.3}} & \textcolor{black}{\textbf{56.0}} \\
                    \bottomrule
                \end{tabular}
                }
                \end{sc}
                \end{small}
                \end{center}
                \vskip 0.15in
            \end{table}            
                    
            \begin{table}[t]
                \caption{\textbf{Computational time} comparison in seconds. All methods are executed on the same hardware: \textit{CPU 11th Gen Intel(R) Core(TM) i7-11700K 3.60GHz}, and \textit{GPU NVIDIA GeForce RTX 2070 SUPER}. MLE-1000 and MLE-500 refer to the MLE parameter estimation with a maximum of 1000 and 500 MLE iterations, respectively, while MoM refers to the method of moments.}
                \label{table_comp_time}
                \begin{center}
                \begin{small}
                \begin{sc}
                \resizebox{\columnwidth}{!}{%
                    \begin{tabular}{lcc}
                        \toprule
                         \multirow{2}{*}{Method} & SVHN$\rightarrow$MNIST & VISDA-C \\
                         & ($N=10 000$, $K=10$) & ($N=55 388$, $K=12$) \\
                        \midrule
                        k-means - feature map & 2.36 & 18.01 \\
                        \midrule
                        k-means - logit/simplex & 0.06 & 1.27 \\
                        GMM & 0.43 & 10.59 \\
                        k-medians & 5.94 & 48.39 \\
                        k-medoids & 8.83 & 195.00 \\
                        k-modes & 0.08 & 4.71 \\
                        KL k-means & 0.27 & 2.79 \\
                        HSC & 8494.54 & $>$One day  \\
                        \rowcolor{Gray}
                        k-sBetas (MLE-1000) & 64.32 & 107.43 \\
                        \rowcolor{Gray}
                        k-sBetas (MLE-500) & 34.22 & 58.68 \\
                        \rowcolor{Gray}
                        k-sBetas (MoM) & 0.48 & 3.61 \\
                        \rowcolor{Gray}
                        k-sBetas (MoM, GPU-based) & 0.13 & 0.49 \\
                        \bottomrule
                    \end{tabular}
                    }
                    \end{sc}
                    \end{small}
                    \end{center}
                    \vskip 0.1in
            \end{table}

        \textbf{Feature maps vs. simplex predictions.} 
        Interestingly, Table \ref{table_bal_scores} shows that k-means, when applied to the logit points, yields scores that are similar or even better than when applied to the feature maps. 
        This suggests that the classifier head, which predicts the logit points, does not deteriorate the semantic information of the bottleneck layer. Table \ref{table_comp_time} shows that logit and simplex clustering are computationally less demanding than clustering the feature maps. This is due to the fact that the dimension of the simplex points is $K$, which is considerably smaller than the dimension of feature maps.
        In addition, Table \ref{table_bal_scores} shows that the proposed simplex-clustering method substantially outperforms the \textsc{k-means} feature-map clustering on the most difficult/realistic dataset, VISDA-C.
        
        It is worth noting that the explored softmax-prediction datasets, SVHN$\rightarrow$MNIST and VISDA-C, are challenging in this particular black-box setting. This enables interesting comparisons with model-agnostic clustering methods. For reproducibility and future improvements, we made these black-box softmax predictions publicly available: \url{https://github.com/fchiaroni/Clustering_Softmax_Predictions}.

    \subsection{Transductive zero-shot and one-shot learning}
    \label{subsec_zero_shot_using_clip_embedding}
    
    We now demonstrate the general applicability and effectiveness of our method in the practical and realistic few-shot problem, which involves only an unknown fraction of the original set of classes in a small, unlabeled query (test) set \cite{MartinNeurIPS22}. We tackle transductive zero- and one-shot inference using the foundational CLIP model \cite{radford2021clip} by clustering its zero-shot softmax predictions. CLIP is based on two deep encoders, one for image representation and the other for text inputs. Coupled with projections, this dual structure yields image and text embeddings within the same low-dimensional space. Trained on a large-scale dataset of text-image pairs, CLIP maximizes the cosine similarity between the embedding of an image and its corresponding text description, which makes it well-suited for zero-shot prediction. At inference time, the classification of a given image into one of $K$ classes is given by the zero-shot softmax prediction of CLIP: ${\sigma(\bm{z})}_j = \frac{\exp(z_j)}{\sum_{l=1}^D \exp(z_l)}$, where logit $z_j$ evaluates the cosine similarity between the image embedding and the representation of a text prompt describing the class, typically `'a photo of a [name of class k]'' \cite{radford2021clip}. 
    In the one-shot setting, we cluster softmax predictions of the same form, except that, for each class, the text embedding is replaced by the visual embedding of the labeled shot of the class, i.e., the single labeled image of the class.
    \\
    
    \textbf{Experimental Setup:}
    We assessed our method using four distinct architectures of the CLIP visual encoder: ResNet-50, ResNet-101 \cite{resnet}, ViT-B/32, and ViT-B/16 \cite{dosovitskiy2021an}. Our experiments were performed across eight datasets: CIFAR-100 \cite{krizhevsky2009learning}, Stanford-Cars \cite{krause20133d}, FGVC Aircraft \cite{maji2013fine}, Caltech101 \cite{1384978}, Food101 \cite{bossard2014food}, Flowers102 \cite{nilsback2008automated}, Sun397-100 \cite{xiao2010sun} and ImageNetv2-100 \cite{recht2019imagenet}. For Sun397-100 
    and ImageNetv2-100, we restrict the evaluation to the first 100 classes. For each dataset, we generated 100 query sets, each containing 64 unlabeled examples sampled from a random selection of 2 to 10 classes.

    We compare our method with the inductive baseline, i.e., CLIP zero-shot prediction, and with transductive state-of-the-art few-shot learning methods, which leverage the statistics of the unlabeled query set, and operate on the feature maps: \textsc{BD-CSPN} \cite{liu2020prototype}, \textsc{Lap-Shot} \cite{ziko2020laplacian} and \textsc{TIM} \cite{boudiaf2020information}. Additional comparisons include \textsc{k-means}, \textsc{KL k-means}, and our \textsc{k-sBetas} method, which we apply directly to the softmax predictions of CLIP. 
    In the one-shot setting, the initialization prototype vector for each class is the CLIP-encoded image from the labeled support set. In the zero-shot setting, the initialization prototype vector for each class is the CLIP-encoded text prompt. Post-convergence, we directly employ the $\text{argmax}$ function for cluster-to-class assignments. This function assigns each estimated cluster to the class whose one-hot vector is nearest to the cluster centroid. \\

   \textbf{Results.} Tables \ref{One_SHOT_clip_embedding_qs_64} and \ref{Zero_SHOT_clip_embedding_qs_64} show the results for these One-shot and Zero-shot settings. 
   A first key observation is the enhanced effectiveness of the probability-simplex methods over those based on the feature 
   maps. On another note, one may observe that feature-map methods TIM and k-means show similar results in these settings. This observation is consistent with a previous work \cite{jabi_deep_2019}, which showed that, under certain mild conditions, along with common posterior models and parameter regularization, such models become equivalent.
   Overall, the proposed \textsc{k-sBetas} method demonstrates a clear and consistent improvement over the inductive CLIP baseline and the other transductive few-shot methods across a variety of datasets and network encoders. This indicates its potential as a robust and versatile approach in enhancing the performance of foundation models, like CLIP, in practical applications with limited labeled data and unknown class distributions.

            \begin{table*}[t]
                \addtolength{\tabcolsep}{-3pt}
                \caption{\textbf{1-shot results.} \textbf{Query set size is 64.} The proposed \textsc{k-sBetas} further boosts the results of the existing CLIP model by clustering its soft predictions on the query set.}
                \label{One_SHOT_clip_embedding_qs_64}
                \begin{center}
                    \begin{small}
                        \begin{sc}
                            \resizebox{\textwidth}{!}{%
                            \begin{tabular}{lcccccccccccccccccc}
                                \toprule
                                \multirow{2}{*}{Method} & \multirow{2}{*}{Layer used} & \multirow{2}{*}{Network} & \multicolumn{2}{c}{cifar-100} & \multicolumn{2}{c}{Stanford-cars} & \multicolumn{2}{c}{FGVC} & \multicolumn{2}{c}{Caltech101} & \multicolumn{2}{c}{Food101} & \multicolumn{2}{c}{Flowers102}  & \multicolumn{2}{c}{Sun397-100} & \multicolumn{2}{c}{ImageNetv2-100} \\
                                \cmidrule(r){4-5} \cmidrule(r){6-7} \cmidrule(r){8-9} \cmidrule(r){10-11} \cmidrule(r){12-13} \cmidrule(r){14-15} \cmidrule(r){16-17} \cmidrule(r){18-19} 
                                 & & & (NMI) & (Acc) & (NMI) & (Acc) & (NMI) & (Acc) & (NMI) & (Acc) & (NMI) & (Acc) & (NMI) & (Acc) & (NMI) & (Acc) & (NMI) & (Acc) \\
\midrule
CLIP & feature map & RN50 & 54.1 & 17.2 & 61.6 & 26.9 & 55.4 & 15.8 & 77.7 & 64.5 & 58.8 & 29.4 & 74.0 & 58.5 & 65.3 & 43.8 & 63.2 & 31.3 \\
BD-CSPN & feature map & RN50 & 54.1 & 13.8 & 60.3 & 19.7 & 56.1 & 13.7 & 70.0 & 51.1 & 57.4 & 22.7 & 70.5 & 46.8 & 60.4 & 32.6 & 61.8 & 25.2 \\
Lap-Shot & feature map & RN50 & 54.1 & 13.9 & 60.3 & 19.7 & 56.0 & 13.7 & 70.2 & 51.3 & 57.4 & 22.8 & 70.5 & 46.8 & 60.5 & 32.8 & 61.8 & 25.2 \\
TIM & feature map & RN50 & \textbf{56.3} & 14.3 & 59.0 & 14.5 & \textbf{57.2} & 11.6 & 63.9 & 35.4 & 57.0 & 17.8 & 68.7 & 34.3 & 55.9 & 22.9 & 60.8 & 18.0 \\
k-means & feature map & RN50 & \textbf{56.3} & 14.3 & 59.0 & 14.5 & \textbf{57.2} & 11.6 & 63.9 & 35.4 & 57.0 & 17.8 & 68.7 & 34.3 & 55.9 & 22.9 & 60.8 & 18.0 \\
k-means & simplex & RN50 & 52.1 & 17.5 & 65.6 & 29.3 & 55.5 & 16.0 & 79.6 & 63.9 & 60.2 & 31.4 & 80.1 & 65.3 & 69.9 & 45.7 & 65.6 & 31.5 \\
KL k-means & simplex & RN50 & 52.1 & 17.6 & 65.5 & 29.3 & 55.4 & 16.1 & 79.4 & 64.1 & 60.1 & 31.2 & 80.0 & 65.0 & 69.7 & 45.8 & 65.7 & 31.5 \\
\rowcolor{Gray}
k-sBetas & simplex & RN50 & 51.9 & \textbf{18.6} & \textbf{66.6} & \textbf{32.6} & 55.4 & \textbf{17.9} & \textbf{82.6} & \textbf{69.5} & \textbf{61.0} & \textbf{34.0} & \textbf{80.6} & \textbf{67.7} & \textbf{71.9} & \textbf{48.7} & \textbf{65.8} & \textbf{37.7} \\
\midrule
CLIP & feature map & RN101 & 56.8 & 21.3 & 66.2 & 31.5 & 55.9 & 19.8 & 78.2 & 67.2 & 64.7 & 37.2 & 76.7 & 61.4 & 68.7 & 51.4 & 66.3 & 37.2 \\
BD-CSPN & feature map & RN101 & 56.6 & 17.4 & 63.5 & 23.6 & 55.8 & 16.3 & 69.3 & 53.1 & 63.4 & 30.2 & 71.9 & 48.6 & 62.8 & 39.1 & 65.1 & 31.1 \\
Lap-Shot & feature map & RN101 & 56.6 & 17.4 & 63.5 & 23.6 & 55.8 & 16.3 & 69.4 & 53.2 & 63.4 & 30.3 & 71.9 & 48.7 & 62.8 & 39.3 & 65.1 & 31.2 \\
TIM & feature map & RN101 & \textbf{59.6} & 19.9 & 61.8 & 16.7 & 56.7 & 14.3 & 64.9 & 40.4 & 63.9 & 27.3 & 73.2 & 42.0 & 59.5 & 28.6 & 64.2 & 23.3 \\
k-means & feature map & RN101 & \textbf{59.6} & 19.9 & 61.8 & 16.7 & 56.7 & 14.3 & 64.9 & 40.4 & 63.9 & 27.3 & 73.2 & 42.0 & 59.5 & 28.6 & 64.2 & 23.3 \\
k-means & simplex & RN101 & 57.1 & 21.8 & 72.6 & 35.0 & \textbf{58.5} & 20.1 & 79.9 & 66.5 & 67.3 & 40.6 & 84.9 & 67.6 & 72.3 & 52.7 & 71.0 & 41.5 \\
KL k-means & simplex & RN101 & 57.1 & 21.9 & 72.4 & 34.8 & 58.4 & 20.0 & 80.0 & 66.6 & 67.3 & 40.6 & 84.9 & 67.6 & 72.4 & 52.5 & 70.8 & 41.7 \\
\rowcolor{Gray}
k-sBetas & simplex & RN101 & 57.4 & \textbf{23.2} & \textbf{74.1} & \textbf{37.2} & 58.3 & \textbf{22.3} & \textbf{81.5} & \textbf{70.2} & \textbf{68.1} & \textbf{42.0} & \textbf{85.2} & \textbf{70.1} & \textbf{74.9} & \textbf{56.7} & \textbf{71.2} & \textbf{46.8} \\
\midrule
CLIP & feature map & ViT-B/32 & 59.0 & 26.2 & 63.0 & 28.0 & 55.8 & 18.8 & 82.0 & 69.9 & 64.8 & 38.4 & 77.2 & 61.8 & 69.1 & 44.4 & 66.4 & 36.6 \\
BD-CSPN & feature map & ViT-B/32 & 58.5 & 21.7 & 60.8 & 21.5 & 55.9 & 15.2 & 73.1 & 57.7 & 62.4 & 30.6 & 71.7 & 49.9 & 63.8 & 36.4 & 65.0 & 30.6 \\
Lap-Shot & feature map & ViT-B/32 & 58.6 & 21.8 & 60.8 & 21.5 & 55.9 & 15.3 & 73.4 & 57.8 & 62.5 & 30.7 & 71.7 & 49.9 & 64.0 & 36.6 & 65.0 & 30.6 \\
TIM & feature map & ViT-B/32 & \textbf{60.9} & 21.1 & 58.0 & 15.4 & 56.7 & 12.8 & 65.5 & 41.2 & 61.5 & 25.4 & 69.0 & 36.7 & 58.7 & 26.1 & 63.3 & 22.7 \\
k-means & feature map & ViT-B/32 & \textbf{60.9} & 21.1 & 58.0 & 15.4 & 56.7 & 12.8 & 65.5 & 41.2 & 61.5 & 25.4 & 69.0 & 36.7 & 58.7 & 26.1 & 63.3 & 22.7 \\
k-means & simplex & ViT-B/32 & 60.2 & 26.4 & 68.0 & 29.9 & 58.1 & 19.3 & 84.4 & 69.7 & 66.1 & 39.8 & 85.3 & 67.2 & 72.7 & 46.0 & 70.9 & 38.2 \\
KL k-means & simplex & ViT-B/32 & 60.0 & 26.2 & 67.9 & 29.9 & 58.1 & 19.3 & 84.3 & 69.6 & 66.0 & 39.8 & 85.3 & 67.3 & 72.6 & 45.9 & 70.8 & 38.6 \\
\rowcolor{Gray}
k-sBetas & simplex & ViT-B/32 & 60.1 & \textbf{27.2} & \textbf{69.6} & \textbf{32.8} & \textbf{58.5} & \textbf{22.3} & \textbf{85.6} & \textbf{72.3} & \textbf{67.4} & \textbf{42.8} & \textbf{85.9} & \textbf{70.3} & \textbf{75.4} & \textbf{48.9} & \textbf{71.1} & \textbf{43.6} \\
\midrule
CLIP & feature map & ViT-B/16 & 60.6 & 31.2 & 66.8 & 33.8 & 56.7 & 23.2 & 81.6 & 75.5 & 68.2 & 48.2 & 80.8 & 68.9 & 66.8 & 46.3 & 69.2 & 43.0 \\
BD-CSPN & feature map & ViT-B/16 & 59.9 & 26.2 & 64.7 & 26.6 & 56.0 & 18.9 & 71.3 & 60.2 & 65.2 & 38.5 & 76.7 & 57.1 & 62.7 & 35.5 & 66.6 & 35.1 \\
Lap-Shot & feature map & ViT-B/16 & 60.0 & 26.3 & 64.8 & 26.7 & 56.0 & 18.9 & 71.5 & 60.2 & 65.2 & 38.7 & 76.8 & 57.1 & 62.7 & 35.4 & 66.8 & 35.2 \\
TIM & feature map & ViT-B/16 & \textbf{63.3} & 26.0 & 61.4 & 18.3 & 56.9 & 15.0 & 63.8 & 44.0 & 62.7 & 28.1 & 75.2 & 45.7 & 58.0 & 26.4 & 63.8 & 25.2 \\
k-means & feature map & ViT-B/16 & \textbf{63.3} & 26.0 & 61.4 & 18.3 & 56.9 & 15.0 & 63.8 & 44.0 & 62.7 & 28.1 & 75.2 & 45.7 & 58.0 & 26.4 & 63.8 & 25.2 \\
k-means & simplex & ViT-B/16 & 61.3 & 30.2 & 71.3 & 34.3 & 61.6 & 23.7 & 82.7 & 73.3 & 70.3 & 48.9 & 87.9 & 74.5 & 70.4 & 47.4 & 76.8 & 46.7 \\
KL k-means & simplex & ViT-B/16 & 61.1 & 30.2 & 71.1 & 34.4 & 61.5 & 23.8 & 82.5 & 73.2 & 70.2 & 48.6 & 87.8 & 74.5 & 70.3 & 47.3 & 76.8 & 47.1 \\
\rowcolor{Gray}
k-sBetas & simplex & ViT-B/16 & 61.2 & \textbf{32.3} & \textbf{73.3} & \textbf{38.3} & \textbf{62.1} & \textbf{25.2} & \textbf{85.9} & \textbf{79.1} & \textbf{72.4} & \textbf{54.7} & \textbf{88.6} & \textbf{78.9} & \textbf{72.7} & \textbf{50.2} & \textbf{77.1} & \textbf{50.5} \\
                                \bottomrule
                            \end{tabular}
                            }
                        \end{sc}
                    \end{small}
                \end{center}
            \end{table*}

            \begin{table*}[t]
                \addtolength{\tabcolsep}{-3pt}
                \caption{\textbf{0-shot results.} \textbf{Query set size is 64.} The proposed \textsc{k-sBetas} further boosts the results of the existing Zero-Shot CLIP model by clustering its soft predictions on the query set.}
                \label{Zero_SHOT_clip_embedding_qs_64}
                \begin{center}
                    \begin{small}
                        \begin{sc}
                            \resizebox{\textwidth}{!}{%
                            \begin{tabular}{lcccccccccccccccccc}
                                \toprule
                                \multirow{2}{*}{Method} & \multirow{2}{*}{Layer used} & \multirow{2}{*}{Network} & \multicolumn{2}{c}{cifar-100} & \multicolumn{2}{c}{Stanford-cars} & \multicolumn{2}{c}{FGVC} & \multicolumn{2}{c}{Caltech101} & \multicolumn{2}{c}{Food101} & \multicolumn{2}{c}{Flowers102}  & \multicolumn{2}{c}{Sun397-100} & \multicolumn{2}{c}{ImageNetv2-100} \\
                                \cmidrule(r){4-5} \cmidrule(r){6-7} \cmidrule(r){8-9} \cmidrule(r){10-11} \cmidrule(r){12-13} \cmidrule(r){14-15} \cmidrule(r){16-17} \cmidrule(r){18-19} 
                                 & & & (NMI) & (Acc) & (NMI) & (Acc) & (NMI) & (Acc) & (NMI) & (Acc) & (NMI) & (Acc) & (NMI) & (Acc) & (NMI) & (Acc) & (NMI) & (Acc) \\
\midrule
CLIP & feature map & RN50 & 61.0 & 38.9 & 74.3 & 55.3 & 57.3 & 17.6 & 87.3 & 81.3 & 77.8 & 71.0 & 81.8 & 63.2 & 76.4 & 71.9 & 74.1 & 60.8 \\
BD-CSPN  & feature map & RN50 & 58.7 & 23.4 & 68.6 & 25.8 & 56.1 & 6.5 & 77.2 & 47.4 & 69.8 & 37.8 & 75.0 & 29.5 & 66.6 & 41.2 & 68.1 & 31.8 \\
Lap-Shot  & feature map & RN50 & 56.3 & 19.6 & 69.6 & 25.3 & 55.1 & 5.5 & 77.4 & 46.3 & 66.3 & 29.8 & 61.5 & 12.7 & 66.9 & 41.5 & 66.2 & 27.1 \\
TIM  & feature map & RN50 & 60.5 & 28.3 & 61.9 & 19.6 & 57.9 & 11.7 & 66.4 & 42.6 & 66.2 & 37.9 & 72.0 & 37.1 & 57.1 & 25.7 & 64.0 & 29.4 \\
k-means & feature map & RN50 & 60.5 & 28.3 & 61.9 & 19.6 & 57.9 & 11.7 & 66.4 & 42.6 & 66.2 & 37.9 & 72.0 & 37.1 & 57.1 & 25.7 & 64.0 & 29.4 \\
k-means & simplex & RN50 & 63.4 & 40.4 & 80.2 & 59.0 & 60.1 & 18.8 & 90.7 & 84.6 & 81.8 & 79.2 & 90.6 & 69.0 & 81.3 & 74.8 & 78.2 & 69.2 \\
KL k-means & simplex & RN50 & 63.4 & 40.5 & 80.2 & 59.0 & 60.0 & 18.9 & 90.8 & 84.6 & 81.8 & 79.2 & 90.5 & 68.8 & 81.4 & 75.1 & 78.2 & 69.3 \\
\rowcolor{Gray}
k-sBetas  & simplex & RN50 & \textbf{64.5} & \textbf{43.2} & \textbf{83.6} & \textbf{64.2} & \textbf{61.1} & \textbf{20.4} & \textbf{91.7} & \textbf{85.4} & \textbf{83.3} & \textbf{81.6} & \textbf{91.0} & \textbf{69.1} & \textbf{84.8} & \textbf{80.7} & \textbf{79.3} & \textbf{72.2} \\
\midrule
CLIP & feature map & RN101 & 59.5 & 42.7 & 77.9 & 62.8 & 57.4 & 18.1 & 86.1 & 83.3 & 80.0 & 74.5 & 80.1 & 61.0 & 78.4 & 71.5 & 77.3 & 67.7 \\
BD-CSPN & feature map & RN101 & 56.9 & 24.4 & 70.7 & 34.0 & 57.5 & 7.7 & 77.3 & 45.9 & 70.6 & 38.5 & 78.1 & 28.9 & 69.8 & 44.4 & 71.0 & 34.1 \\
Lap-Shot & feature map & RN101 & 53.7 & 22.6 & 72.7 & 35.3 & 56.3 & 7.1 & 78.7 & 45.0 & 70.4 & 37.4 & 78.1 & 28.0 & 70.6 & 45.3 & 72.0 & 35.1 \\
TIM & feature map & RN101 & 63.3 & 36.8 & 63.6 & 25.4 & 59.4 & 14.1 & 69.4 & 51.9 & 70.2 & 49.4 & 74.5 & 38.8 & 64.1 & 36.5 & 67.7 & 38.6 \\
k-means & feature map & RN101 & 63.3 & 36.8 & 63.6 & 25.4 & 59.4 & 14.1 & 69.4 & 51.9 & 70.2 & 49.4 & 74.5 & 38.8 & 64.1 & 36.5 & 67.7 & 38.6 \\
k-means & simplex & RN101 & \textbf{65.2} & 46.3 & 83.2 & 66.0 & 62.5 & 20.0 & 89.6 & 85.2 & 84.1 & 82.1 & 90.0 & 66.3 & 83.0 & 74.5 & 83.3 & 77.0 \\
KL k-means & simplex & RN101 & \textbf{65.2} & 46.4 & 83.2 & 66.1 & 62.5 & 20.0 & 89.6 & 85.1 & 84.0 & 82.2 & 89.9 & 66.3 & 82.8 & 74.3 & 83.3 & 77.0 \\
\rowcolor{Gray}
k-sBetas & simplex & RN101 & 64.2 & \textbf{47.6} & \textbf{86.9} & \textbf{70.5} & \textbf{63.9} & \textbf{21.7} & \textbf{91.3} & \textbf{86.8} & \textbf{84.9} & \textbf{83.3} & \textbf{90.6} & \textbf{66.7} & \textbf{85.0} & \textbf{78.0} & \textbf{84.9} & \textbf{79.6} \\
\midrule
CLIP & feature map & ViT-B/32 & 70.4 & 58.9 & 75.8 & 59.1 & 57.8 & 17.6 & 89.8 & 85.4 & 78.9 & 76.2 & 81.7 & 62.0 & 79.0 & 72.9 & 77.1 & 63.6 \\
BD-CSPN & feature map & ViT-B/32 & 65.7 & 36.8 & 68.5 & 30.2 & 58.0 & 10.5 & 78.9 & 51.0 & 69.1 & 46.5 & 77.9 & 28.3 & 68.9 & 46.0 & 70.7 & 36.3 \\
Lap-Shot & feature map & ViT-B/32 & 64.9 & 35.2 & 69.1 & 29.5 & 57.7 & 9.9 & 79.6 & 50.2 & 69.0 & 43.4 & 75.2 & 21.6 & 69.5 & 46.3 & 69.7 & 32.3 \\
TIM & feature map & ViT-B/32 & 65.0 & 38.9 & 62.2 & 22.0 & 58.1 & 12.7 & 66.8 & 44.5 & 65.5 & 42.1 & 73.6 & 38.7 & 59.1 & 30.4 & 66.2 & 33.7 \\
k-means & feature map & ViT-B/32 & 65.0 & 38.9 & 62.2 & 22.0 & 58.1 & 12.7 & 66.8 & 44.5 & 65.5 & 42.1 & 73.6 & 38.7 & 59.1 & 30.4 & 66.2 & 33.7 \\
k-means & simplex & ViT-B/32 & 74.7 & 65.3 & 81.1 & 61.2 & 60.9 & 17.8 & 92.0 & 88.0 & 84.4 & 83.6 & 91.1 & 66.4 & 83.3 & 75.9 & 83.0 & 73.0 \\
KL k-means & simplex & ViT-B/32 & 74.8 & 65.4 & 81.1 & 61.3 & 60.8 & 17.4 & 91.9 & 88.0 & 84.4 & 83.7 & 91.1 & 66.3 & 83.4 & 76.2 & 83.0 & 73.0 \\
\rowcolor{Gray}
k-sBetas & simplex & ViT-B/32 & \textbf{75.9} & \textbf{67.7} & \textbf{85.2} & \textbf{68.2} & \textbf{61.9} & \textbf{18.9} & \textbf{94.5} & \textbf{88.9} & \textbf{86.1} & \textbf{86.0} & \textbf{91.2} & \textbf{67.1} & \textbf{85.7} & \textbf{79.5} & \textbf{83.5} & \textbf{74.5} \\
\midrule
CLIP & feature map & ViT-B/16 & 71.0 & 63.8 & 79.1 & 63.7 & 61.1 & 26.0 & 91.0 & 86.2 & 85.1 & 82.9 & 83.1 & 66.4 & 79.7 & 75.6 & 79.6 & 70.5 \\
BD-CSPN & feature map & ViT-B/16 & 66.3 & 43.6 & 71.5 & 34.5 & 61.6 & 13.9 & 80.7 & 62.3 & 74.8 & 56.2 & 80.7 & 34.6 & 69.5 & 50.8 & 71.9 & 38.2 \\
Lap-Shot & feature map & ViT-B/16 & 65.7 & 40.7 & 72.9 & 33.6 & 61.6 & 12.8 & 81.4 & 61.1 & 74.9 & 54.2 & 81.5 & 31.3 & 70.0 & 51.8 & 72.7 & 36.1 \\
TIM & feature map & ViT-B/16 & 65.8 & 44.4 & 64.2 & 25.0 & 60.6 & 15.8 & 70.4 & 49.5 & 70.1 & 48.9 & 75.1 & 43.8 & 60.0 & 31.3 & 66.3 & 38.8 \\
k-means & feature map & ViT-B/16 & 65.8 & 44.4 & 64.2 & 25.0 & 60.6 & 15.8 & 70.4 & 49.5 & 70.1 & 48.9 & 75.1 & 43.8 & 60.0 & 31.3 & 66.3 & 38.8 \\
k-means & simplex & ViT-B/16 & 75.6 & 68.0 & 84.1 & 65.7 & 66.3 & 26.4 & 93.0 & 87.8 & 89.9 & 88.9 & \textbf{91.3} & 70.1 & 83.7 & 77.4 & 85.6 & 77.2 \\
KL k-means & simplex & ViT-B/16 & 75.6 & 68.1 & 84.1 & 65.8 & 66.3 & 26.6 & 92.9 & 87.8 & 89.9 & 88.9 & \textbf{91.3} & 70.1 & 83.7 & 77.4 & 85.8 & 77.2 \\
\rowcolor{Gray}
k-sBetas & simplex & ViT-B/16 & \textbf{77.3} & \textbf{71.6} & \textbf{87.3} & \textbf{70.4} & \textbf{68.2} & \textbf{28.7} & \textbf{93.9} & \textbf{88.1} & \textbf{90.8} & \textbf{90.3} & 91.2 & \textbf{70.6} & \textbf{86.9} & \textbf{82.8} & \textbf{86.5} & \textbf{79.7} \\
                                \bottomrule
                            \end{tabular}
                            }
                        \end{sc}
                    \end{small}
                \end{center}
            \end{table*}

    \subsection{Real-Time UDA for road segmentation}
    \label{subsec_real_time_UDA_road_seg}
    
        We now consider a question that was, to the best of our knowledge, hitherto unaddressed in UDA segmentation: Can we adapt predictions from a black-box source model on a target set in real-time?
        
        \textbf{Setup.} We address this question on the \textit{road GTA5$\rightarrow$road Cityscapes} challenge. 
        A source model is trained on GTA5 \cite{richter2016playing} and applied on the Cityscapes validation set \cite{Cordts_2016_CVPR}. Following \cite{vu2019advent}, we use Deeplab-V2 \cite{chen2017deeplab} as the semantic segmentation network. In our case, the model is exclusively trained for the binary task of road segmentation. The validation set contains 500 images with a 1024*2048 resolution. In order to simulate a real-time application, we treat each image independently, i.e. as a new clustering task, in which each pixel represents a point to cluster. \\
        
        \textbf{Towards Real-Time running time:} Furthermore, to maximize the speed of execution, we downsample the model's output probability map prior to fitting the clustering methods and then use the obtained densities to perform inference at the original resolution. Downsampling by a factor of 8 allows for an increase in the frame rate by almost two orders of magnitude, without incurring any loss in mIoU\footnote{See Table \ref{table_comp_time_modulo_seg} in the Appendix for more details.}. Specifically, \textsc{k-sBetas} takes on average 0.0165 seconds for clustering on each 128*256 subset of pixels and 0.0056 seconds for predictions on original 1024*2048 images, which represents a processing frequency of 45 images per second. Hardware used: \textit{CPU 11th Gen Intel(R) Core(TM) i7-11700K 3.60GHz}, and \textit{GPU NVIDIA GeForce RTX 2070 SUPER}. \\
        
        \textbf{Results.} Corresponding NMI and mean IoU scores are displayed on Table \ref{table_road_seg_uda}, and show \textsc{k-sBetas} outperforming low computational demanding competitors \textsc{k-means} and \textsc{KL k-means} by a large margin, with an improvement of 14 mIoU points over the pixel-wise inductive baseline. Fig. \ref{fig_GTA_to_Cityscapes_road_seg} provides qualitative results, in which \textsc{k-sBetas} yields a definite visual improvement over the baseline.
        In addition, and consistently with the previous results observed in Table \ref{table_bal_scores} for the UDA classification challenge, Table \ref{table_road_seg_uda} also shows that clustering the probability simplex points may be more effective than clustering the logits.
            \begin{table}[t]
                \addtolength{\tabcolsep}{-3pt}
                \caption{\textbf{Real-Time UDA} for road segmentation  per image on the challenge GTA5$\rightarrow$Cityscapes, by using probability simplex clustering.}
                \label{table_road_seg_uda}
                \begin{center}
                \begin{small}
                \begin{sc}
                \resizebox{0.65\columnwidth}{!}{%
                \begin{tabular}{lcc}
                    \toprule
                    \multirow{2}{*}{Approach} & \multicolumn{2}{c}{GTA5$\rightarrow$Cityscapes} \\
                     \cmidrule(r){2-3}
                     & (NMI) & (mIoU) \\
                    \midrule
                    k-means - logits & 21.4 & 39.4 \\
                    \midrule
                    argmax & 19.7 & 49.2 \\ 
                    k-means & 23.6 & 52.4 \\ 
                    KL k-means & 24.9 & 52.2 \\ 
                    \rowcolor{Gray}
                    k-sBetas & \textcolor{black}{\textbf{35.8}} & \textcolor{black}{\textbf{65.7}} \\ 
                    \bottomrule
                \end{tabular}
                }
                \end{sc}
                \end{small}
                \end{center}
            \end{table}

    \subsection{Ablation studies}
    \label{subsec_ablation_study}

        \textbf{Joint v.s. Factorised density.} Our proposed method draws inspiration from a factorized $\mathtt{Beta}$ density, i.e. each component of the simplex vector is considered independent from each other. While such a simplifying assumption has significant analytical and computational benefits, it may also fail to capture the complexity of the target distribution. We explore this trade-off by comparing \textsc{k-Betas} with \textsc{k-Dirs}, in which the full joint $\mathtt{Beta}$ density -- corresponding to a Dirichlet density -- is fitted at each iteration. Because there exists no closed-form solution to this problem, we resort to the approximate estimation procedure described in \cite{minka2000estimating}. Results in Table \ref{table_bal_scores} show that \textsc{k-Dirs} and \textsc{k-Betas} produce similar scores on mixtures of Dirichlet distributions. In the meantime, \textsc{k-Dirs} outperforms \textsc{k-Betas} on SVHN$\rightarrow$MNIST but fails to converge on the more challenging VISDA-C challenge, making it ill-suited to real-world applications. \\

        \textbf{Effect of class imbalance.} To investigate the problem of class imbalance, we generate heavily imbalanced datasets. First, we create an imbalanced version of our synthetic \textit{Simu} dataset, which we refer to as \textit{iSimu}. Specifically, we weighted the 3 components of the \textit{Simu} mixture with six different combinations using the imbalanced class proportions $\{0.75, 0.2, 0.05\}$. Second, we consider a variant of \textit{VISDA-C}, where class-proportions are sampled from a Dirichlet distribution with $\bm{\alpha}=\bm{1}_{K}$. We refer to this variant as \textit{iVISDA-Cs}. We perform 10 random re-runs. 
        
        Table \ref{table_imb_scores} compares the clustering models on these two datasets. Displayed NMI scores for \textit{iSimus} correspond to the average NMI scores obtained across the six different mixture proportions. NMI and IoU scores displayed for iVISDA-Cs correspond to the average scores obtained across ten different highly imbalanced subset variants of VISDA-C. \textsc{k-sBetas (biased)} refers to the proposed approach without the marginal probability term $\pi_k$ in eq. \ref{eq_unbiased_ksbetas}. These comparative results clearly highlight the benefit of the proposed \textsc{k-sBetas} formulation. Note that our unbiased formulation could theoretically apply to metric-based approaches but was systematically found to produce degenerated solutions in which all examples are assigned to a unique cluster. \\
            \begin{table}[t]
                \caption{Comparative \textbf{probability simplex clustering} on \textbf{highly imbalanced} \textit{iSimus} and iVISDA-Cs datasets.}
                \label{table_imb_scores}
                \vspace{-0.015\textwidth}
                \begin{center}
                    \begin{sc}
                    \resizebox{0.75\columnwidth}{!}{%
                    \begin{tabular}{lccc}
                    \toprule
                     & \multicolumn{1}{c}{iSimus} & \multicolumn{2}{c}{iVISDA-Cs} \\
                     \cmidrule(r){2-2} \cmidrule(r){3-4}
                    Approach & (NMI) & (NMI) & (mIoU) \\
                    \midrule
                    argmax   &  55.5 & 31.6 & 22.7 \\ 
                    k-means   &  62.3 & 32.8 & 24.2 \\
                    GMM &  64.5 & 34.4 & 21.1 \\ 
                    k-medians   &  60.4 & 33.1 & 22.4 \\ 
                    k-medoids   &  62.6 & 32.1 & 22.5 \\ 
                    k-modes   &  55.1 & 32.0 & 22.8 \\ 
                    KL k-means  &  59.9 & 35.2 & 24.9 \\ 
                    HSC   & 17.7 & 28.9 & 16.3 \\
                    \rowcolor{Gray}
                    k-sBetas (biased) &  55.3 & 35.4 & 25.6 \\ 
                    \rowcolor{Gray}
                    k-sBetas & 72.4 & \textcolor{black}{\textbf{36.4}} & \textcolor{black}{\textbf{27.1}} \\
                    \bottomrule
                    \end{tabular}
                    }
                    \end{sc}
                \end{center}
            \end{table}
        
            \begin{table}[t]
                \addtolength{\tabcolsep}{-3pt}
                \caption{Accuracy depending on the unimodal constraint. The constraint is enabled in $\checkmark$ columns. k-Dirs parameter estimation \cite{minka2000estimating} fails to converge on VISDA-C.}
                \label{tab_unimodal_constraint}
                \vspace{-0.015\textwidth}
                \begin{center}
                \begin{small}
                \begin{sc}
                \resizebox{0.7\columnwidth}{!}{%
                \begin{tabular}{lcccc}
                    \toprule
                     & \multicolumn{2}{c}{SVHN$\rightarrow$MNIST} & \multicolumn{2}{c}{VISDA-C} \\
                    \cmidrule(r){2-3} \cmidrule(r){4-5}
                    Unimodal constraint & $\times$ & $\checkmark$ & $\times$ & $\checkmark$ \\
                    \midrule
                    k-Dirs & 65.3 & \textbf{68.8} & fails & fails \\
                    k-Betas  & 19.5 & \textbf{51.5} & 11.9 & \textbf{19.8} \\ 
                    \rowcolor{Gray}
                    k-sBetas (MLE-500)  & \textbf{76.2} & \textbf{76.2} & \textbf{55.0} & \textbf{55.0} \\
                    \rowcolor{Gray}
                    k-sBetas & \textbf{76.6} & \textbf{76.6} & \textbf{56.0} & \textbf{56.0} \\ 
                    \bottomrule
                \end{tabular}
                }
                \end{sc}
                \end{small}
                \end{center}
                \vskip -0.1in
            \end{table}        

        \textbf{Parameter estimation.} Concerning the parameter estimation for \textsc{k-sBetas}, we can observe in Table \ref{table_comp_time} and Table \ref{tab_unimodal_constraint} that using the method of moments (MoM) is more beneficial than the iterative MLE in practice, both in terms of computational cost and prediction performances. \\

        \textbf{Effect of the unimodal constraint.} Table \ref{tab_unimodal_constraint} empirically confirms that constraining a density-based clustering model to only consider mixtures of unimodal distributions is appropriate with softmax predictions from pre-trained source models. Yet, we can observe that disabling this constraint has no impact on \textsc{k-sBetas} results. The following paragraph explains such observations by the presence of $\delta$. \\
        
        \textbf{Effect of $\delta$.} As a matter of fact, Table \ref{tab_unimodal_constraint} results suggest that $\delta$ interest is finally twofold in our experiments: It enables a more flexible clustering. Meanwhile, it also prevents estimating bimodal density functions.
        The latter point can be explained as follows. Bimodal distributions have a higher variance than uniform or unimodal distributions because they partition the major portion of the density at the vicinity of two opposite interval boundaries simultaneously. Thus, when $\delta$ is sufficiently large (e.g. with $\delta=0.15$ in these experiments), the scaled variance of the projection of $\bm{x}_{i}$ coordinates into the interval $[\frac{\delta}{1+2\delta},\frac{1+\delta}{1+2\delta}]$ becomes sufficiently small to inhibit the modeling of bimodal distributions. Complementary, Fig. \ref{fig_sBeta_depending_on_delta_and_iters} shows that setting $\delta=0.15$ provides consistent outperforming results on different datasets, and it allows fast convergence in terms of clustering iterations. Thus, we set $\delta=0.15$ along all the other presented experiments. \\
        
            \begin{figure}%
                \centering
                \includegraphics[width=\columnwidth]{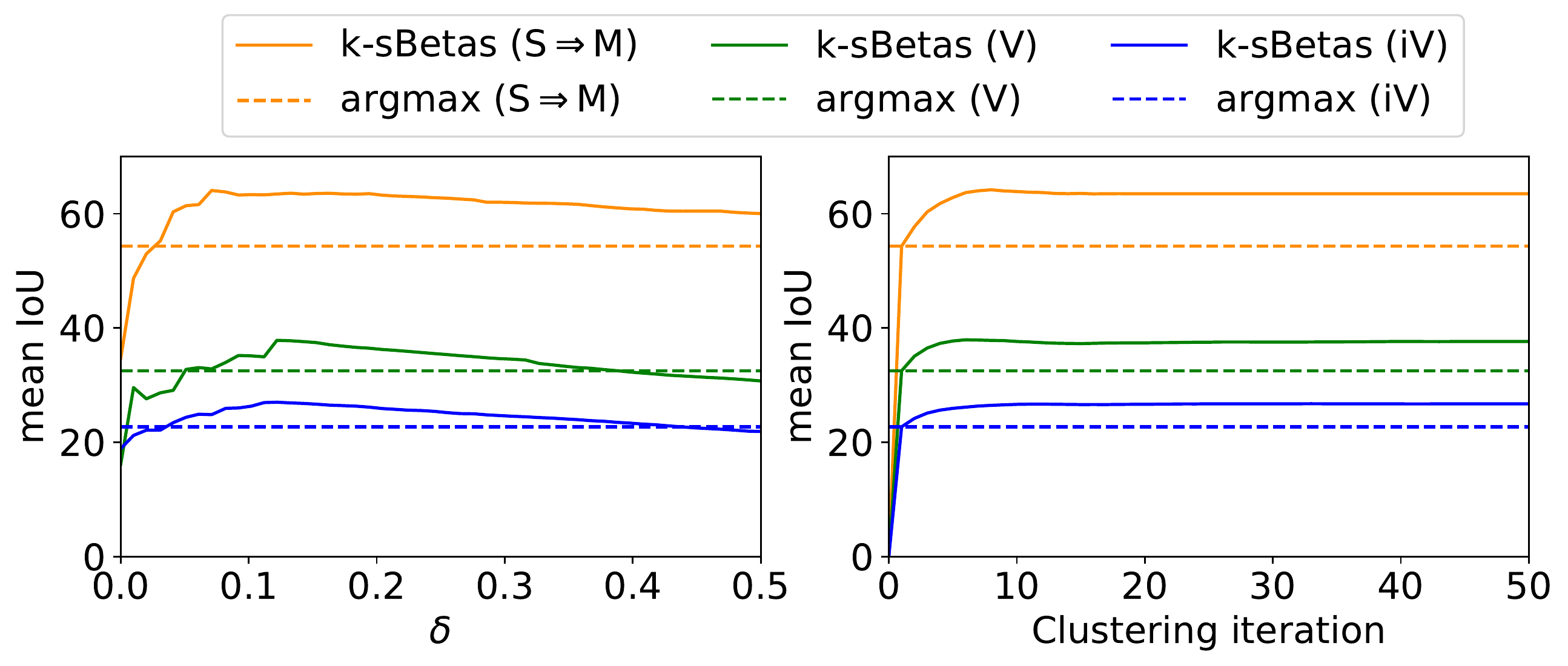}%
                \centering
                \vspace{-0.015\textwidth}
                \caption{\textsc{k-sBetas} mean IoU scores depending on $\delta$ and on the clustering iteration. We respectively set 25 clustering iterations on the left figure and $\delta=0.15$ on the right one. S$\rightarrow$M, V, and iV respectively refer to SVHN$\rightarrow$MNIST, VISDA-C, and iVISDA-Cs datasets.}%
                \label{fig_sBeta_depending_on_delta_and_iters}%
            \end{figure}
        
        \textbf{Pre-clustering: Centroid initialization.} Parameter initialization plays a crucial role in unsupervised clustering. In Table \ref{tab:initialization}, we compare the widely-used k-means++ initialization, designed for generic clustering, with our simplex-tailored vertex initialization. Regardless of the assignment method used, our initialization demonstrates up to a $7\%$ improvement in accuracy over k-means++\footnote{Table \ref{table_kmeansplusplus_init_vs_vertices_init} in the Appendix further extends this comparison to other clustering approaches, showing similar benefits across every tested approach.}.
        \\

        \textbf{Post-clustering: Cluster to class assignment.} We also show on Table \ref{tab:initialization} the interest of the Hungarian algorithm, referred to as Hung, for aligning cluster-labels with class-labels. Compared to the argmax assignment for centroids in the context of such closed-set challenges, we recall that Hung aims to ensure that each cluster is assigned to a separate class. We can observe that the proposed Hung strategy is particularly interesting when using the vertex init. This suggests that optimal transport assignment is naturally relevant in situations where each cluster is likely to represent a separate class.
        
        In order to provide a fair comparison with the state-of-the-art, we jointly applied these \textit{pre-} and \textit{post-}clustering strategies on every compared approach.
        
            \begin{table}[t]
                \addtolength{\tabcolsep}{-3pt}
                \caption{\textsc{k-sBetas} prediction performances depending on the parameter initialization and cluster to class assignment. We compare k-means++ with the proposed \textit{vertex init}, and also the centroid assignment argmax with the proposed optimal transport strategy (Hung).}
                \label{tab:initialization}
                \vspace{-0.015\textwidth}
                \begin{center}
                \begin{small}
                \begin{sc}
                \resizebox{\columnwidth}{!}{%
                \begin{tabular}{lcccc}
                    \toprule
                     &  & & SVHN$\rightarrow$MNIST & VISDA-C \\
                    \cmidrule(r){4-4} \cmidrule(r){5-5}
                    Approach & Init & assignment & (Acc) & (Acc) \\
                    \midrule
                    \multirow{4}{*}{k-sBetas} & k-mean++ & argmax & 69.0  $\pm$  6.1 & 50.0  $\pm$  5.0 \\
                     & k-means++ & hung & 69.8  $\pm$  8.1 & 47.2  $\pm$  5.2 \\
                     & vertex init & argmax & 73.5 $\pm$  0.0 & 53.8  $\pm$  0.0 \\
                     & vertex init & Hung & \textbf{76.6  $\pm$  0.0} & \textbf{56.0 $\pm$ 0.0} \\ 
                    \bottomrule
                \end{tabular}
                }
                \end{sc}
                \end{small}
                \end{center}
                \vskip -0.1in
            \end{table}

\section{Discussion}
\label{sec_Discussion}

This section discusses the potential limitations and extensions of the proposed approach. \\

\textbf{Maximum performance may be upper-bounded.} We have shown that clustering softmax predictions with the proposed approach can efficiently improve source model prediction performances at a reasonable computational footprint. However, it is worth noting that we could not reach the prediction performances of the recent state-of-the-art self-training \cite{liang2021source}, which exploits and updates the model parameters end-to-end through an iterative process. This is because, as for every clustering algorithm, the maximum performance of the proposed approach is upper-bounded by the quality of the output probability simplex domain representation, over which we have no control. \\

\textbf{Spatial analysis could be complementary.}
The clustering model does not consider global or local spatial information. That would be worthwhile in challenges such as semantic pixel-wise segmentation to complement it with spatial post-processing techniques \cite{badrinarayanan2017segnet}. \\

\textbf{Clustering softmax predictions for self-training.} 
Throughout this article, we have motivated the use of simplex clustering for efficient prediction adjustment of black-box source models. Nevertheless, simplex clustering could also play the role of pseudo-labeling along self-training strategies. This could be a simple and light generic alternative to feature map clustering \cite{caron2018deep}, which would not imply cumbersome manipulation of high-dimensional and hidden feature maps. \\

\textbf{If the number of classes is large.} In some contexts, the number of classes $K$ may be potentially large. For example, K may correspond to millions of words in neural language modeling. It would be then interesting to envision strategies that reorganize the softmax layer for more efficient calculation \cite{chen-etal-2016-strategies}. In addition, distortion-based and probabilistic clustering methods usually require a sufficient number of points per class to produce consistent partitioning. Thus, if $K$ is too large with respect to the target set size, then the clustering could produce degenerate solutions. It would be interesting to explore hierarchical clustering strategies \cite{murtagh2012algorithms, ahalya2015data} on the probability simplex domain, as these models can deal with a small amount of data points per class. \\

\textbf{Low-powered devices.} Systems such as mobile robots and embedded systems may not have the resources to perform costly neural network parameter updates on the application time. The presented framework can be viewed as an efficient plug-in solution for prediction adjustment in the wild when using such low-powered devices.

\section{Conclusion}
\label{sec_Conclusion}

In this paper, we have tackled the simplex clustering of softmax predictions from black-box deep networks. We found that existing distortion-based objectives of the state-of-the-art do not adequately approximate real-world simplex distributions, which can, for example, be highly peaked near the simplex vertices. Thus, we have introduced a novel probabilistic clustering objective that integrates a generalization of the $\mathtt{Beta}$ density, which we coin $\mathtt{sBeta}$. This scaled variant of $\mathtt{Beta}$ is relatively more permissive; meanwhile, we constrain it to fit only strictly unimodal distributions and to avoid degenerate solutions. In order to optimize our clustering objective, we proceed with a block-coordinate descent approach, which alternates optimization w.r.t assignment variables and parameter estimation of cluster distributions, s.t. the introduced parameter constraints. The resulting clustering model, which we refer to as \textsc{k-sBetas}, is both highly competitive and efficient on the proposed simulated and real-world softmax-prediction datasets, including class-imbalance scenarios, along which we performed our comparisons. Additionally, we emphasized the utility of probability simplex clustering through several practical applications: zero-shot and one-shot image classification, and real-time adjustment for road segmentation on full-size images.

Our future perspectives include extending the insights of probability simplex clustering to other challenges, such as pseudo-labeling for self-training procedures and other applications that require rapid correction of source models. Moreover, it may be interesting to complement the clustering with spatial and temporal information depending on the target data properties.

Overall, we hope that the presented simplex clustering framework and softmax-prediction benchmarks will encourage the community to give more attention to such facilitating, low-computational demanding, model-agnostic, plug-in solutions.

\ifCLASSOPTIONcompsoc
  \section*{Acknowledgments}
\else
  \section*{Acknowledgment}
\fi

This research was supported by computational resources provided by Compute Canada.

\ifCLASSOPTIONcaptionsoff
  \newpage
\fi



\bibliographystyle{IEEEtran}
\bibliography{main}
%



%

\begin{IEEEbiography}[{\includegraphics[width=1in,height=2in,clip,keepaspectratio]{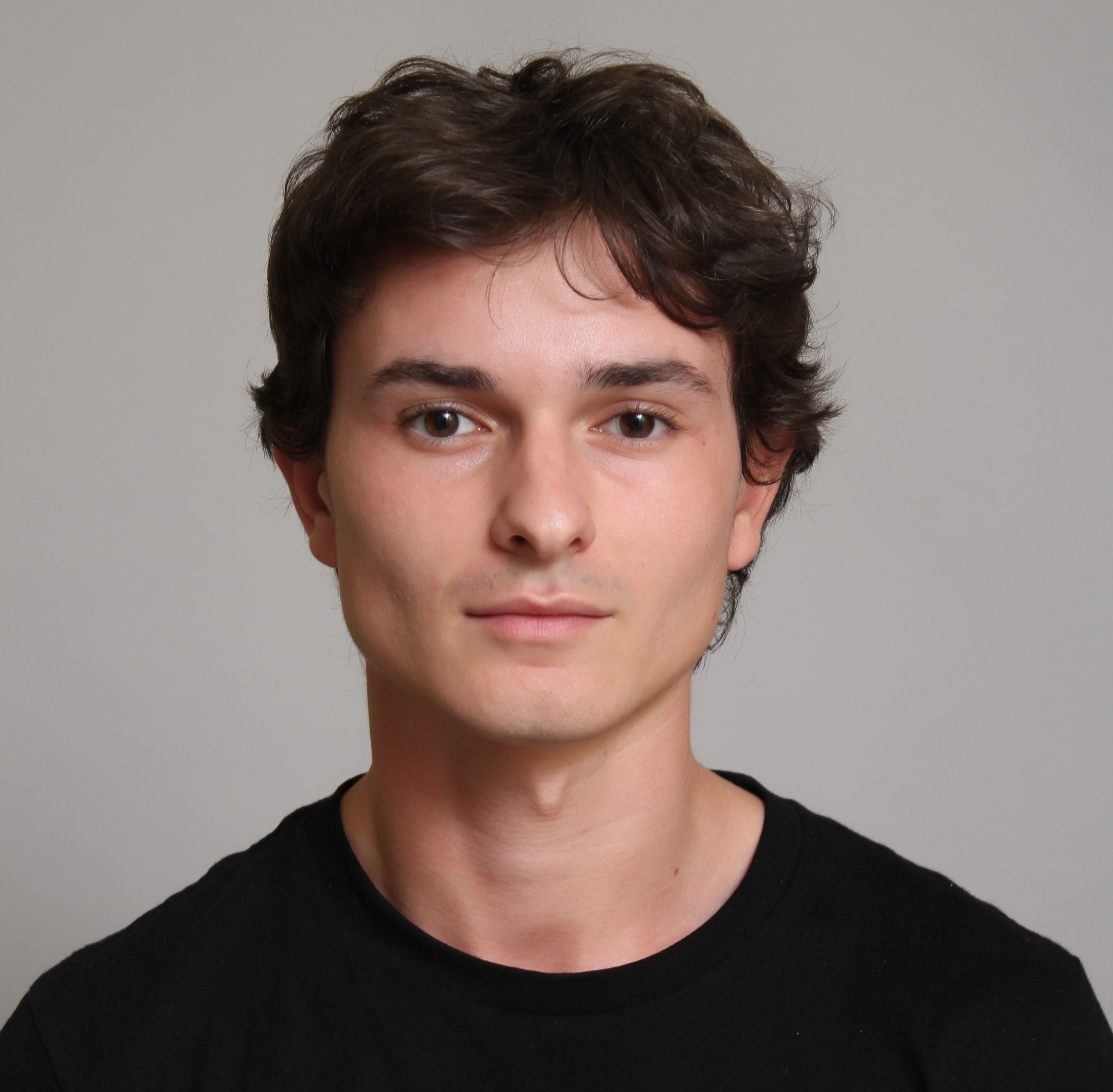}}]{Florent} \textbf{Chiaroni} received his Dipl-Ing degree in computer science and electronics from ESTIA,
France, in 2016, and his MSc in robotics and embedded systems from the University of Salford Manchester, United Kingdom, in the same year. He earned his Ph.D. in signal and image processing from the University of Paris Saclay in 2020, with VEDECOM Institute, Versailles, France, and Université Paris-Saclay, Centre national de la recherche scientifique (CNRS), CentraleSupélec, Gif-Sur-Yvette, France. He during the course of this research, was a Postdoctoral Fellow at \'ETS Montreal, Canada and Institut National de la Recherche Scientifique (INRS), Montreal, Canada. He is currently a Research Scientist at Thales Research and Technology (TRT) Canada, Thales Digital Solutions (TDS), CortAIx, Montreal, Canada. His current research interests include civil unmanned aerial vehicles (UAVs) and efficient weakly-supervised learning for visual pattern analysis.
\end{IEEEbiography}
\vspace{-2mm} 

\begin{IEEEbiography}[{\includegraphics[width=1in,height=2in,clip,keepaspectratio]{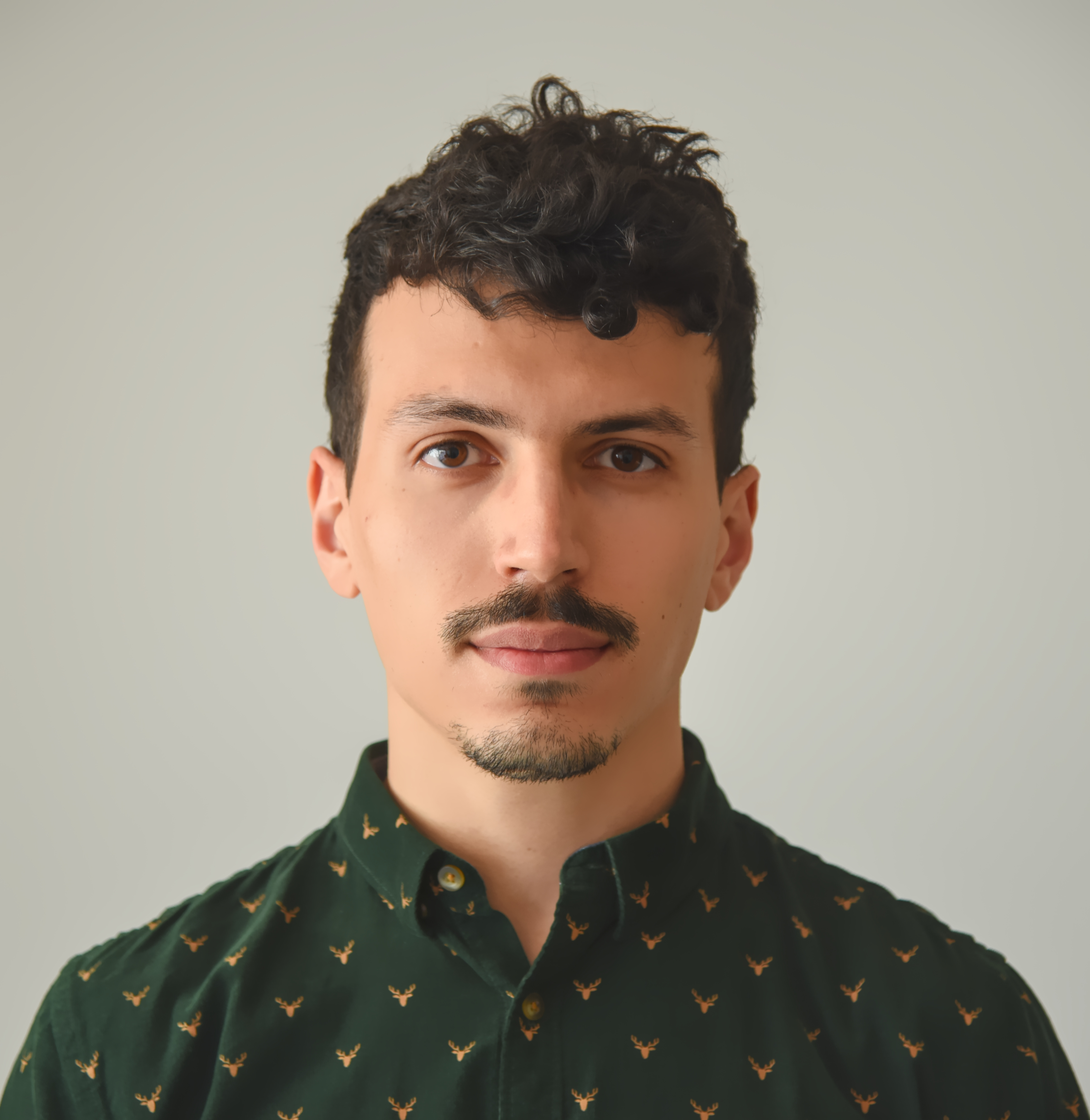}}]{Malik} \textbf{Boudiaf} obtained his MSc in Aeronautics \& Astronautics from Stanford University in 2019, and his M.Eng in Aerospace Engineering from ISAE-Supaero, France in 2017, and his PhD degree from ÉTS Montréal, Canada, supervised by Prof. Ismail Ben Ayed and Prof. Pablo Piantanida. His research lies between Computer Vision, Information Theory, Optimization, and their application to few-shot/unsupervised learning. He is currently a Senior Machine Learning Engineer in Computer Vision and Natural Language Processing at Stealth Startup, Montreal, Canada.
\end{IEEEbiography}
\vspace{-2mm} 

\begin{IEEEbiography}[{\includegraphics[width=1in,height=2in,clip,keepaspectratio]{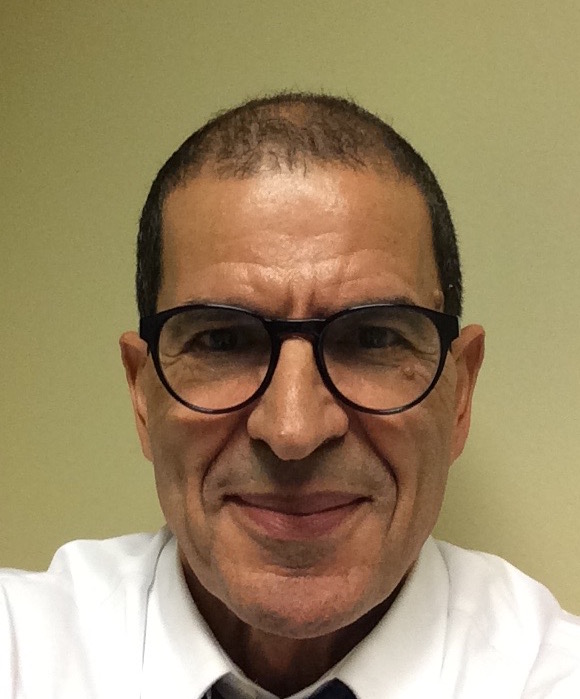}}]{Amar Mitiche} holds the Licence Es Sciences degree in mathematics from the University of
Algiers and the Ph.D. degree in computer science from the University of Texas at Austin. He is
currently a Professor with the Department of Telecommunications (INRS-EMT), Institut
National de la Recherche Scientifique (INRS), Montreal, QC, Canada. His research is in
computer vision and pattern recognition. He has written several articles on the subjects, as well
as three books: Computational Analysis of Visual Motion (Plenum Press, 1994), Variational and
Level Set Methods in Image Segmentation (Springer, 2011), with Ismail Ben Ayed, and
Computer Vision Analysis of Image Motion by Variational Methods (Springer, 2014), with J. K.
Aggarwal. His current interests include image segmentation, image motion analysis, and
pattern classification by neural networks.
\end{IEEEbiography}
\vspace{-2mm} 

\begin{IEEEbiography}[{\includegraphics[width=1in,height=2in,clip,keepaspectratio]{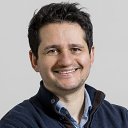}}]{Ismail Ben Ayed} is currently a Full Professor at \'ETS Montreal. He is also affiliated with the CRCHUM. His interests are in computer vision, optimization, machine learning, and medical image analysis algorithms. Ismail authored over 100 fully peer-reviewed papers, mostly published in the top venues of those areas, along with 2 books and 7 US patents.  In recent years, he gave over 30 invited talks, including 4 tutorials at flagship conferences (\textsc{Miccai}'14, \textsc{Isbi}'16, \textsc{Miccai}'19 and \textsc{Miccai}'20). His research has been covered in several visible media outlets, such as Radio Canada (CBC), Quebec Science Magazine and Canal du Savoir. His research team received several recent distinctions, such as the \textsc{Midl}'19 best paper runner-up award and several top-ranking positions in internationally visible contests. Ismail served as Program Committee for \textsc{Miccai}'15, \textsc{Miccai}'17 and \textsc{Miccai}'19, and as Program Chair for \textsc{Midl}'20. Also, he regularly serves as a reviewer for the main scientific journals of his field and was selected several times among the top reviewers of prestigious conferences (such as CVPR'15 and \textsc{NeurIPS}'20).
\end{IEEEbiography}






\newpage
\clearpage
\appendices
\section{sBeta complementary details}
\label{subsec_all_properties_demos}

We provide in this section the demonstrations for the mean, the variance, the mode, and parameter estimation of the presented $\mathtt{sBeta}$ density function. Table \ref{table_properties_Beta_sBeta} summarizes the resulting properties.

\begin{table*}[h]
\caption{$\mathtt{sBeta}$ and related $\mathtt{Beta}$ properties depending on their respective probability density function.}
\label{table_properties_Beta_sBeta}
\vskip 0.15in
\begin{center}
\begin{small}
\begin{sc}
\resizebox{0.95\textwidth}{!}{%
\begin{tabular}{l|c|c}
\toprule
 & sBeta & Beta \\
\midrule
Density function & $\displaystyle f_{\mathtt{sBeta}} = \frac{(x+\delta)^{\alpha-1} (1+\delta-x)^{\beta-1}}{B(\alpha,\beta)(1+2\delta)^{\alpha+\beta-2}}$ & $\displaystyle f_{\mathtt{Beta}} = \frac{x^{\alpha-1} {(1-x)}^{\beta-1}}{B(\alpha,\beta)}$ \\
\midrule
mean (Sec. \ref{subsubsec_mean_and_variance}) & $\displaystyle E_{sB}[X]=\frac{\alpha}{\alpha+\beta}(1+2\delta) - \delta$ & $\displaystyle E_{B}[X]=\frac{\alpha}{(\alpha+\beta)}$ \\
\midrule
variance (Sec. \ref{subsubsec_mean_and_variance}) & $\displaystyle V_{sB}[X]=\frac{\alpha\beta}{(\alpha+\beta)^2(\alpha+\beta+1)}(1+2\delta)^2$ & $\displaystyle V_{B}[X]=\frac{\alpha \beta}{(\alpha+\beta)^2(\alpha+\beta+1)}$ \\
\midrule
mode (Sec. \ref{subsec_mode_and_concentration}) & $\displaystyle m_{sB} = \frac{\alpha-1 + \delta(\alpha-\beta)}{\alpha + \beta - 2}$ & $\displaystyle m_B = \frac{\alpha-1}{\alpha+\beta-2}$ \\
\midrule
$\begin{aligned} &\textrm{concentration} \\ &\textrm{(Sec. \ref{subsubsec_mode_concentration})}\end{aligned}$ & $\displaystyle \lambda = \alpha + \beta - 2$ & $\displaystyle \lambda = \alpha + \beta - 2$ \\
\midrule
$\begin{aligned} &\textrm{method of moments} \\ &\textrm{(Sec. \ref{subsec_clustering_algo})}\end{aligned}$
& $\displaystyle \begin{array}{ll}
    &\alpha = ( \frac{\displaystyle \mu_{\delta}(1-\mu_{\delta})(1+2\delta)^2}{V_{sB}[X]}-1 ) \mu_{\delta} \\ \\
    &\beta = ( \frac{\displaystyle \mu_{\delta}(1-\mu_{\delta})(1+2\delta)^2}{V_{sB}[X]}-1 ) (1-\mu_{\delta}), \\ \\
    &\textrm{with } \mu_{\delta}=\frac{\displaystyle E_{sB}[X]+\delta}{1+2\delta}
  \end{array}$ 
& $\displaystyle \begin{array}{ll}
    &\alpha = ( \frac{\displaystyle E_{B}[X](1-E_{B}[X])}{V_{B}[X]}-1 ) E_{B}[X] \\ \\
    &\beta  = ( \frac{\displaystyle E_{B}[X](1-E_{B}[X])}{V_{B}[X]}-1 )(1-E_{B}[X])
  \end{array}$ \\
\bottomrule
\end{tabular}
}
\end{sc}
\end{small}
\end{center}
\vskip -0.1in
\end{table*}

\subsection{Mean and Variance}
    \label{subsubsec_mean_and_variance}
    
    The mean of the probability density function $f(x)$ is defined as $E[X]=\int x f(x) \,dx$. Meanwhile, the variance of $f(x)$ can be defined as $V=E[X^2]-{E[X]}^2$ with $E[X^2]=\int x^2 f(x) \,dx$.
    Based on these two statements, we propose to estimate the mean and the variance of the proposed $\mathtt{sBeta}$ density function $f_{\mathtt{sBeta}}$.
    
    \textbf{sBeta mean:} To express the first moment $E_{sB}[X]$ of $\mathtt{sBeta}$, as a function its parameters, we propose to replace $x$ with $(x+\delta-\delta)$ as follow:
    
    \begin{equation}
        \begin{aligned}
            & E_{sB}[X] = \int x f_{\mathtt{sBeta}}(x) \,dx \\
                 =& \int x \frac{(x+\delta)^{\alpha-1} (1+\delta-x)^{\beta-1}}{B(\alpha,\beta)(1+2\delta)^{\alpha+\beta-2}} \,dx \\
                 =& \int \frac{\overbrace{(x+\delta-\delta)}^{\textrm{replaces $x$}} (x+\delta)^{\alpha-1} (1+\delta-x)^{\beta-1}}{B(\alpha,\beta)(1+2\delta)^{\alpha+\beta-2}}   \,dx \\
                 =& \frac{1}{B(\alpha,\beta)(1+2\delta)^{\alpha+\beta-2}} \times \\
                 & \int  (x+\delta)^{(\alpha+1)-1} (1+\delta-x)^{\beta-1} \\
                 & - \delta(x+\delta)^{\alpha-1} (1+\delta-x)^{\beta-1} \,dx \\
                 =& \frac{1}{B(\alpha,\beta)(1+2\delta)^{\alpha+\beta-2}} \times \\
                 & \int (x+\delta)^{(\alpha+1)-1} (1+\delta-x)^{\beta-1} \,dx \\
                 & - \delta \underbrace{\int \frac{(x+\delta)^{\alpha-1} (1+\delta-x)^{\beta-1}}{B(\alpha,\beta)(1+2\delta)^{\alpha+\beta-2}} \,dx}_{=f_{\mathtt{sBeta}} \textrm{, so it integrates to } 1} \\
                 =& \frac{1}{B(\alpha,\beta)(1+2\delta)^{\alpha+\beta-2}} \times \\
                 & \int (x+\delta)^{(\alpha+1)-1} (1+\delta-x)^{\beta-1} \,dx - \delta \\
                 =& \frac{B(\alpha+1,\beta)(1+2\delta)^{(\alpha+1)+\beta-2}}{B(\alpha,\beta)(1+2\delta)^{\alpha+\beta-2}} \times \\
                 & \underbrace{\int \frac{(x+\delta)^{(\alpha+1)-1} (1+\delta-x)^{\beta-1}}{B(\alpha+1,\beta)(1+2\delta)^{(\alpha+1)+\beta-2}} \,dx}_{=1} - \delta \\
                 =& \frac{B(\alpha+1,\beta)(1+2\delta)}{B(\alpha,\beta)} - \delta, \\
        \end{aligned}
        \label{eq_integrated_sBeta_mean}
    \end{equation}
    with $B(z_1,z_2)=\frac{\Gamma(z_1)\Gamma(z_2)}{\Gamma(z_1+z_2)}$. Then, with respect to the property $\Gamma(z+1)=z\Gamma(z)$, we have:
    \begin{equation}
        \begin{aligned}
                E_{sB}[X] =& \frac{\Gamma(\alpha+1)\Gamma(\beta)\Gamma(\alpha+\beta)}{\Gamma(\alpha+1+\beta)\Gamma(\alpha)\Gamma(\beta)}(1+2\delta) - \delta \\
                 =& \frac{\Gamma(\alpha+1)\Gamma(\alpha+\beta)}{\Gamma(\alpha+1+\beta)\Gamma(\alpha)}(1+2\delta) - \delta \\
                 =& \frac{\alpha\Gamma(\alpha) \Gamma(\alpha+\beta)}{(\alpha+\beta)\Gamma(\alpha+\beta)\Gamma(\alpha)}(1+2\delta) - \delta \\
                 =& \frac{\alpha}{\alpha+\beta}(1+2\delta) - \delta.
        \end{aligned}
        \label{eq_integrated_sBeta_mean_suite}
    \end{equation}

    \textbf{sBeta variance:} We first estimate the second moment $E_{sB}[X^2]$ by replacing $x^2$ with $(x+\delta-\delta)^2=(x+\delta)^2-2\delta(x+\delta)+\delta^2$ as follow:
    
    \begin{equation}
        \begin{aligned}
            E_{sB}&[X^2] = \int x^2 f_{\mathtt{sBeta}}(x) \,dx \\
                   =& \int (x+\delta-\delta)^2 f_{\mathtt{sBeta}}(x) \,dx \\
                   =& \int ((x+\delta)^2-2\delta(x+\delta)+\delta^2) f_{\mathtt{sBeta}}(x) \,dx \\
                   =& \int (x+\delta)^2 f_{\mathtt{sBeta}}(x) \,dx -2\delta \int (x+\delta)f_{\mathtt{sBeta}}(x) \,dx \\
                   & +\delta^2 \underbrace{\int f_{\mathtt{sBeta}}(x) \,dx}_{=1} \\
                   =& \underbrace{\frac{(\alpha+1)\alpha}{(\alpha+1+\beta)(\alpha+\beta)}(1+2\delta)^2}_{\textrm{Using Eq. \eqref{eq_first_term}}} -\underbrace{2\delta \frac{\alpha}{\alpha+\beta}(1+2\delta)}_{\textrm{Using Eq. \eqref{eq_second_term}}} +\delta^2.
        \end{aligned}
        \label{eq_second_order_moment_sBeta}
    \end{equation}
    
    The term $\int (x+\delta)^2 f_{\mathtt{sBeta}}(x) \,dx$ from Eq. \eqref{eq_second_order_moment_sBeta} can be developed as: 
    \begin{equation}
        \begin{aligned}
            \int (x+\delta)^2 & f_{\mathtt{sBeta}}(x) \,dx \\
            =& \int (x+\delta)^2 \frac{(x+\delta)^{\alpha-1} (1+\delta-x)^{\beta-1}}{B(\alpha,\beta)(1+2\delta)^{\alpha+\beta-2}} \,dx \\
            =& \frac{1}{B(\alpha,\beta)(1+2\delta)^{\alpha+\beta-2}} \times \\
            & \int (x+\delta)^{(\alpha+2)-1} (1+\delta-x)^{\beta-1} \,dx \\
            =& \frac{B(\alpha+2,\beta)(1+2\delta)^{(\alpha+2)+\beta-2}}{B(\alpha,\beta)(1+2\delta)^{\alpha+\beta-2}} \times \\
            & \underbrace{\int \frac{(x+\delta)^{(\alpha+2)-1} (1+\delta-x)^{\beta-1}}{B(\alpha+2,\beta)(1+2\delta)^{(\alpha+2)+\beta-2}} \,dx}_{=1} \\
            =& \frac{B(\alpha+2,\beta)(1+2\delta)^{(\alpha+\beta-2)+2}}{B(\alpha,\beta)(1+2\delta)^{\alpha+\beta-2}} \\
            =& B(\alpha+2,\beta)\frac{1}{B(\alpha,\beta)}(1+2\delta)^2 \\
            =& \frac{\Gamma(\alpha+2)\Gamma(\beta)}{\Gamma(\alpha+2+\beta)}\frac{\Gamma(\alpha+\beta)}{\Gamma(\alpha)\Gamma(\beta)}(1+2\delta)^2 \\
            =& \frac{(\alpha+1)\alpha\Gamma(\alpha)\Gamma(\beta)\Gamma(\alpha+\beta)}{(\alpha+1+\beta)(\alpha+\beta)\Gamma(\alpha+\beta)\Gamma(\alpha)\Gamma(\beta)}(1+2\delta)^2 \\
            =& \frac{(\alpha+1)\alpha}{(\alpha+1+\beta)(\alpha+\beta)}(1+2\delta)^2.
        \end{aligned}
        \label{eq_first_term}
    \end{equation}
    Similarly, the term $-2\delta \int (x+\delta)f_{\mathtt{sBeta}}(x) \,dx$ from Eq. \eqref{eq_second_order_moment_sBeta} can be developed as:
    \begin{equation}
        \begin{aligned}
            -2\delta \int & (x+\delta) f_{\mathtt{sBeta}}(x) \,dx \\ 
            &= -2\delta \underbrace{\int \frac{(x+\delta)^{(\alpha+1)-1} (1+\delta-x)^{\beta-1}}{B(\alpha,\beta)(1+2\delta)^{\alpha+\beta-2}} \,dx}_{=E_{sB}[X]+\delta=\frac{\alpha}{\alpha+\beta}(1+2\delta) \textrm{ using Eq. \eqref{eq_integrated_sBeta_mean}}} \\
            &= -2\delta \frac{\alpha}{\alpha+\beta}(1+2\delta).
        \end{aligned}
        \label{eq_second_term}
    \end{equation}

    Finally, we can estimate the variance (i.e., the second central moment) $V_{sB}[X]$, as a function of $\mathtt{sBeta}$ density parameters, using Eq. \eqref{eq_integrated_sBeta_mean} and Eq. \eqref{eq_second_order_moment_sBeta} as follow:
    \begin{equation}
        \begin{aligned}
            V_{sB}[X] =& E_{sB}[X^2] - E_{sB}[X]^2 \\
                 =& \frac{(\alpha+1)\alpha}{(\alpha+1+\beta)(\alpha+\beta)}(1+2\delta)^2 \\
                 & -2\delta \frac{\alpha}{\alpha+\beta}(1+2\delta) +\delta^2 \\
                 & - \left[ \frac{\alpha}{\alpha+\beta}(1+2\delta) - \delta \right]^2 \\
                 =& \frac{\alpha\beta}{(\alpha+\beta)^2(\alpha+\beta+1)}(1+2\delta)^2.
        \end{aligned}
    \end{equation}

    \textbf{Linear property:} With respect to the expectation linearity, it is worth noting that we have $\frac{E[X]+\delta}{1+2\delta} = E\left[ \frac{X+\delta}{1+2\delta} \right]$ and
    \begin{equation}
        \begin{aligned}
            V\left[\frac{X+\delta}{1+2\delta}\right] 
                                            &= E \left[ \left( \frac{X+\delta}{1+2\delta} - E \left[ \frac{X+\delta}{1+2\delta} \right] \right)^2 \right] \\
                                            &= \frac{E \left[ (X - E[X])^2 \right]}{(1+2\delta)^2} \\
                                            &= \frac{V[X]}{(1+2\delta)^2}.
        \end{aligned}
    \label{eq_var_equiv_using_expectation_linearity_rule}
    \end{equation} \\

\subsection{Mode}\label{subsec_mode_and_concentration}

    The mode of a density function $f(x)$ corresponds to the value of $x$ at which $f(x)$ achieves its maximum. \\
    
    \textbf{sBeta mode:}
    To find the mode of $\mathtt{sBeta}$ depending on $\alpha$ and $\beta$ parameters, we estimate at which value of $x$ the derivative of $f_{\mathtt{sBeta}}$ is equal to $0$. We first estimate the derivative as follows:
    
    \begin{equation}
        \begin{aligned}
            & f_{\mathtt{sBeta}}'(x) = \frac{d f_{\mathtt{sBeta}}(x)}{d x}
                                   = \frac{d }{d x} \frac{(x+\delta)^{\alpha-1}(1+\delta-x)^{\beta-1}}{B(\alpha,\beta)(1+2\delta)^{\alpha+\beta-2}} \\
                                   =& \frac{1}{B(\alpha,\beta)(1+2\delta)^{\alpha+\beta-2}}\frac{d }{d x}(x+\delta)^{\alpha-1}(1+\delta-x)^{\beta-1} \\
                                   =& \frac{(\alpha-1)(x+\delta)^{\alpha-2}(1+\delta-x)^{\beta-1}}{B(\alpha,\beta)(1+2\delta)^{\alpha+\beta-2}} \\
                                   & - \frac{(x+\delta)^{\alpha-1}(\beta-1)(1+\delta-x)^{\beta-2}}{B(\alpha,\beta)(1+2\delta)^{\alpha+\beta-2}}.
        \end{aligned}
    \end{equation} \\
    
    Then, we can solve $\frac{d f_{\mathtt{sBeta}}(x)}{d x}=0$ as follow:
    
    \begin{equation}
        \begin{aligned}
            &\frac{d f_{\mathtt{sBeta}}(x)}{d x} = 0 \\
            \Rightarrow& (\alpha-1)(x+\delta)^{\alpha-2}(1+\delta-x)^{\beta-1} \\
            &- (x+\delta)^{\alpha-1}(\beta-1)(1+\delta-x)^{\beta-2} = 0 \\
            \Rightarrow& (\alpha-1)(1+\delta-x) - (\beta-1)(x+\delta) = 0 \\
            \Rightarrow& x = \frac{\alpha-1 + \delta(\alpha-\beta)}{\alpha + \beta - 2}.
        \end{aligned}
    \end{equation} \\
    
    Thus, the mode of $\mathtt{sBeta}$ with $\alpha,\beta>1$ is defined as $m_{sB} = \frac{\alpha-1 + \delta(\alpha-\beta)}{\alpha + \beta - 2}$. \\

\subsection{Solving the necessary conditions for minimizing the objective w.r.t sBeta parameters} \label{subsec_mle}
    
    Let us compute the partial derivatives of our objective $L_{\mathtt{sBeta}}$ \eqref{eq_unbiased_ksbetas} for $\alpha_{k,n}$ and $\beta_{k,n}$:
    
    \begin{align} \label{eq:partial_derivatives}
        \begin{split}
            \frac{\partial L_{\mathtt{sBeta}}}{\partial \alpha_{k,n}} =& \sum_{i=1}^N u_{i,k} \log(x_{i,n} + \delta) - (\sum_{i=1}^N u_{i,k}) \log(1+2\delta) \\
            & - (\sum_{i=1}^N u_{i,k})(\psi(\alpha_{k,n}) - \psi(\alpha_{k,n}+\beta_{k,n})) \\
            \frac{\partial L_{\mathtt{sBeta}}}{\partial \beta_{k,n}} =& \sum_{i=1}^N u_{i,k}\log(1+\delta - x_{i,n}) - (\sum_{i=1}^N u_{i,k}) \log(1+2\delta) \\
            & - (\sum_{i=1}^N u_{i,k})(\psi(\beta_{k,n}) - \psi(\alpha_{k,n}+\beta_{k,n}))    
        \end{split}
    \end{align}
    where $\psi$ stands for the di-gamma function. Setting partial derivatives in \eqref{eq:partial_derivatives} to 0 leads to the following coupled system:

    \begin{align} \label{eq:coupled_equations}
        \begin{split}
            \alpha_{k,n} =& \psi^{-1}(\psi(\alpha_{k,n}+\beta_{k,n}) \\
                          &+ \frac{1}{\sum_{i=1}^N u_{i,k}}\sum_{i=1}^N u_{i,k} \log(\frac{x_{i,n} +\delta}{1+2\delta})) \\
        \beta_{k,n} =& \psi^{-1}(\psi(\alpha_{k,n}+\beta_{k,n}) \\
                     &+ \frac{1}{\sum_{i=1}^N u_{i,k}}\sum_{i=1}^N u_{i,k} \log(\frac{1+\delta - x_{i,n}}{1+2\delta}))
        \end{split}
    \end{align}
    
    While \eqref{eq:coupled_equations} does not have any analytical solution, we approximate updates in \eqref{eq:coupled_equations} by fixing the right-hand side of each equation, leading the following vectorized updates of parameters $\bm{\alpha}_k$ and $\bm{\beta}_k$ for each cluster $k$:
    
    \begin{align}
        \bm{\alpha}_k^{(t+1)} =& \psi^{-1}(\psi(\bm{\alpha}_k^{(t)}+\bm{\beta}_k^{(t)}) \\ 
        &+ \frac{1}{\sum_{i=1}^N u_{i,k}}\sum_{i=1}^N u_{i,k} \log(\frac{\bm{x}_i +\delta}{1+2\delta})) \\
        \bm{\beta}_k^{(t+1)} =& \psi^{-1}(\psi(\bm{\alpha}_k^{(t)}+\bm{\beta}_k^{(t)}) \\ 
        &+ \frac{1}{\sum_{i=1}^N u_{i,k}}\sum_{i=1}^N u_{i,k} \log(\frac{1+\delta - \bm{x}_i}{1+2\delta})) \nonumber
    \end{align}
    
    We empirically validated such optimization procedure converged to the MLE estimate on the synthetically generated dataset.

\clearpage
\section{Supplementary experiments}
\begin{figure}[h]%
\centering
\vskip -0.05in
\includegraphics[width=0.99\columnwidth]{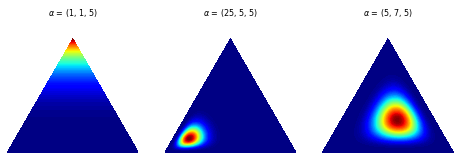}%
\vskip -0.1in
\centering
\caption{Visualization on $\Delta^2$ of the three Dirichlet distributions used to generate random artificial datasets \textit{Simu} and \textit{iSimus}, introduced in Sec. \ref{sec_Experiments}.}%
\vskip -0.15in
\label{fig_three_simul_diri_distr}%
\end{figure}
\begin{table}[h]
\caption{Balanced clustering on \textit{Simu} depending on the dataset size.}
\label{table_bal_simu}
\begin{center}
\begin{small}
\begin{sc}
\vskip -0.15in
\resizebox{0.99\columnwidth}{!}{%
\begin{tabular}{lcccc}
\toprule
Dataset size & 100 000 & 10 000 & 1 000 & 100 \\
Approach & (NMI) & (NMI) & (NMI) & (NMI) \\
\midrule
argmax & 60.1  $\pm$  0.5 & 59.8  $\pm$0.7 & 60.4  $\pm$3.9  & 62.0$\pm$17.9\\ 
k-means & 76.6  $\pm$  0.1  &  76.7  $\pm$0.5 & 76.9  $\pm$6.2 & 78.2  $\pm$21.8\\ 
KL k-means & 76.2  $\pm$  0.2 &  76.3  $\pm$0.7 & 76.5  $\pm$6.2 & 78.1  $\pm$22.8\\ 
GMM & 75.8  $\pm$  0.3  &  75.8  $\pm$1.2 & 75.9  $\pm$6.8 & 77.3  $\pm$28.4\\ 
k-medians  & 76.8  $\pm$  0.1 &  77.1  $\pm$0.7 & 77.1  $\pm$5.6 & 78.7  $\pm$22.2\\ 
k-medoids  & 60.8  $\pm$  14.4 & 64.2  $\pm$  12.1 & 65.3  $\pm$  19.4 & 71.8  $\pm$  24.2 \\ 
k-modes  & 76.2  $\pm$  0.2 &  76.3  $\pm$1.0 & 76.5  $\pm$5.3 & 77.5  $\pm$24.9\\ 
HSC & 9.3  $\pm$  1.9 & 15.0  $\pm$  2.7 & 30.6  $\pm$  16.3 & 41.2  $\pm$  27.9 \\ 
\rowcolor{Gray}
k-sBetas & \textcolor{black}{\textbf{79.5$\pm$0.1}}  &  \textcolor{black}{\textbf{79.5$\pm$0.8}} & \textcolor{black}{\textbf{79.8$\pm$6.4}} & \textcolor{black}{\textbf{80.0$\pm$25.4}} \\ 
\bottomrule
\end{tabular}
}
\end{sc}
\end{small}
\end{center}
\vskip -0.15in
\end{table}

\begin{table}[h]
\caption{Imbalanced clustering on \textit{iSimus} depending on the dataset size.}
\label{table_imb_simu}
\begin{center}
\begin{small}
\begin{sc}
\vskip -0.15in
\resizebox{0.99\columnwidth}{!}{%
\begin{tabular}{lccc}
\toprule
Dataset size & 100 000 & 10 000 & 1 000 \\
Approach & (NMI) & (NMI) & (NMI) \\
\midrule
argmax   &  55.5  $\pm$  25.5 & 55.6  $\pm$  26.8 & 55.6  $\pm$  31.5 \\ 
k-means  &  62.3  $\pm$  22.6 & 62.2  $\pm$  23.4 & 62.4  $\pm$  28.9 \\
KL k-means  &  59.9  $\pm$  25.2 & 59.9  $\pm$  26.0 & 60.2  $\pm$  31.5 \\ 
GMM   &  60.6  $\pm$  23.9 & 63.8  $\pm$  29.3 & 63.9  $\pm$  35.0 \\ 
k-medians   &  60.4  $\pm$  24.8 & 60.3  $\pm$  25.6 & 60.3  $\pm$  32.2 \\ 
k-medoids   &  47.2  $\pm$  33.0 & 55.4  $\pm$  30.0 & 57.8  $\pm$  35.9 \\ 
k-modes   &  55.1  $\pm$  30.9 & 54.9  $\pm$  32.4 & 54.8  $\pm$  36.5 \\ 
HSC   & 17.7  $\pm$  12.1 & 13.6  $\pm$  11.7 & 29.1  $\pm$  31.8 \\ 
\rowcolor{Gray}
k-sBetas   &  \textcolor{black}{\textbf{72.4$\pm$17.2}} & \textcolor{black}{\textbf{72.2$\pm$20.1}} & \textcolor{black}{\textbf{73.3$\pm$29.2}} \\ 
\bottomrule
\end{tabular}
}
\end{sc}
\end{small}
\end{center}
\vskip -0.1in
\end{table}

\begin{table}[H]
\addtolength{\tabcolsep}{-3pt}
\caption{\textbf{Running time} of \textsc{k-sBetas} (GPU-based) on \textit{GTA5 road $\rightarrow$Cityscapes road} depending on the subset size. This GPU-based implementation uses the pytorch library. We fixed to 10 the maximum number of \textsc{k-sBetas} clustering iterations. We assume that the segmentation model and \textsc{k-sBetas} method run on the same GPU, i.e., no loading time is implied by the softmax prediction set. The CPU used is 11th Gen Intel(R) Core(TM) i7-11700K 3.60GHz. GPU used: NVIDIA GeForce RTX 2070 SUPER. Presented prediction scores were all obtained on the full-size (1024*2048) set.}
\label{table_comp_time_modulo_seg}
\begin{center}
\begin{small}
\begin{sc}
\vskip -0.1in
\resizebox{0.90\columnwidth}{!}{%
\begin{tabular}{lcccc}
\toprule
  & \multicolumn{2}{c}{Running time (seconds)} & \multicolumn{2}{|c}{Full size Scores}\\
Subset size & (Clustering) & (Prediction) & (NMI) & (mIoU) \\
\midrule
Full size & \multirow{2}{*}{0.2784} &  \multirow{2}{*}{0.0056} & \multirow{2}{*}{35.8} & \multirow{2}{*}{65.7} \\ 
(1024*2048) \\
\midrule
Modulo 2 & \multirow{2}{*}{0.0812} & \multirow{2}{*}{0.0019}  & \multirow{2}{*}{35.8} & \multirow{2}{*}{65.7} \\
(512*1024) \\
\midrule
Modulo 4 & \multirow{2}{*}{0.0282} & \multirow{2}{*}{0.0009} & \multirow{2}{*}{35.8} & \multirow{2}{*}{65.7} \\ 
(256*512) \\
\midrule
Modulo 8 & \multirow{2}{*}{0.0165} & \multirow{2}{*}{0.0004} & \multirow{2}{*}{35.8} & \multirow{2}{*}{65.7}  \\
(128*256) \\
\bottomrule
\end{tabular}
}
\end{sc}
\end{small}
\end{center}
\vskip -0.15in
\end{table}

\newpage
\begin{table}[h]
    \caption{\textbf{k-means++} versus \textbf{vertex init} (proposed). Balanced clustering on softmax predictions. Performances are averaged over ten executions. Note that k-means++ is stochastic while the vertex init is not.}
    \label{table_kmeansplusplus_init_vs_vertices_init}
    \begin{center}
        \begin{small}
        \begin{sc}
        \vskip -0.15in
        \resizebox{0.99\columnwidth}{!}{%
        \begin{tabular}{lcccc}
        \toprule
        Dataset & \multicolumn{2}{c}{SVHN$\rightarrow$ MNIST} & \multicolumn{2}{c}{VISDA-C} \\
         \cmidrule(r){2-3} \cmidrule(r){4-5}
        Initialization & k-means++ & vertex init & k-means++ & vertex init \\
        Approach & (Acc) & (Acc) & (Acc) & (Acc) \\
        \midrule
        k-means   &  66.6  $\pm$  5.4 &  \textcolor{black}{\textbf{68.9  $\pm$  0.0}} & 44.5  $\pm$  3.3 & \textcolor{black}{\textbf{47.9 $\pm$ 0.0}} \\ 
        KL k-means  &  72.6  $\pm$  7.8 &  \textcolor{black}{\textbf{75.5  $\pm$  0.0}} & 50.0  $\pm$  2.7 & \textcolor{black}{\textbf{51.2 $\pm$ 0.0}} \\ 
        GMM   &  60.6  $\pm$  9.6 &  \textcolor{black}{\textbf{69.2  $\pm$  0.0}} & 43.8  $\pm$  5.2 & \textcolor{black}{\textbf{49.4 $\pm$ 0.0}} \\
        k-medians   &  67.4  $\pm$  9.2 & \textcolor{black}{\textbf{68.8  $\pm$  0.0}} & 38.5  $\pm$  4.0 & \textcolor{black}{\textbf{40.0 $\pm$ 0.0}} \\ 
        k-medoids   &  51.9  $\pm$  4.9 &  \textcolor{black}{\textbf{71.3 $\pm$  0.0}} & 40.6  $\pm$  8.6 & \textcolor{black}{\textbf{46.8 $\pm$  0.0}} \\ 
        k-modes   &  60.6  $\pm$  13.0 &  \textcolor{black}{\textbf{71.3  $\pm$  0.0}} & 30.2  $\pm$  6.3 & \textcolor{black}{\textbf{31.1 $\pm$ 0.0}} \\ 
        \rowcolor{Gray}
        k-sBetas   &  69.8  $\pm$  8.1 &  \textcolor{black}{\textbf{76.6  $\pm$  0.0}} & 47.2  $\pm$  5.2 & \textcolor{black}{\textbf{56.0 $\pm$ 0.0}} \\ 
        \bottomrule
        \end{tabular}
        }
        \end{sc}
        \end{small}
    \end{center}
    \vskip -0.1in
\end{table}

    \subsubsection{One-Shot Learning}
    \label{subsec_one_shot}

            \begin{table}[H]
                \addtolength{\tabcolsep}{-3pt}
                \caption{\textbf{1-shot plug-in results.} The proposed \textsc{k-sBetas} further boosts results of existing One-Shot methods by clustering their soft predictions on the query set.}
                \vskip -0.15in
                \label{one_SHOTv3}
                \begin{center}
                    \begin{small}
                        \begin{sc}
                            \resizebox{\columnwidth}{!}{%
                            \begin{tabular}{lccccc}
                                \toprule
                                \multirow{2}{*}{Method} & \multirow{2}{*}{Network} & \multicolumn{2}{c}{\textit{mini}ImageNet} & \multicolumn{2}{c}{\textit{tiered}ImageNet}  \\
                                \cmidrule(r){3-6}
                                 & & (NMI) & (Acc) & (NMI) & (Acc) \\
                                \midrule
                                SimpleSHOT & RN-18 & 49.1 & 62.7 & 57.5 & 69.2 \\
                                \rowcolor{Gray}
                                SimpleSHOT \textbf{+ k-sBetas} & RN-18 & \textcolor{black}{\textbf{52.2}} & \textcolor{black}{\textbf{64.4}} & \textcolor{black}{\textbf{60.5}} & \textcolor{black}{\textbf{71.0}} \\
                                \midrule
                                BD-CSPN & RN-18 & 58.2 & 68.9 & 67.2 & 76.0 \\
                                \rowcolor{Gray}
                                BD-CSPN \textbf{+ k-sBetas} & RN-18 & \textcolor{black}{\textbf{60.5}} & \textcolor{black}{\textbf{69.8}} & \textcolor{black}{\textbf{68.9}} & \textcolor{black}{\textbf{76.6}} \\
                                \midrule
                                SimpleSHOT & WRN-28-10 & 53.6 & 65.7 & 59.1 & 70.4 \\
                                \rowcolor{Gray}
                                SimpleSHOT \textbf{+ k-sBetas} & WRN-28-10 & \textcolor{black}{\textbf{56.8}} & \textcolor{black}{\textbf{67.3}} & \textcolor{black}{\textbf{61.8}} & \textcolor{black}{\textbf{72.4}} \\
                                \midrule
                                BD-CSPN & WRN-28-10 & 62.4 & 72.1 & 69.0 & 77.5 \\
                                \rowcolor{Gray}
                                BD-CSPN \textbf{+ k-sBetas} & WRN-28-10 & \textcolor{black}{\textbf{64.0}} & \textcolor{black}{\textbf{72.4}} & \textcolor{black}{\textbf{70.7}} & \textcolor{black}{\textbf{78.3}} \\
                                \bottomrule
                            \end{tabular}
                            }
                        \end{sc}
                    \end{small}
                \end{center}
                \vskip -0.1in
            \end{table}

        We consider the One-Shot classification problem, in which a model is evaluated based on its ability to generalize to new classes from a single labeled example per class. Typically, One-Shot methods use the labeled support set $\mathcal{S}$ of each task to build a classifier and obtain soft predictions for unlabelled query samples. Once soft predictions have been obtained, proposed \textsc{k-sBetas} can be used to further refine predictions by clustering soft predictions of the entire query set. \\
        
        \textbf{Setup.} We use two standard benchmarks for One-Shot classification: \textit{mini}-Imagenet \cite{mini_imagenet} and \textit{tiered}-Imagenet \cite{tiered_imagenet}. The \textit{mini}-Imagenet benchmark is composed of 60,000 color images \cite{mini_imagenet} equally split among 100 classes, themselves split between train, val, and test following \cite{ravi2016optimization}. The \textit{tiered}-Imagenet benchmark is a larger dataset with 779,165 images and 608 classes, split following \cite{tiered_imagenet}. All images are resized to $84 \times 84$. Regarding the networks, we use the pre-trained RN-18 \cite{resnet} and WRN28-10 \cite{zagoruyko2016wide} provided by \cite{boudiaf2020information}. Only 15 unlabeled query data points per class are available during each separate task, so we use the biased version of \textsc{k-sBetas}. As for the methods, we select one inductive method: SimpleSHOT \cite{wang2019simpleshot} and one transductive method: BD-CSPN \cite{liu2020prototype}. Each One-Shot method is reproduced using the set of hyperparameters suggested in the original papers. \\
        
        \textbf{Results.} Table \ref{one_SHOTv3} shows that all along these experiments, \textsc{k-sBetas} consistently improves in terms of NMI and Accuracy scores the output predictions of SimpleSHOT and BD-CSPN.

\begin{table*}[t]
\addtolength{\tabcolsep}{-3pt}
\caption{\textbf{iVISDA-Cs imbalanced proportions.} To obtain 10 imbalanced subsets of VISDA-C that we refer to \textit{iVISDA-Cs}, we have randomly generated ten random vectors from a Dirichlet distribution. VISDA-C contains 12 classes, so the dimension of this Dirichlet distribution is equal to 12. Dirichlet parameters are set as $\bm{\alpha}=\{\alpha_n=1\}_{n=1}^{D=12}$, such that this dirichlet distribution is uniform on $\Delta^{11}$. Below is the list of the ten sets of rounded proportions obtained and the corresponding number of examples for each VISDA-C class. For every class, examples from the original dataset are always picked in the same order, starting at the beginning of the target image list of VISDA-C.}
\label{table_props_iVISDA_C}
\begin{center}
\begin{small}
\begin{sc}
\vskip -0.1in
\resizebox{0.88\textwidth}{!}{%
\begin{tabular}{lcccccccccccc}
\toprule
Classes  & plane & bcycl & bus & car & horse & knife & mcycl & person & plant & sktbrd & train & truck \\
\midrule
VISDA-C proportions & 0.0658 & 0.0627 & 0.0847 & 0.1878 & 0.0847 & 0.0375 & 0.1046 & 0.0722 & 0.0821 & 0.0412 & 0.0765 & 0.1002 \\
Number of examples & 3646 & 3475 & 4690 & 10401 & 4691 & 2075 & 5796 & 4000 & 4549 & 2281 & 4236 & 5548 \\
\midrule
Proportions set 1  &  0.0406 & 0.2315 & 0.0953 & 0.0257 & 0.2367 & 0.0443 &
  0.0930 & 0.0252 & 0.1347 & 0.0601 & 0.0012 & 0.0116 \\
Number of examples & 275 & 1568 & 645 & 173 & 1603 & 300 & 630 & 170 & 912 & 407 & 8  & 78 \\
\midrule
Proportions set 2 & 0.0996 & 0.0234 & 0.0086 & 0.0112 & 0.1490 & 0.0064
 & 0.1900 & 0.1223 & 0.1287 & 0.0387 & 0.0876 & 0.1346 \\
Number of examples & 675 & 158 &  58  & 75 & 1009 &  43 & 1286 & 828 & 871 & 262 & 593 & 912 \\
\midrule
Proportions set 3 & 0.0511 & 0.1170 & 0.0779 & 0.0661 & 0.0127 & 0.0575 & 0.0166 & 0.1958 & 0.0047 & 0.0795 & 0.2147 & 0.1064 \\
Number of examples & 346 & 792 & 527 & 448 &  85 & 389 & 112 & 1326  & 31 & 538 & 1454 & 720 \\
\midrule
Proportions set 4 & 0.1382 & 0.0014 & 0.1726 & 0.0017 & 0.0030 &  0.1600 & 0.2470 & 0.0726 & 0.0419 & 0.0524 & 0.0558 & 0.0534 \\
Number of examples & 936 & 9 & 1169 &  11 &  20 & 1084 & 1673 & 492 &  284 & 354 & 378 & 361\\
\midrule
Proportions set 5 & 0.0013 & 0.0227 & 0.1451 & 0.1049 & 0.2725 & 0.0511 & 0.0196 & 0.0500 & 0.1401 & 0.0193 & 0.0322 & 0.1411 \\
Number of examples & 8 & 154 & 983 & 710 & 1846 & 346 & 132 & 338 & 948 & 130 & 218 & 956\\
\midrule
Proportions set 6 & 0.1450 & 0.0202 & 0.0343 & 0.1604 & 0.0574 & 0.0183 & 0.0318 & 0.0368 & 0.2668 & 0.0356 & 0.1361 & 0.0570\\
Number of examples & 982 & 137 & 232 & 1086 & 388 & 123 & 215 & 249 & 1807 & 241 & 922 & 386\\
\midrule
Proportions set 7 & 0.1200 & 0.1304 & 0.0310 & 0.0451 & 0.0308 & 0.0071 & 0.1194 & 0.2701 & 0.0122 & 0.1311 & 0.0911 & 0.0116\\
Number of examples & 812 & 883 & 209 & 305 & 208 &  48 & 809 & 1830  & 82 & 888 & 617 & 78\\
\midrule
Proportions set 8 & 0.0166 & 0.0521 & 0.0457 & 0.0358 & 0.1175 & 0.1917 & 0.0258 & 0.2303 & 0.1503 & 0.0379 & 0.0600 & 0.0362\\
Number of examples & 112 & 352 & 309 & 242 & 796 & 1298 & 174 & 1560 & 1018 & 256 & 406 & 245\\
\midrule
Proportions set 9 & 0.1523 & 0.1316 & 0.0196 & 0.0427 & 0.0380 & 0.0806 & 0.0423 & 0.2075 & 0.0383 & 0.0644 & 0.1074 & 0.0754\\
Number of examples & 1031 & 891 & 132 & 289 & 257 & 546 & 286 & 1405  & 259 & 436 & 727 & 510\\
\midrule
Proportions set 10 & 0.0102 & 0.1336 & 0.0063 & 0.0131 & 0.1291   & 0.0812 & 0.0019 & 0.0967 & 0.3063 & 0.0770 & 0.1107 & 0.0337\\
Number of examples & 69 & 905 &  42 &  88 & 874 & 549 &  12 & 655 & 2074 & 521 & 750 & 228\\
\bottomrule
\end{tabular}
}
\end{sc}
\end{small}
\end{center}
\vskip -0.1in
\end{table*}

\begin{figure*}[b]%
\centering
\includegraphics[width=1.\textwidth]{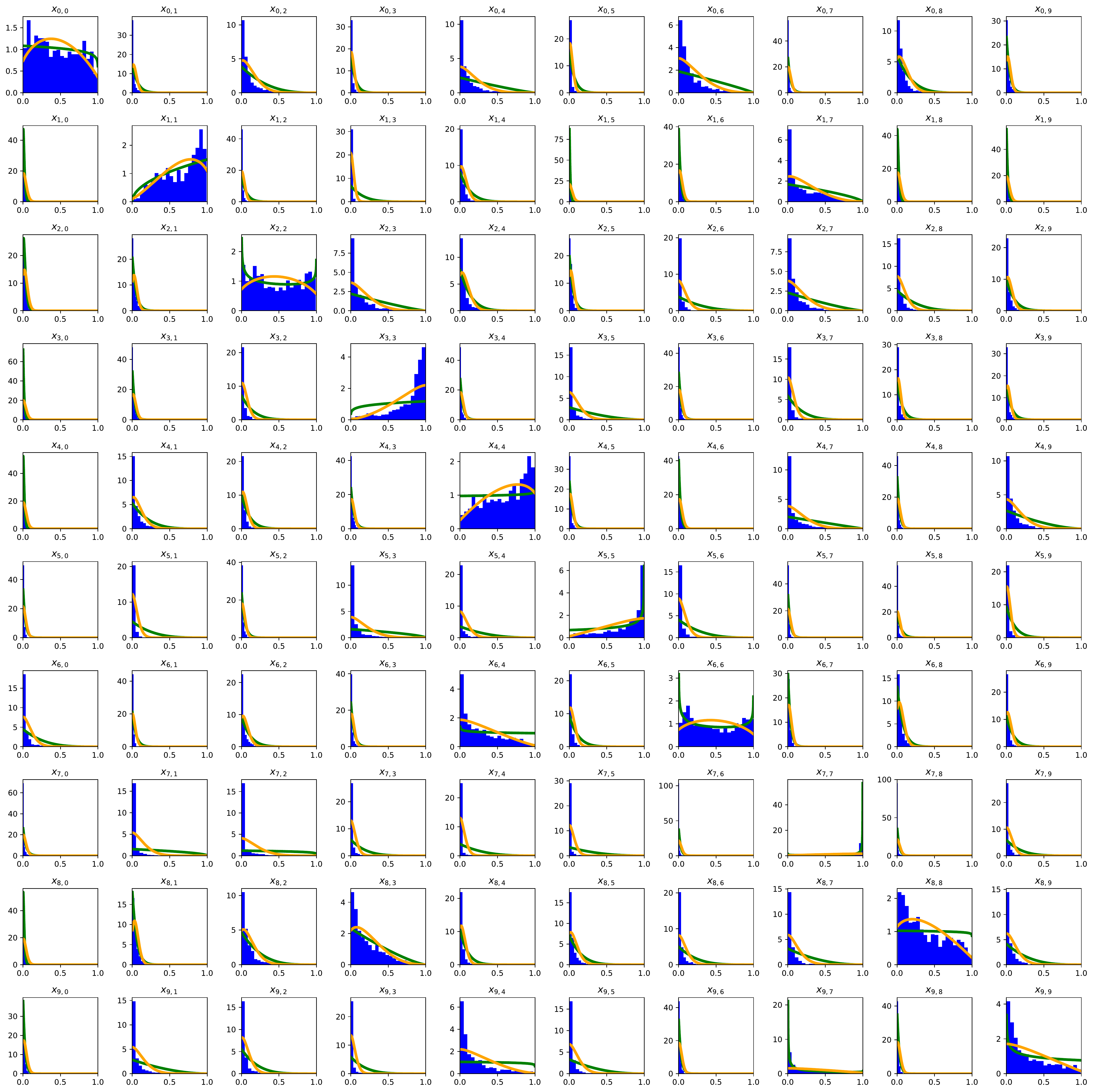}%
\centering
\caption{Histograms per coordinate of softmax predictions obtained on MNIST, with a source model pre-trained on SVHN. Rows correspond to softmax predictions for a given class. Columns refer to the softmax prediction coordinates. Green curves represent $\mathtt{Beta}$ density estimations, and orange curves represent $\mathtt{sBeta}$ density estimations. \textit{We invite the reader on the pdf version to zoom on these vectorized figures.}}%
\label{fig_S_to_M_hists}%
\end{figure*}

\begin{figure*}[b]%
\centering
\includegraphics[width=1.\textwidth]{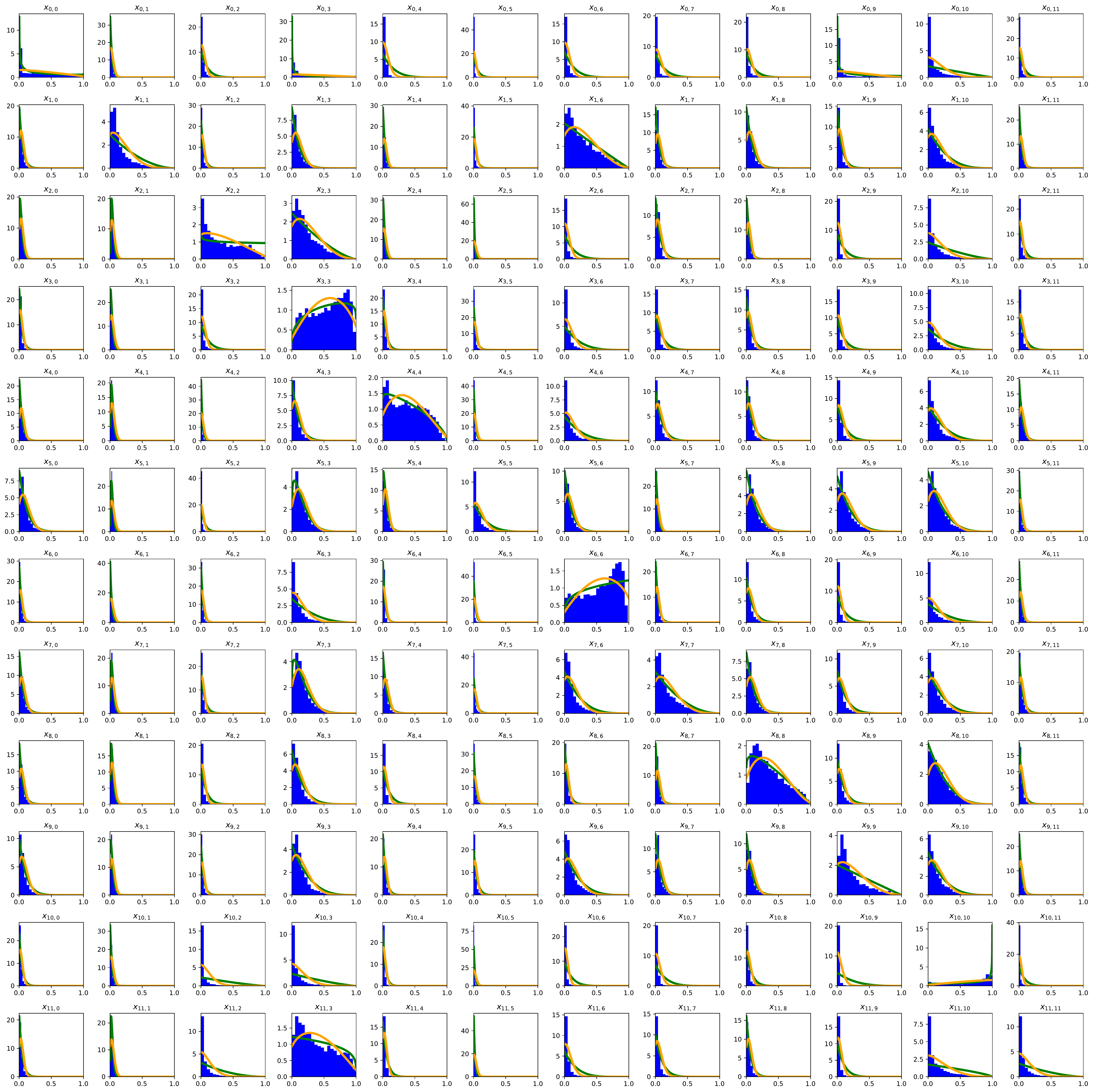}%
\centering
\caption{Histograms per coordinate of softmax predictions on the challenge VISDA-C. Rows correspond to softmax predictions for a given class. Columns refer to softmax prediction coordinates. Green curves represent $\mathtt{Beta}$ density estimations, and orange curves represent $\mathtt{sBeta}$ density estimations. \textit{We invite the reader on the pdf version to zoom on these vectorized figures.}}%
\label{fig_VISDA_C_hists}%
\end{figure*}
\end{document}